\documentclass{article}

\PassOptionsToPackage{numbers,compress}{natbib}
\usepackage[main, final]{neurips_2026}     %


\usepackage[utf8]{inputenc}
\usepackage[T1]{fontenc}
\usepackage[hidelinks]{hyperref}
\usepackage{url}
\usepackage{booktabs}
\usepackage{amsfonts}
\usepackage{amssymb}
\usepackage{amsmath}
\usepackage{nicefrac}
\usepackage{microtype}
\usepackage[table]{xcolor}

\usepackage{pgfplots}
\pgfplotsset{compat=1.18}
\usepgfplotslibrary{groupplots}
\usetikzlibrary{arrows.meta}

\definecolor{accent}{HTML}{E0452C}    %
\definecolor{charcoal}{HTML}{262626}  %
\definecolor{refgray}{HTML}{8C8C8C}   %

\usepackage{graphicx}
\usepackage{subcaption}
\usepackage{afterpage}
\usepackage{multirow}
\usepackage{algorithm}
\usepackage{algpseudocode}
\usepackage{float}
\usepackage{makecell}
\usepackage{threeparttable}
\usepackage{colortbl}

\newcommand{\cmark}{$\checkmark$}
\newcommand{\xmark}{$\times$}

\definecolor{ourrow}{RGB}{220, 232, 245}

\title{$\boldsymbol{A^2}$: Smaller Self-Supervised ViTs\\Localize Better than Larger Ones}

\hypersetup{
  pdftitle={A2: Smaller Self-Supervised ViTs Localize Better than Larger Ones},
  pdfauthor={Sreehari Rammohan, Huy Ha, Carl Vondrick},
  pdfsubject={NeurIPS 2026},
  pdfkeywords={self-supervised learning, vision transformers, attention, localization, spurious correlations}
}

\definecolor{affilcolor}{RGB}{0, 100, 50}

\newcommand{\mi}[1]{\textsuperscript{\textcolor{affilcolor}{\textit{#1}}}}

\author{%
  \textbf{Sreehari Rammohan}\mi{a} \quad
  \textbf{Huy Ha}\mi{b} \quad
  \textbf{Carl Vondrick}\mi{a} \\[6pt]
  {\normalfont
    \mi{a}\,Columbia University \quad
    \mi{b}\,Stanford University
  }%
}

\begin{document}

\maketitle

\begin{figure}[h!]
  \centering
  \input{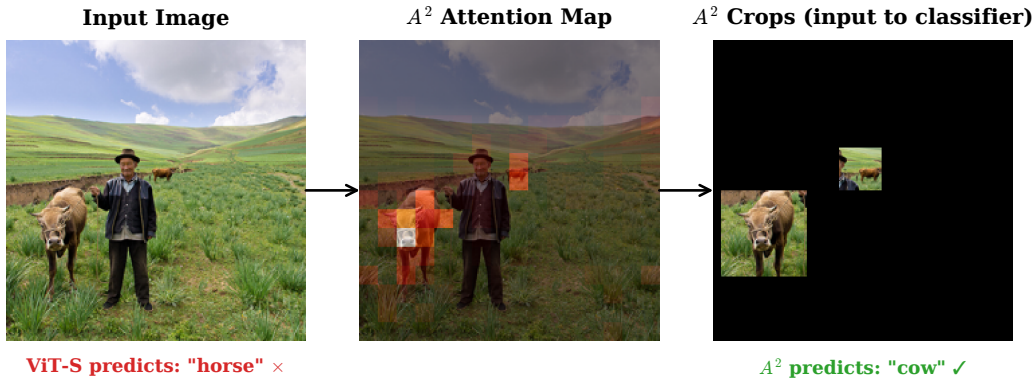}
  \caption{\textbf{Robust Recognition through Exclusion.} Images contain spurious correlations that often mislead classifiers, e.g.\ a man holding a rein is usually associated with horses. $A^2$ removes spatially spurious correlations by automatically selecting crops based on a pre-trained model's attention map. While the original model is misled (left), our $A^2$ trained classifier explicitly attends to only the animal body and correctly predicts ``cow'' (right).}
  \label{fig:teaser}
\end{figure}

\begin{abstract}
Robust visual classification often depends on localizing the main foreground objects in an image while ignoring contextual distractors. Surprisingly, we find that the last layer attention maps of smaller self-supervised ViTs localize foreground objects better than those of larger ViTs. However, we still need large ViTs, because they extract richer representations from each patch. To get the best of both worlds, good localization and rich representations, we propose $A^2$, a simple method that leverages this inverse scaling finding by decoupling where to look (a small attention model) from what to extract (a large embedding model): we crop around the attention peaks of a small model and embed the crops with a larger model. $A^2$ uses entirely pretrained features, requires no group labels, and does not require per-dataset attention or backbone training. Across 5 benchmarks, $A^2$ is competitive with backbone-matched loss-level methods like DFR, and outperforms end-to-end attention training under stronger distribution shifts. Code can be found here: \url{https://github.com/SreehariRamMohan/A2}
\end{abstract}

\section{Introduction}
\label{sec:intro}

Large vision models can approximate a wide range of functions given sufficient training data. However, this flexibility also means the model will exploit all biases and shortcuts \cite{shortcutLearning} in the dataset to minimize the training loss. For example, Figure \ref{fig:teaser} (left) shows a linear classifier trained on ViT \cite{vit} embeddings that falsely predicts that the animal is a horse, when it is actually a cow. Spurious correlations, such as the co-occurrence of horses and people, are pervasive in visual data. Despite progress in curating benchmarks and developing debiasing algorithms, recognition in the presence of such distractors remains a longstanding challenge for computer vision.

Existing methods have tried to correct for these biases by altering the loss function or augmenting the training data. These methods use group labels \cite{groupDRO, dfr}, multiple training passes \cite{jtt, cnc}, end-to-end training of attention \cite{ifam}, an LLM to identify spurious features \cite{roboshot}, or prompts specifying what to segment \cite{sam2, sam3}. Most of these methods try to correct for the spurious correlation after it has already entered the model, or require external prompts or supervision specifying the bias.

A natural question is whether a pretrained attention model can be used to filter out these spurious cues on its own. In many cases, these tasks can simply be solved by isolating the primary foreground object, something that pretrained vision transformers are already good at doing. We find a surprising result: the attention maps in smaller self-supervised ViTs are better at localizing the foreground object than the attention maps in larger ViTs. This inverse scaling trend runs contrary to the common belief that larger parameter-count models are always better \cite{scalingVITs, unreasonableEffectivenessData, scalingTo22B}.

We design $A^2$ to leverage this finding by decoupling where to look (the attention model) from what to extract (the embedding model): one model attends to another model's attention, hence \emph{Attending on Attention}. While pretrained attention might not always be enough to localize the features of interest, for many real-world applications (e.g.\ security camera footage, species identification, medical imaging) the foreground objects are the most task-relevant. This property is also baked into common distribution shift benchmarks (Waterbirds, Spawrious, MetaShift). 

\textbf{Contributions}. We make four contributions: 1) we find that smaller pretrained self-supervised ViTs localize foreground objects substantially better than larger ones in last-layer attention, and that localization quality tracks downstream task performance, 2) we leverage this finding to propose $A^2$, a simple method that decouples \emph{where to look} (the attention model) from \emph{what to extract} (the embedding model) using only frozen pretrained features and is competitive with purpose-built methods for learning under distribution shift, 3) we show that larger models localize in earlier layers, and using an intermediate layer (chosen on a held-out ImageNet validation set) can regain some of the performance of smaller models, and 4) we introduce the MetaShift Animals Dataset, an 8-class animal benchmark with disjoint train and test contexts.

\section{Attention in Smaller Models Localizes Better}

\label{sec:attention_iou}

\afterpage{\begin{figure}[t]
\centering
\definecolor{famDinoOne}{HTML}{7C4607}%
\definecolor{famDinoThree}{HTML}{005586}%
\definecolor{famIbot}{HTML}{0E8E4F}%
\definecolor{famMae}{HTML}{7E5CB7}%
\begin{tikzpicture}[font=\fontsize{8}{9.5}\selectfont]
\begin{groupplot}[
  group style={group size=2 by 1, horizontal sep=0.40in},
  scale only axis, width=2.22in, height=1.05in,
  xmode=log, xmin=15, xmax=9500,
  xtick={21,86,300,1100,6716}, xticklabels={21M,86M,300M,1.1B,6.7B},
  axis lines*=left, axis line style={line width=0.5pt, -},
  tick align=outside, major tick length=2pt, minor tick length=0pt,
  tick style={line width=0.5pt, black},
  tick label style={font=\fontsize{8}{9.5}\selectfont, /pgf/number format/assume math mode=true},
  label style={font=\fontsize{8}{9.5}\selectfont},
  xlabel={Attention model size (parameters)}, xlabel style={yshift=2pt},
  title style={font=\fontsize{8}{9.5}\selectfont\bfseries, at={(0,1)}, anchor=south west, inner sep=0pt, yshift=4pt},
  clip=false,
]
\nextgroupplot[title={Waterbirds}, ylabel={Attention mass in GT box (\%)}, ylabel style={yshift=-2pt},
  ymin=34, ymax=86, ytick={40,50,60,70,80},
  legend to name=loclegend-mass,
  legend style={draw=none, fill=none, legend columns=6, font=\fontsize{8}{9.5}\selectfont,
    /tikz/every even column/.append style={column sep=9pt}}]
\addlegendimage{color=accent, line width=1.2pt, mark=*, mark size=2.4pt, mark options={fill=accent, draw=accent, line width=0pt}}\addlegendentry{\textbf{DINOv2}}
\addlegendimage{color=famDinoOne, line width=0.8pt, mark=square*, mark size=1.9pt, mark options={fill=famDinoOne!35, draw=famDinoOne, line width=0.6pt}}\addlegendentry{DINOv1}
\addlegendimage{color=famDinoThree, line width=0.8pt, mark=diamond*, mark size=1.9pt, mark options={fill=famDinoThree!35, draw=famDinoThree, line width=0.6pt}}\addlegendentry{DINOv3}
\addlegendimage{color=famIbot, line width=0.8pt, mark=triangle*, mark size=1.9pt, mark options={fill=famIbot!35, draw=famIbot, line width=0.6pt}}\addlegendentry{iBOT}
\addlegendimage{color=famMae, line width=0.8pt, mark=*, mark size=1.9pt, mark options={fill=famMae!35, draw=famMae, line width=0.6pt}}\addlegendentry{MAE}
\addlegendimage{color=refgray, line width=0.8pt, mark=pentagon*, mark size=1.9pt, mark options={fill=refgray!35, draw=refgray, line width=0.6pt}}\addlegendentry{OpenCLIP (image--text)}
\addplot[color=refgray, line width=0.8pt, mark=pentagon*, mark size=1.9pt, mark options={fill=refgray!35, draw=refgray, line width=0.6pt}] table[x=params, y=wb_mass, col sep=comma] {Figures/attention_scaling/openclip.csv};
\addplot[color=famDinoOne, line width=0.8pt, mark=square*, mark size=1.9pt, mark options={fill=famDinoOne!35, draw=famDinoOne, line width=0.6pt}] table[x=params, y=wb_mass, col sep=comma] {Figures/attention_scaling/dinov1.csv};
\addplot[color=famDinoThree, line width=0.8pt, mark=diamond*, mark size=1.9pt, mark options={fill=famDinoThree!35, draw=famDinoThree, line width=0.6pt}] table[x=params, y=wb_mass, col sep=comma] {Figures/attention_scaling/dinov3.csv};
\addplot[color=famIbot, line width=0.8pt, mark=triangle*, mark size=1.9pt, mark options={fill=famIbot!35, draw=famIbot, line width=0.6pt}] table[x=params, y=wb_mass, col sep=comma] {Figures/attention_scaling/ibot.csv};
\addplot[color=famMae, line width=0.8pt, mark=*, mark size=1.9pt, mark options={fill=famMae!35, draw=famMae, line width=0.6pt}] table[x=params, y=wb_mass, col sep=comma] {Figures/attention_scaling/mae.csv};
\addplot[color=accent, line width=1.2pt, mark=*, mark size=2.4pt, mark options={fill=accent, draw=accent, line width=0pt}] table[x=params, y=wb_mass, col sep=comma] {Figures/attention_scaling/dinov2.csv};
\nextgroupplot[title={ImageNet val}, ymin=47, ymax=77, ytick={50,60,70}]
\addplot[color=refgray, line width=0.8pt, mark=pentagon*, mark size=1.9pt, mark options={fill=refgray!35, draw=refgray, line width=0.6pt}] table[x=params, y=in_mass, col sep=comma] {Figures/attention_scaling/openclip.csv};
\addplot[color=famDinoOne, line width=0.8pt, mark=square*, mark size=1.9pt, mark options={fill=famDinoOne!35, draw=famDinoOne, line width=0.6pt}] table[x=params, y=in_mass, col sep=comma] {Figures/attention_scaling/dinov1.csv};
\addplot[color=famDinoThree, line width=0.8pt, mark=diamond*, mark size=1.9pt, mark options={fill=famDinoThree!35, draw=famDinoThree, line width=0.6pt}] table[x=params, y=in_mass, col sep=comma] {Figures/attention_scaling/dinov3.csv};
\addplot[color=famIbot, line width=0.8pt, mark=triangle*, mark size=1.9pt, mark options={fill=famIbot!35, draw=famIbot, line width=0.6pt}] table[x=params, y=in_mass, col sep=comma] {Figures/attention_scaling/ibot.csv};
\addplot[color=famMae, line width=0.8pt, mark=*, mark size=1.9pt, mark options={fill=famMae!35, draw=famMae, line width=0.6pt}] table[x=params, y=in_mass, col sep=comma] {Figures/attention_scaling/mae.csv};
\addplot[color=accent, line width=1.2pt, mark=*, mark size=2.4pt, mark options={fill=accent, draw=accent, line width=0pt}] table[x=params, y=in_mass, col sep=comma] {Figures/attention_scaling/dinov2.csv};
\end{groupplot}
\node[anchor=south] at ([xshift=0.20in, yshift=15pt]group c1r1.north east) {\pgfplotslegendfromname{loclegend-mass}};
\end{tikzpicture}%
\caption{\textbf{Smaller self-supervised ViTs localize better.} We show the proportion of last layer attention mass that is inside ground-truth bounding boxes versus model size across six pretraining families, on Waterbirds (left) and ImageNet Val (right). All five self-supervised families (DINOv1, DINOv2, DINOv3, MAE, iBOT) slope downward as model size grows; the contrastive image-text family (OpenCLIP, gray) does not.
}
\label{fig:attention_mass_scaling}
\end{figure}
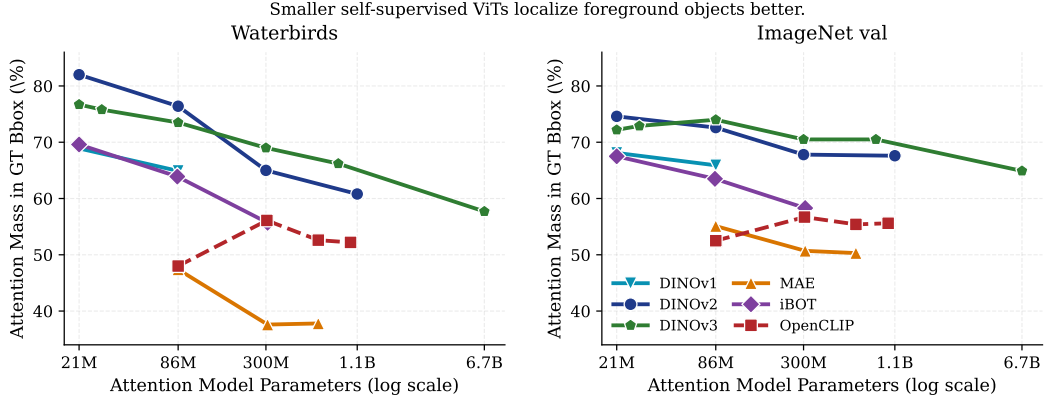}%

To directly test whether smaller self-supervised ViTs \cite{dinov1} localize better than larger ones, we use ImageNet \cite{imagenet} ($1000$ object classes, $N{=}50{,}000$) and Waterbirds ($2$ classes, $N{=}5{,}794$) where annotated bounding boxes are available. We measure the proportion of last-layer CLS attention captured within the bounding box (taking the union if there are multiple) and also report the hit rate (whether or not the highest attention map value falls within a bounding box). 

Figure~\ref{fig:attention_mass_scaling} shows that the trend appears across all five self-supervised families we evaluate. The original DINO \cite{dinov1} shows it at its smallest scale: ViT-S/16 puts $68.9\%$ of its attention mass inside Waterbirds ground truth bounding boxes vs.\ $64.9\%$ for ViT-B/16. For DINOv2 \cite{dinov2}, there is a monotonic decrease in attention mass and hit rate as the ViT gets larger. For ImageNet Val, ViT-S puts $74.5\%$ of its mass inside the ground truth object box vs.\ $67.6\%$ for ViT-G. The same pattern persists at a much larger scale with DINOv3 \cite{dinov3}: on Waterbirds, ViT-S/16 ($21$M) puts $76.7\%$ of its mass inside GT bboxes vs.\ only $57.7\%$ for ViT-7B/16 ($6.7$B), a $19$-point drop while scaling $320\times$ in parameters. Although not monotonic, we observe a similar trend in another self-supervised ViT trained with reconstruction loss, MAE \cite{mae}, with the model best at localization being the smallest, ViT-B/16 (for Waterbirds, ViT-B/16 puts $47.4\%$ of its mass inside the ground truth object box vs. $37.7\%$ for ViT-H/14). The same monotonic decrease in attention mass holds for iBOT \cite{ibot}, a fifth self-supervised family combining masked image modeling with self-distillation: ViT-S/16 puts $69.6\%$ of its mass inside Waterbirds GT bboxes vs.\ $55.7\%$ for ViT-L/16. In contrastive language-image pretrained ViTs (OpenCLIP) the trend does not hold: ViT-B/16 is both the smallest and the worst at localization with a particularly low hit rate of $0.166$ compared with ViT-L/14 ($0.815$). Appendix Table~\ref{tab:attention_iou_all} reports precise values for all metrics.

\textbf{On patch size, objective, and pretraining data.} Finer patches localize better. DINOv1 ViT-S/8 and ViT-B/8 beat their corresponding same-sized patch size 16 counterparts (ViT-S/8 outperforms ViT-S/16). Keeping the patch size fixed, the smaller models localize better (ViT-S/16 is better than ViT-B/16). The plateau of MAE in localization when going from MAE ViT-L/16 to MAE ViT-H/14 (Figure \ref{fig:attention_mass_scaling}) is likely due to the smaller patch size offsetting the inverse scaling trend when moving up in model size. When going from MAE B/16 to MAE L/16, Waterbirds mass in bbox declines 0.098, consistent with inverse scaling, but when going from MAE L/16 to MAE H/14, the smaller patch size appears to counteract inverse scaling, resulting in a slight increase (+0.002) in Waterbirds mass in bbox. 

For objective, we compare DeiT-S/16 (supervised) with DINOv1 ViT-S/16 (self-distilled) as these models are identical except for the objective (both trained on ImageNet-1K). Small size alone is not enough for strong localization, as the DeiT model localizes poorly. 

Finally, the image-text pretraining trend is not stable, varying the pretraining data and comparing OpenCLIP vs. OpenAI CLIP (LAION-2B vs. WIT-400M) suggests that image-text pretrained ViTs show no consistent trend with localization. The same architecture and objective trained on two different corpora yield very different attention hit rates (0.815 vs. 0.172) on Waterbirds. Full results in Appendix Section $\ref{sec:patch_objective_controls}$.

\section{Attending on Attention}

\afterpage{\begin{figure}[t]
\centering
\input{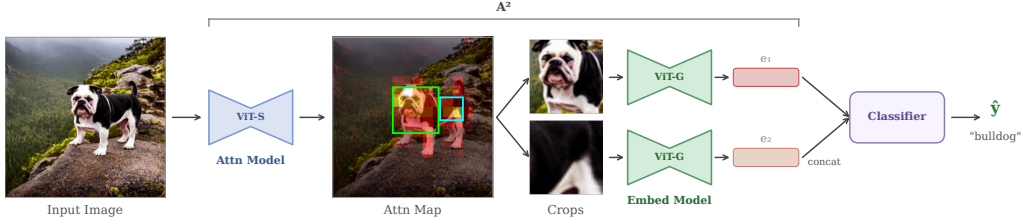}
\caption{\textbf{Overview of $\boldsymbol{A^2}$.} Given an input image, $x$, we use an attention model to select crops, $\mathrm{Crop}_i(x;\;\mathrm{Attn}(x))$, which are resized and embedded using a (possibly different) embedding model, $\mathcal{T}$. The concatenated embeddings
$[\,\mathcal{T}(c_1),\;\ldots,\;\mathcal{T}(c_k)\,]$ are passed to a
classifier $f$ to predict $\hat{y}$.}
\label{fig:method}
\end{figure}}%

Motivated by this finding, we propose Attending on Attention ($A^2$), a method where a larger model attends to the attention of a smaller model.  
$A^2$ has two variants: $A^2_\text{ZS}$, which is fully zero-shot, and $A^2_\text{LR}$ which fits a logistic regression head using crop embeddings from a frozen pretrained model. As shown by Figure \ref{fig:method}, $A^2$ operates in two stages: 1) selection of crops 2) classification based on crops.

Given an image $x$, we predict the label $\hat{y}$ by independently embedding image crops $c_i$ using a transformer $\mathcal{T}$, concatenating the representations, and making a prediction using a classifier $f$. The main idea of our approach is that the crops are computed using the attention maps from a second (possibly different) transformer:
\begin{align}
    \hat{y} = f\!\left(\left[\,\mathcal{T}(c_1),\; \ldots,\; \mathcal{T}(c_k)\,\right]\right) \quad \text{where} \quad c_i = \mathrm{Crop}_i\!\left(x;\; \mathrm{Attn}(x)\right)
\end{align}
where $\mathcal{T}(\cdot)$ is the CLS embedding of a vision transformer. This approach effectively turns the soft-attention $\textrm{Attn}$ into a hard attention \cite{hardAttention}. While $\mathcal{T}$ and $\textrm{Attn}$ could be from the same model, this decoupling also allows us to use one transformer to estimate the attention and a different transformer to predict the label.

\subsection{Crops as Hard Attention}

We crop the image, $x$, around the highest-attention regions in the attention map $\textrm{Attn(x)}$ computed from a vision transformer. This converts the transformer's soft attention to hard attention, allowing only the most attended-to and relevant regions to enter the downstream representation.

Concretely, we use a ViT (e.g., DINOv2/v3 or CLIP) to produce an attention map by taking the dot product of the CLS token's query with every key \cite{attentionAllYouNeed}. Because large ViT models have multiple heads, we choose to compute the mean attention map over all heads (same attention computation as is used in Section \ref{sec:attention_iou}).
We upsample the attention matrix to be the same size as the input image using bilinear interpolation.

Given an $M \times M$ attention map, we greedily select non-overlapping crops $c_1,\dots,c_n$ from a predefined list of crop sizes $s_1,\dots,s_n$ (ordered from largest to smallest). Each crop is chosen so as to maximize the sum of attention mass within it while also not intersecting with a previously placed crop. Formal definitions for both head aggregation and greedy crop selection are given in Appendix~\ref{sec:method_details}.

\subsection{Classification}

Each crop, $c_i$, is resized to $224 \times 224$ before being embedded using $\mathcal{T}$ into a $d$ dimensional representation. $\mathcal{T}$ need not be the same model used to compute $\textrm{Attn(x)}$. In the $A^2_\text{LR}$ variant, we concatenate the crop embeddings $[\,\mathcal{T}(c_1),\;\ldots,\;\mathcal{T}(c_k)\,] \in \mathbb{R}^{k \cdot d}$ and fit a logistic regression classifier to predict $\hat{y}$. Training happens on these frozen crop embeddings and does not require group labels. In the zero-shot variant, $A^2_\text{ZS}$, we select a single crop, $c_1$, embed it using CLIP \cite{clip}  ($\mathcal{T} = \text{CLIP}$), and classify by comparing the cosine similarity of the crop embedding with those of the possible label embeddings (``a photo of a \{\}'').

While the attention map produced from the SSL model is task agnostic, its large scale pretraining allows it to appropriately focus on the relevant foreground objects \cite{dinov1} -- this is often sufficient for performing well on tasks, as shown by $A^2$'s performance. Learning a task conditioned selector using an approach like iFAM \cite{ifam} or GradCAM \cite{gradcam} often overfits to tasks with spurious distractors, underperforming task agnostic selection.

\section{Experiments}

In our experiments, we use five datasets: Spawrious O2O Hard, Spawrious M2M Hard \cite{lynch2023spawrious}, Waterbirds \cite{groupDRO}, MetaShift Cat vs.\ Dog \cite{MetaShift}, and a new dataset we created, MetaShift Animals (Figure~\ref{fig:metashift_animals_main}). Each MetaShift image carries a \emph{context tag} from Visual Genome (e.g., \texttt{tree}, \texttt{water}, \texttt{rope}) which acts as the spurious feature; we split the context tags in each class such that those appearing in training are disjoint from those at test time. The full dataset card with all eight classes and per-class image counts is in Appendix~\ref{sec:metashift_animals_data_card}. Code to construct the MetaShift Animals dataset can be found here: \url{https://github.com/SreehariRamMohan/MetaShift_Animals}

\begin{figure}[t]
\centering
\input{Figures/metashift_animals/giraffe_main.tex}
\caption{\textbf{MetaShift Animals (ours): disjoint train/test contexts.} One of our eight classes (\texttt{giraffe}) shown across several Visual Genome context tags. The top row of each pair contains training images and the bottom row contains test images; the per-image label is the context tag (spurious feature). Train tags (e.g., \texttt{tree}, \texttt{man}, \texttt{dirt}) and test tags (e.g., \texttt{ground}, \texttt{wall}, \texttt{rock}) are disjoint by construction. Full panels for four representative classes in Appendix Figure~\ref{fig:metashift_animals_context_examples}.}
\label{fig:metashift_animals_main}
\end{figure}

We design experiments to answer the following questions:
\begin{itemize}
    \item \textbf{Robustness}: Does $A^2$ improve robustness to common spatial spurious correlations over full-image, last-layer retraining (DFR \cite{dfr}), and attention retraining baselines \cite{ifam}? Can $A^2$ be used with existing methods for spurious correlations that operate at the loss level?
    \item \textbf{Decoupling}: Can decoupling the model used to compute attention and embed crops improve performance as observed in Section \ref{sec:attention_iou}? Is localization performance tied to downstream task performance?
\end{itemize}

For our main $A^2_{\text{LR}}$ cross model results, we use a shared crop configuration across all five benchmarks: one $64\times64$ crop followed by one $32\times32$ crop. Crops are chosen using DINOv2 ViT-S attention and embedded with DINOv2 ViT-G. We believe these settings work well in most cases, but investigate various crop configurations and settings (sizes, upscale options, top-$k$ attention patches instead), along with computational costs in Appendix \ref{sec:crop_count_ablation}.

\subsection{$\boldsymbol{A^2}$ improves robustness and complements loss-level methods}

\begin{table*}[t]
\centering
\caption{\textbf{Worst-Group Accuracy.} This table reports worst-group accuracy (WGA, \%) on five distribution-shift benchmarks. MetaShift Animals reports worst-class accuracy (WC) since its groups are class-context pairs with disjoint train/test contexts. $A^2$ improves worst-group accuracy over full image ViT-G linear probes on three of five benchmarks. Methods that require group labels at training time are shown above the horizontal line and are not directly comparable to the methods below. Backbones are DINOv2~\cite{dinov2} unless prefixed (e.g., OpenCLIP); for $A^2$ rows, the Backbone column reads attention $\times$ embedding. Bold indicates the best value per column among methods not requiring group labels. Values are means over $5$ random seeds where applicable; standard deviations in Appendix Tables~\ref{tab:spawrious}, \ref{tab:waterbirds}, \ref{tab:metashift_catdog}, \ref{tab:metashift_animals}.}
\label{tab:main-wga}
\setlength{\tabcolsep}{3.3pt}%
\scriptsize
\begin{tabular*}{\textwidth}{@{\extracolsep{\fill}}llccccc@{}}
\toprule
Method & Backbone (attn $\times$ emb) & Spawrious O2O & Spawrious M2M & Waterbirds & Cat vs.\ Dog & Animals (WC) \\
\midrule
\multicolumn{7}{@{}l}{\textbf{Deep Feature Reweighting (requires group labels)}} \\
DFR~\cite{dfr}                       & ViT-S (21M)          & 74.9 & 49.9 & 90.4 & 54.5 & 72.0 \\
DFR~\cite{dfr}                       & ViT-B (86M)          & 84.5 & 69.7 & 94.0 & 65.9 & 73.5 \\
DFR~\cite{dfr}                       & ViT-G (1.1B)         & 89.9 & 73.9 & 96.4 & 74.2 & 76.6 \\
DFR$^\dagger$ + $A^2$                & ViT-S $\times$ ViT-S & 88.8 & 64.6 & 91.0 & 67.7 & 72.9 \\
\midrule
\multicolumn{7}{@{}l}{\textbf{Linear probe on full image (no crops)}} \\
LR (full image)                      & ViT-S (21M)          & 71.7 & 25.0 & 72.7 & 56.0 & 63.2 \\
LR (full image)                      & ViT-G (1.1B)         & 77.7 & 52.4 & 92.8 & 76.8 & \textbf{75.9} \\
\addlinespace[3pt]
\multicolumn{7}{@{}l}{\textbf{Attention learning}} \\
iFAM~\cite{ifam} ($K{=}4$)           & ViT-B (86M)          & 80.6 & 71.4 & 96.0 & 72.1 & 68.8 \\
iFAM~\cite{ifam} ($K{=}8$)           & ViT-B (86M)          & 76.1 & 72.4 & \textbf{96.1} & 73.5 & 63.5 \\
GradCAM~\cite{gradcam} & ViT-S (21M) & 45.5 & 16.4 & 68.4 & 43.7 & 61.7 \\
\addlinespace[3pt]
\multicolumn{7}{@{}l}{\textbf{Zero-shot CLIP-based references (no crops)}} \\
TTR~\cite{ttr}                       & OpenCLIP ViT-L-14    & 70.2 & 83.6 & 76.9 & 82.7 & 49.1 \\
OpenCLIP ZS (full image)             & ViT-L-14             & 82.0 & 82.1 & 46.9 & \textbf{88.8} & 58.8 \\
\addlinespace[3pt]
\multicolumn{7}{@{}l}{\textbf{Attending on Attention (ours)}} \\
$A^2_\text{LR}$ cross-model          & ViT-S $\times$ ViT-G & 79.9 & 64.6 & 95.3 & 76.6 & 70.8 \\
$A^2_\text{LR}$                      & ViT-S $\times$ ViT-S & 80.2 & 58.1 & 80.4 & 68.3 & 67.0 \\
$A^2_\text{ZS}$                      & ViT-S $\times$ OpenCLIP ViT-L-14 & \textbf{85.0} & \textbf{85.4} & 64.5 & \textbf{88.8} & 54.7 \\
\bottomrule
\end{tabular*}
\noindent{\scriptsize $\dagger$ = uses group labels via the DFR step.}
\end{table*}

\begin{table*}[t]
\centering
\caption{\textbf{O.O.D.\ Classification Accuracy.} ${A^2}$ does not sacrifice test accuracy, it exceeds full image ViT-G linear probes on three of the five benchmarks and comes within $1.7$ points on the other two. Layout and notation are the same as Table \ref{tab:main-wga}.}
\label{tab:main-acc}
\setlength{\tabcolsep}{4pt}
\scriptsize
\begin{tabular*}{\textwidth}{@{\extracolsep{\fill}}llccccc@{}}
\toprule
Method & Backbone (attn $\times$ emb) & Spawrious O2O & Spawrious M2M & Waterbirds & Cat vs.\ Dog & Animals \\
\midrule
\multicolumn{7}{@{}l}{\textbf{Deep Feature Reweighting (requires group labels)}} \\
DFR~\cite{dfr}                       & ViT-S (21M)          & 86.1 & 73.4 & 92.4 & 58.7 & 88.3 \\
DFR~\cite{dfr}                       & ViT-B (86M)          & 90.1 & 84.9 & 95.5 & 68.6 & 90.3 \\
DFR~\cite{dfr}                       & ViT-G (1.1B)         & 95.7 & 88.0 & 98.1 & 76.0 & 91.4 \\
DFR$^\dagger$ + $A^2$                & ViT-S $\times$ ViT-S & 95.2 & 86.2 & 91.9 & 69.4 & 88.9 \\
\midrule
\multicolumn{7}{@{}l}{\textbf{Linear probe on full image (no crops)}} \\
LR (full image)                      & ViT-S (21M)          & 80.5 & 66.6 & 88.7 & 59.8 & 83.2 \\
LR (full image)                      & ViT-G (1.1B)         & 87.8 & 79.7 & 97.8 & 78.6 & \textbf{90.6} \\
\addlinespace[3pt]
\multicolumn{7}{@{}l}{\textbf{Attention learning}} \\
iFAM~\cite{ifam} ($K{=}4$)           & ViT-B (86M)          & 92.0 & 86.1 & \textbf{98.9} & 78.3 & 86.6 \\
iFAM~\cite{ifam} ($K{=}8$)           & ViT-B (86M)          & 91.2 & 86.0 & 98.8 & 81.1 & 88.2 \\
GradCAM~\cite{gradcam} & ViT-S (21M) & 67.1 & 50.7 & 86.3 & 48.1 & 77.7 \\
\addlinespace[3pt]
\multicolumn{7}{@{}l}{\textbf{Zero-shot CLIP-based references (no crops)}} \\
TTR~\cite{ttr}                       & OpenCLIP ViT-L-14    & 84.4 & 92.1 & 87.2 & 88.1 & 69.2 \\
OpenCLIP ZS (full image)             & ViT-L-14             & 89.8 & 92.4 & 73.5 & \textbf{92.4} & 90.4 \\
\addlinespace[3pt]
\multicolumn{7}{@{}l}{\textbf{Attending on Attention (ours)}} \\
$A^2_\text{LR}$ cross-model          & ViT-S $\times$ ViT-G & 89.2 & 86.0 & 98.3 & 77.3 & 88.9 \\
$A^2_\text{LR}$                      & ViT-S $\times$ ViT-S & 91.0 & 82.4 & 93.0 & 69.8 & 85.1 \\
$A^2_\text{ZS}$                      & ViT-S $\times$ OpenCLIP ViT-L-14 & \textbf{92.8} & \textbf{94.0} & 80.3 & 89.7 & 89.7 \\
\bottomrule
\end{tabular*}
\noindent{\scriptsize $\dagger$ = uses group labels via the DFR step.}
\end{table*}

We show our consolidated results in Tables \ref{tab:main-wga} (worst-group accuracy) and \ref{tab:main-acc} (test accuracy), with the full results and ablations in Appendix Tables \ref{tab:spawrious}, \ref{tab:spawrious_wga_results}, \ref{tab:waterbirds}, \ref{tab:metashift_catdog}, \ref{tab:metashift_animals}. To keep the main comparison backbone-matched (ViT throughout), we omit the historical ResNet-50 baselines (ERM, CORAL, CausIRL, JTT, GroupDRO) from Tables~\ref{tab:main-wga} and \ref{tab:main-acc}. They are reported in Appendix Tables~\ref{tab:spawrious} and~\ref{tab:waterbirds}, where $A^2$ is competitive with them on out-of-domain test accuracy.

\textbf{$\boldsymbol{A^2}$ significantly improves performance when the spatial spurious signal dominates.} At fixed backbone, adding $A^2$ crops to a ViT-S linear probe improves Spawrious M2M Hard WGA from $25.0\%$ to $58.1\%$ ($+33.1\%$). In zero-shot, holding the OpenCLIP ViT-L-14 embedder fixed, $A^2$ crops improve Waterbirds WGA from $46.9\%$ to $64.5\%$ ($+17.6\%$). On benchmarks where the full-image ViT-G probe is already strong (MetaShift Animals, Cat vs.\ Dog accuracy), $A^2$ trails slightly.

\textbf{$\boldsymbol{A^2}$ is complementary to loss-level methods} (e.g.\ DFR, which does last-layer retraining on an equally weighted dataset) and can be used in tandem with these methods to improve performance. Fixing the backbone size to ViT-S, $A^2_\text{LR}$ outperforms DFR on $4/5$ datasets in test accuracy ($3/5$ in worst-group accuracy). $A^2 + \text{DFR}$ adds another $+13.9$ points WGA over DFR alone on Spawrious O2O Hard. $A^2_\text{ZS}$ outperforms TTR (an input filtering method \cite{ttr}) on $4/5$ datasets. On the remaining dataset, Waterbirds, $A^2_\text{LR}$ outperforms TTR by $5.8$ points in accuracy and $3.5$ points in WGA. Full details of our TTR implementation are in Appendix \ref{sec:ttr}.

\textbf{$\boldsymbol{A^2}$ vs.\ end-to-end attention training (iFAM).} We compare against iFAM \cite{ifam}, an attention tuning method that learns its attention head and ViT-B backbone jointly on a target dataset. We ran iFAM at both $K{=}4$ and $K{=}8$ part-discovery configurations on every benchmark and report both in Tables~\ref{tab:main-wga} and~\ref{tab:main-acc}. In Waterbirds, where train and test contexts overlap, iFAM wins ($96.1\%$ vs.\ our $95.3\%$ WGA at $K{=}8$), though iFAM's WGA drops $18.2$ points when tested on a subset of off-center birds (while $A^2$ drops $10.6$ points), showing it may be overfitting to the center bias of the birds in the dataset (Appendix Table \ref{tab:waterbirds_distance_tails}). On the harder distribution-shift dataset MetaShift Animals, $A^2$ wins in both accuracy and worst-class accuracy. The pattern suggests that frozen pretrained DINOv2 attention generalizes across context shifts in a way that iFAM's per-dataset training does not.

In MetaShift Cat vs.\ Dog (binary classification), the OpenCLIP \cite{openclip} zero-shot classifier outperforms the $A^2_\text{LR}$ classifier. The OpenCLIP classifier possesses very strong image-text grounding for the cat and dog labels through its large-scale pretraining. The logistic regression classifier does not start with these priors and the slightest amount of spurious information that leaks into the representations can lead to decreased performance. When the task becomes more difficult ($8$-way classification in MetaShift Animals), $A^2_\text{LR}$ improves over OpenCLIP zero-shot, $+12$-point gain in worst-class accuracy.

Finally, in Table \ref{tab:crop_selector_controls}, to isolate the additional benefit of attention-guided selection over generic cropping, we compare various crop selection policies while keeping the embedding model and crop configuration constant. Small model attention for crop selection achieves the highest worst group accuracy, outperforming random, center, and GradCAM chosen crops.

\begin{table}[h!]
\centering
\caption{$A^2$ vs. other generic cropping policies}
\label{tab:crop_selector_controls}
\small
\begin{tabular}{@{}lcc@{}}
\toprule
Crop selector & Spawrious O2O & Waterbirds \\
\midrule
Full image (no crops) & 77.7 & 92.8 \\
\midrule
Random placement & $25.9 \pm 1.8$ & $47.4 \pm 2.0$ \\
Center placement & 43.5 & 86.3 \\
GradCAM~\cite{gradcam} & 49.2 & 62.9 \\
ViT-G attention, last layer & 73.8 & 89.9 \\
\textbf{ViT-S attention, last layer ($A^2$)} & \textbf{79.9} & \textbf{94.4} \\
\midrule
GT-box-centered crop (uses annotations) & --- & 93.8 \\
\bottomrule
\end{tabular}
\end{table}

\subsection{Localization quality predicts downstream task performance}
\label{sec:decoupling}

\begin{figure}[t]
\centering
\definecolor{dsSky}{HTML}{1C90DD}%
\definecolor{dsIndigo}{HTML}{4B4589}%
\definecolor{dsGreen}{HTML}{168E4D}%
\definecolor{dsOrchid}{HTML}{9F57A5}%
\definecolor{dsSienna}{HTML}{844500}%
\begin{tikzpicture}[font=\fontsize{8}{9.5}\selectfont]
\begin{groupplot}[
  group style={group size=2 by 1, horizontal sep=0.42in},
  scale only axis, width=2.06in, height=1.05in,
  xmode=log, axis lines*=left, axis line style={line width=0.5pt, -},
  tick align=outside, major tick length=2pt, minor tick length=0pt,
  tick style={line width=0.5pt, black},
  tick label style={font=\fontsize{8}{9.5}\selectfont, /pgf/number format/assume math mode=true},
  label style={font=\fontsize{8}{9.5}\selectfont}, xlabel style={yshift=2pt},
  title style={font=\fontsize{8}{9.5}\selectfont\bfseries, at={(0,1)}, anchor=south west, inner sep=0pt, yshift=4pt},
  clip=false,
]
\nextgroupplot[title={Where to look}, xlabel={Selector size (embedder: DINOv2 ViT-G)}, xmin=15.6, xmax=1485.0, xtick={21,86,300,1100}, xticklabels={21M,86M,300M,1.1B}, ymin=-9.07, ymax=0.67, ytick={-8,-6,-4,-2,0}, yticklabels={\textminus8,\textminus6,\textminus4,\textminus2,0}, ylabel={Change vs.\ smallest (points)}, ylabel style={yshift=-2pt},
  legend to name=rellegend-where_what,
  legend style={draw=none, fill=none, legend columns=6, font=\fontsize{8}{9.5}\selectfont,
    /tikz/every even column/.append style={column sep=6pt}},
  legend image code/.code={\draw[mark repeat=2, mark phase=2] plot coordinates {(0cm,0cm) (0.25cm,0cm) (0.5cm,0cm)};}]
\addlegendimage{color=accent, line width=1.2pt, mark=*, mark size=2.4pt, mark options={fill=accent, draw=accent, line width=0pt}}\addlegendentry{\textbf{Mean}}
\addlegendimage{color=dsSky, line width=0.8pt, mark=*, mark size=1.9pt, mark options={fill=dsSky!35, draw=dsSky, line width=0.6pt}}\addlegendentry{Spawrious O2O}
\addlegendimage{color=dsIndigo, line width=0.8pt, mark=square*, mark size=1.9pt, mark options={fill=dsIndigo!35, draw=dsIndigo, line width=0.6pt}}\addlegendentry{Spawrious M2M}
\addlegendimage{color=dsGreen, line width=0.8pt, mark=triangle*, mark size=1.9pt, mark options={fill=dsGreen!35, draw=dsGreen, line width=0.6pt}}\addlegendentry{Waterbirds}
\addlegendimage{color=dsOrchid, line width=0.8pt, mark=diamond*, mark size=1.9pt, mark options={fill=dsOrchid!35, draw=dsOrchid, line width=0.6pt}}\addlegendentry{Cat vs.\ Dog}
\addlegendimage{color=dsSienna, line width=0.8pt, mark=pentagon*, mark size=1.9pt, mark options={fill=dsSienna!35, draw=dsSienna, line width=0.6pt}}\addlegendentry{Animals}
\addplot[refgray, line width=0.5pt, forget plot] coordinates {(15.6,0) (1485.0,0)};
\addplot[color=dsSky, line width=0.8pt, mark=*, mark size=1.9pt, mark options={fill=dsSky!35, draw=dsSky, line width=0.6pt}] table[x=params, y=o2o, col sep=comma] {Figures/relative_scaling/where_what_left.csv};
\addplot[color=dsIndigo, line width=0.8pt, mark=square*, mark size=1.9pt, mark options={fill=dsIndigo!35, draw=dsIndigo, line width=0.6pt}] table[x=params, y=m2m, col sep=comma] {Figures/relative_scaling/where_what_left.csv};
\addplot[color=dsGreen, line width=0.8pt, mark=triangle*, mark size=1.9pt, mark options={fill=dsGreen!35, draw=dsGreen, line width=0.6pt}] table[x=params, y=wb, col sep=comma] {Figures/relative_scaling/where_what_left.csv};
\addplot[color=dsOrchid, line width=0.8pt, mark=diamond*, mark size=1.9pt, mark options={fill=dsOrchid!35, draw=dsOrchid, line width=0.6pt}] table[x=params, y=catdog, col sep=comma] {Figures/relative_scaling/where_what_left.csv};
\addplot[color=dsSienna, line width=0.8pt, mark=pentagon*, mark size=1.9pt, mark options={fill=dsSienna!35, draw=dsSienna, line width=0.6pt}] table[x=params, y=animals, col sep=comma] {Figures/relative_scaling/where_what_left.csv};
\addplot[color=accent, line width=1.2pt, mark=*, mark size=2.4pt, mark options={fill=accent, draw=accent, line width=0pt}] table[x=params, y=mean, col sep=comma] {Figures/relative_scaling/where_what_left.csv};
\node[anchor=west, xshift=3pt, fill=white, inner sep=1pt] at (axis cs:1100,-4.10) {\textminus4.1};
\nextgroupplot[title={What to extract}, xlabel={Embedder size (selector: DINOv2 ViT-S)}, xmin=15.6, xmax=1485.0, xtick={21,86,300,1100}, xticklabels={21M,86M,300M,1.1B}, ymin=-3.06, ymax=15.26, ytick={0,5,10,15}, yticklabels={0,+5,+10,+15}]
\addplot[refgray, line width=0.5pt, forget plot] coordinates {(15.6,0) (1485.0,0)};
\addplot[color=dsSky, line width=0.8pt, mark=*, mark size=1.9pt, mark options={fill=dsSky!35, draw=dsSky, line width=0.6pt}] table[x=params, y=o2o, col sep=comma] {Figures/relative_scaling/where_what_right.csv};
\addplot[color=dsIndigo, line width=0.8pt, mark=square*, mark size=1.9pt, mark options={fill=dsIndigo!35, draw=dsIndigo, line width=0.6pt}] table[x=params, y=m2m, col sep=comma] {Figures/relative_scaling/where_what_right.csv};
\addplot[color=dsGreen, line width=0.8pt, mark=triangle*, mark size=1.9pt, mark options={fill=dsGreen!35, draw=dsGreen, line width=0.6pt}] table[x=params, y=wb, col sep=comma] {Figures/relative_scaling/where_what_right.csv};
\addplot[color=dsOrchid, line width=0.8pt, mark=diamond*, mark size=1.9pt, mark options={fill=dsOrchid!35, draw=dsOrchid, line width=0.6pt}] table[x=params, y=catdog, col sep=comma] {Figures/relative_scaling/where_what_right.csv};
\addplot[color=dsSienna, line width=0.8pt, mark=pentagon*, mark size=1.9pt, mark options={fill=dsSienna!35, draw=dsSienna, line width=0.6pt}] table[x=params, y=animals, col sep=comma] {Figures/relative_scaling/where_what_right.csv};
\addplot[color=accent, line width=1.2pt, mark=*, mark size=2.4pt, mark options={fill=accent, draw=accent, line width=0pt}] table[x=params, y=mean, col sep=comma] {Figures/relative_scaling/where_what_right.csv};
\node[anchor=west, xshift=3pt, fill=white, inner sep=1pt] at (axis cs:1100,5.58) {+5.6};
\end{groupplot}
\node[anchor=south] at ([xshift=0.21in, yshift=15pt]group c1r1.north east) {\pgfplotslegendfromname{rellegend-where_what}};
\end{tikzpicture}%
\caption{\textbf{Where and What Scaling Laws.} Larger embedding models but smaller attention models improve performance. Each panel plots each dataset's standard metric (test accuracy for Spawrious; worst-group accuracy for Waterbirds and Cat vs.\ Dog; worst-class accuracy for Animals) relative to the smallest model (thin colored) and the mean of those relative changes across 5 datasets (bold red). \textbf{Left (Where to Look)}: With the embedding model fixed to DINOv2 ViT-G, every dataset curve slopes downward as the attention model grows. \textbf{Right (What to Extract)}: With the attention model fixed to DINOv2 ViT-S, most dataset curves slope upward as the embedding model grows. Absolute values are in Appendix Tables \ref{tab:cross_o2o}, \ref{tab:cross_m2m}, \ref{tab:cross_waterbirds}, \ref{tab:cross_catdog}, and \ref{tab:cross_animals}. The DINOv3 \cite{dinov3} family shows the same inverse-scaling-of-attention pattern (Appendix Figure \ref{fig:dinov3_where_to_look_scaling_mean}).}
\label{fig:where_what_scaling}
\end{figure}

\begin{figure}[t]
\centering
\definecolor{dsSky}{HTML}{1C90DD}%
\definecolor{dsIndigo}{HTML}{4B4589}%
\definecolor{dsGreen}{HTML}{168E4D}%
\definecolor{dsOrchid}{HTML}{9F57A5}%
\definecolor{dsSienna}{HTML}{844500}%
\begin{tikzpicture}[font=\fontsize{8}{9.5}\selectfont]
\begin{groupplot}[
  group style={group size=2 by 1, horizontal sep=0.42in},
  scale only axis, width=2.06in, height=1.05in,
  xmode=log, axis lines*=left, axis line style={line width=0.5pt, -},
  tick align=outside, major tick length=2pt, minor tick length=0pt,
  tick style={line width=0.5pt, black},
  tick label style={font=\fontsize{8}{9.5}\selectfont, /pgf/number format/assume math mode=true},
  label style={font=\fontsize{8}{9.5}\selectfont}, xlabel style={yshift=2pt},
  title style={font=\fontsize{8}{9.5}\selectfont\bfseries, at={(0,1)}, anchor=south west, inner sep=0pt, yshift=4pt},
  clip=false,
]
\nextgroupplot[title={DINOv2 selectors}, xlabel={Selector size (embedder: OpenCLIP ViT-L/14)}, xmin=15.6, xmax=1485.0, xtick={21,86,300,1100}, xticklabels={21M,86M,300M,1.1B}, ymin=-3.61, ymax=0.91, ytick={-3,-2,-1,0}, yticklabels={\textminus3,\textminus2,\textminus1,0}, ylabel={Test acc.\ change (points)}, ylabel style={yshift=-2pt},
  legend to name=rellegend-clipzs_acc,
  legend style={draw=none, fill=none, legend columns=6, font=\fontsize{8}{9.5}\selectfont,
    /tikz/every even column/.append style={column sep=6pt}},
  legend image code/.code={\draw[mark repeat=2, mark phase=2] plot coordinates {(0cm,0cm) (0.25cm,0cm) (0.5cm,0cm)};}]
\addlegendimage{color=accent, line width=1.2pt, mark=*, mark size=2.4pt, mark options={fill=accent, draw=accent, line width=0pt}}\addlegendentry{\textbf{Mean}}
\addlegendimage{color=dsSky, line width=0.8pt, mark=*, mark size=1.9pt, mark options={fill=dsSky!35, draw=dsSky, line width=0.6pt}}\addlegendentry{Spawrious O2O}
\addlegendimage{color=dsIndigo, line width=0.8pt, mark=square*, mark size=1.9pt, mark options={fill=dsIndigo!35, draw=dsIndigo, line width=0.6pt}}\addlegendentry{Spawrious M2M}
\addlegendimage{color=dsGreen, line width=0.8pt, mark=triangle*, mark size=1.9pt, mark options={fill=dsGreen!35, draw=dsGreen, line width=0.6pt}}\addlegendentry{Waterbirds}
\addlegendimage{color=dsOrchid, line width=0.8pt, mark=diamond*, mark size=1.9pt, mark options={fill=dsOrchid!35, draw=dsOrchid, line width=0.6pt}}\addlegendentry{Cat vs.\ Dog}
\addlegendimage{color=dsSienna, line width=0.8pt, mark=pentagon*, mark size=1.9pt, mark options={fill=dsSienna!35, draw=dsSienna, line width=0.6pt}}\addlegendentry{Animals}
\addplot[refgray, line width=0.5pt, forget plot] coordinates {(15.6,0) (1485.0,0)};
\addplot[color=dsSky, line width=0.8pt, mark=*, mark size=1.9pt, mark options={fill=dsSky!35, draw=dsSky, line width=0.6pt}] table[x=params, y=o2o, col sep=comma] {Figures/relative_scaling/clipzs_acc_left.csv};
\addplot[color=dsIndigo, line width=0.8pt, mark=square*, mark size=1.9pt, mark options={fill=dsIndigo!35, draw=dsIndigo, line width=0.6pt}] table[x=params, y=m2m, col sep=comma] {Figures/relative_scaling/clipzs_acc_left.csv};
\addplot[color=dsGreen, line width=0.8pt, mark=triangle*, mark size=1.9pt, mark options={fill=dsGreen!35, draw=dsGreen, line width=0.6pt}] table[x=params, y=wb, col sep=comma] {Figures/relative_scaling/clipzs_acc_left.csv};
\addplot[color=dsOrchid, line width=0.8pt, mark=diamond*, mark size=1.9pt, mark options={fill=dsOrchid!35, draw=dsOrchid, line width=0.6pt}] table[x=params, y=catdog, col sep=comma] {Figures/relative_scaling/clipzs_acc_left.csv};
\addplot[color=dsSienna, line width=0.8pt, mark=pentagon*, mark size=1.9pt, mark options={fill=dsSienna!35, draw=dsSienna, line width=0.6pt}] table[x=params, y=animals, col sep=comma] {Figures/relative_scaling/clipzs_acc_left.csv};
\addplot[color=accent, line width=1.2pt, mark=*, mark size=2.4pt, mark options={fill=accent, draw=accent, line width=0pt}] table[x=params, y=mean, col sep=comma] {Figures/relative_scaling/clipzs_acc_left.csv};
\node[anchor=west, xshift=3pt, fill=white, inner sep=1pt] at (axis cs:1100,-0.98) {\textminus1.0};
\nextgroupplot[title={OpenCLIP selectors}, xlabel={Selector size (embedder: OpenCLIP ViT-L/14)}, xmin=63.7, xmax=1350.0, xtick={86,304,632,1000}, xticklabels={86M,304M,632M,1.0B}, ymin=-0.63, ymax=8.53, ytick={0,2,4,6,8}, yticklabels={0,+2,+4,+6,+8}]
\addplot[refgray, line width=0.5pt, forget plot] coordinates {(63.7,0) (1350.0,0)};
\addplot[color=dsSky, line width=0.8pt, mark=*, mark size=1.9pt, mark options={fill=dsSky!35, draw=dsSky, line width=0.6pt}] table[x=params, y=o2o, col sep=comma] {Figures/relative_scaling/clipzs_acc_right.csv};
\addplot[color=dsIndigo, line width=0.8pt, mark=square*, mark size=1.9pt, mark options={fill=dsIndigo!35, draw=dsIndigo, line width=0.6pt}] table[x=params, y=m2m, col sep=comma] {Figures/relative_scaling/clipzs_acc_right.csv};
\addplot[color=dsGreen, line width=0.8pt, mark=triangle*, mark size=1.9pt, mark options={fill=dsGreen!35, draw=dsGreen, line width=0.6pt}] table[x=params, y=wb, col sep=comma] {Figures/relative_scaling/clipzs_acc_right.csv};
\addplot[color=dsOrchid, line width=0.8pt, mark=diamond*, mark size=1.9pt, mark options={fill=dsOrchid!35, draw=dsOrchid, line width=0.6pt}] table[x=params, y=catdog, col sep=comma] {Figures/relative_scaling/clipzs_acc_right.csv};
\addplot[color=dsSienna, line width=0.8pt, mark=pentagon*, mark size=1.9pt, mark options={fill=dsSienna!35, draw=dsSienna, line width=0.6pt}] table[x=params, y=animals, col sep=comma] {Figures/relative_scaling/clipzs_acc_right.csv};
\addplot[color=accent, line width=1.2pt, mark=*, mark size=2.4pt, mark options={fill=accent, draw=accent, line width=0pt}] table[x=params, y=mean, col sep=comma] {Figures/relative_scaling/clipzs_acc_right.csv};
\node[anchor=west, xshift=3pt, fill=white, inner sep=1pt] at (axis cs:1000,3.64) {+3.6};
\end{groupplot}
\node[anchor=south] at ([xshift=0.21in, yshift=15pt]group c1r1.north east) {\pgfplotslegendfromname{rellegend-clipzs_acc}};
\end{tikzpicture}%
\caption{\textbf{Downstream Task Performance.}  ${A^2_{\text{ZS}}}$ tracks localization quality instead of model size. We fix the embedder to OpenCLIP ViT-L/14 and vary the attention model. Each panel plots per-dataset performance relative to the smallest model in that family (thin colored) and the mean of those relative changes across 5 datasets (bold red). \textbf{Left}: With a DINOv2 crop selector, performance \emph{decreases} as the attention model grows on most datasets (curves below zero). \textbf{Right}: An OpenCLIP selector does not show the same inverse scaling, ViT-B/16 (the smallest) is the worst. Appendix Figure~\ref{fig:clip_zs_family_scaling_wga} shows the worst-group accuracy version; Appendix Table~\ref{tab:clip_zs_attn_scaling} reports the precise absolute values.}
\label{fig:clip_zs_attn_scaling}
\end{figure}

A natural question is whether downstream $A^2$ performance mimics the trends seen in Section~\ref{sec:attention_iou}.

For $A^2_\text{LR}$, we fix the crop embedder to be DINOv2 ViT-G, and vary the model used to compute attention (left panel in Figure \ref{fig:where_what_scaling}). To test the direct scaling of embedding size, we fix the attention model to be DINOv2 ViT-S, and vary the model used to embed the crops (right panel in Figure \ref{fig:where_what_scaling}). For $A^2_\text{ZS}$, we fix the embedding model to OpenCLIP ViT-L/14 and vary the crop selector within the DINOv2 family and OpenCLIP family.

Across both $A^2$ variants, localization ability of the underlying self-supervised model is indicative of downstream task performance. We see the same downward sloping trend in task performance as the DINOv2 model used to select a crop scales (Figure \ref{fig:clip_zs_attn_scaling}). OpenCLIP does not show this trend, and worst-group accuracy follows the same family pattern (Appendix Figure~\ref{fig:clip_zs_family_scaling_wga}). On Spawrious and Waterbirds, $A^2_\text{ZS}$ using ViT-S attention improves the test accuracy and worst-group accuracy over standard CLIP classification using the full image (improving WGA for Waterbirds from $46.9\%$ to $64.5\%$). Across all $5$ datasets, in $18/20$ cases, the best results are achieved with ViT-S (the smallest model). In fact, pairing a small selector with a large embedder (ViT-S $\times$ ViT-G) outperforms using the large model for both (ViT-G $\times$ ViT-G) on every dataset. Full cross-model pairings in Appendix Tables \ref{tab:cross_o2o}-\ref{tab:cross_animals}.

\section{Discussion}

\textbf{Inverse Scaling in Self-Supervised ViTs.} 

We first rule out simple head averaging as the contributor (ViT-G has $24$ attention heads vs.\ $6$ in ViT-S, and averaging more heads should, by concavity of entropy, yield more diffuse attention). We explored trying different head aggregation strategies (Appendix Table \ref{tab:agg_ablation_imagenet_vitg}), and while this ultimately reduced the attention entropy of the selection map, it did not increase downstream task performance (the models were concentrating attention on objects besides the foreground). 

Next, we look at downstream localization from the attention map at every layer in the ViT --- we notice a premature peak in localization ability for larger models (which also coincides with the model shifting a majority of its attention from patch tokens to register tokens, Figure~\ref{fig:register_attention_vs_localization}). In Appendix Figure \ref{fig:per_layer_attention_hit_noreg}, we are able to reproduce the observation in \cite{vitregister} of high-norm tokens appearing in the attention maps of larger ViTs midway through the model, but note that this inverse scaling effect still holds in DINOv2 \textit{with registers} (see Figure \ref{fig:per_layer_attention_mass}). 

Lappe et al.~\cite{lappe} observed that even with registers, while the attention maps in models with registers are visually cleaner, they do not properly aggregate local information and pay a disproportionate amount of attention to register tokens over patch tokens. In ViT-G on Waterbirds, the fraction of CLS attention on register tokens grows from $41\%$ at the peak-localization layer (block~32) to $55\%$ at the last block, while the patch-side attention falls from $57\%$ to $40\%$ and the patch-side mass-in-bbox drops $13$ points over the same span. ViT-S and ViT-B never undergo this register takeover and continue to localize through their last block without a premature peak. 

\textbf{Rethinking the use of last layer attention}

The late stage degradation seen in Appendix Figure \ref{fig:register_attention_vs_localization} naturally raises the question of whether we can use the same model for selection and embedding, provided we select from the image at the peak localization layer. 

We use the ImageNet Validation set to calibrate, choosing the layer whose attention map places most mass in the ground truth bounding boxes. Once found, we keep this selection choice fixed for our five experiments to test transfer. We see particularly pronounced gains on the largest model size from the MAE backbone, where using block $18$ instead of block 31 for attention yields over $20$ points more in worst group/class accuracy for Waterbirds and MetaShift Animals (Appendix Table \ref{tab:imagenet_layer_transfer}). In the $60$ cases where the selected layer index differs from the terminal index, $40$ result in improved performance, $19$ reduce performance, and $1$ leads to no change. 

Using the smallest model in the family as a dedicated selector is still preferable over using the peak localization layer in the largest model in each family for three reasons: 1) using the smallest model in a family as a dedicated selector scores higher in $16/25$ cases compared with using the largest model's peak localization layer (see Appendix Table \ref{tab:a2_vs_peak_localization}), 2) the smallest model in the family is computationally more efficient for attention computation, and 3) $A^2$ forgoes the need for calibration to find the peak localization layer.

\begin{figure}[t!]
\centering
\definecolor{sizeS}{HTML}{0FA49D}%
\definecolor{sizeB}{HTML}{C65B2D}%
\definecolor{sizeL}{HTML}{943F69}%
\definecolor{sizeG}{HTML}{363FB4}%
\begin{tikzpicture}[font=\fontsize{8}{9.5}\selectfont]
\begin{groupplot}[
  group style={group size=2 by 1, horizontal sep=0.40in},
  scale only axis, width=2.18in, height=1.05in,
  xmin=-1.5, xmax=41.5, xtick={0,11,23,39}, xlabel={Block (0-indexed)}, xlabel style={yshift=2pt},
  axis lines*=left, axis line style={line width=0.5pt, -},
  tick align=outside, major tick length=2pt, minor tick length=0pt,
  tick style={line width=0.5pt, black},
  tick label style={font=\fontsize{8}{9.5}\selectfont, /pgf/number format/assume math mode=true},
  label style={font=\fontsize{8}{9.5}\selectfont},
  title style={font=\fontsize{8}{9.5}\selectfont\bfseries, at={(0,1)}, anchor=south west, inner sep=0pt, yshift=4pt},
  clip=false,
]
\nextgroupplot[title={Waterbirds}, ymin=29.8, ymax=85.0, ytick={40,50,60,70,80}, ylabel={Attention mass in GT box (\%)}, ylabel style={yshift=-2pt},
  legend to name=perlayer-mass-reg,
  legend style={draw=none, fill=none, legend columns=6, font=\fontsize{8}{9.5}\selectfont,
    /tikz/every even column/.append style={column sep=8pt}}]
\addlegendimage{color=sizeS, line width=1pt}\addlegendentry{ViT-S}
\addlegendimage{color=sizeB, line width=1pt}\addlegendentry{ViT-B}
\addlegendimage{color=sizeL, line width=1pt}\addlegendentry{ViT-L}
\addlegendimage{color=sizeG, line width=1pt}\addlegendentry{ViT-G}
\addlegendimage{only marks, mark=*, mark size=1.9pt, mark options={fill=white, draw=black, line width=0.7pt}}\addlegendentry{peak block}
\addlegendimage{only marks, mark=*, mark size=1.9pt, mark options={fill=black, draw=black, line width=0.7pt}}\addlegendentry{last block}
\addplot[color=sizeS, line width=1pt] table[x=block, y=mass, col sep=comma] {Figures/per_layer/reg_wb_vits.csv};
\addplot[only marks, mark=*, mark size=1.9pt, mark options={fill=sizeS, draw=sizeS, line width=0.7pt}] coordinates {(11,82.03)};
\addplot[color=sizeB, line width=1pt] table[x=block, y=mass, col sep=comma] {Figures/per_layer/reg_wb_vitb.csv};
\addplot[only marks, mark=*, mark size=1.9pt, mark options={fill=sizeB, draw=sizeB, line width=0.7pt}] coordinates {(11,76.39)};
\addplot[color=sizeL, line width=1pt] table[x=block, y=mass, col sep=comma] {Figures/per_layer/reg_wb_vitl.csv};
\addplot[only marks, mark=*, mark size=1.9pt, mark options={fill=white, draw=sizeL, line width=0.7pt}] coordinates {(21,76.77)};
\addplot[only marks, mark=*, mark size=1.9pt, mark options={fill=sizeL, draw=sizeL, line width=0.7pt}] coordinates {(23,65.01)};
\addplot[color=sizeG, line width=1pt] table[x=block, y=mass, col sep=comma] {Figures/per_layer/reg_wb_vitg.csv};
\addplot[only marks, mark=*, mark size=1.9pt, mark options={fill=white, draw=sizeG, line width=0.7pt}] coordinates {(32,73.59)};
\addplot[only marks, mark=*, mark size=1.9pt, mark options={fill=sizeG, draw=sizeG, line width=0.7pt}] coordinates {(39,60.84)};
\node[anchor=west, xshift=2.5pt, yshift=0pt, inner sep=1pt] at (axis cs:39,60.84) {\textminus13};
\node[anchor=north east, xshift=-1.5pt, yshift=-1.5pt, inner sep=1pt] at (axis cs:23,65.01) {\textminus12};
\nextgroupplot[title={ImageNet val}, ymin=44.6, ymax=76.2, ytick={50,60,70}]
\addplot[color=sizeS, line width=1pt] table[x=block, y=mass, col sep=comma] {Figures/per_layer/reg_in_vits.csv};
\addplot[only marks, mark=*, mark size=1.9pt, mark options={fill=sizeS, draw=sizeS, line width=0.7pt}] coordinates {(11,74.55)};
\addplot[color=sizeB, line width=1pt] table[x=block, y=mass, col sep=comma] {Figures/per_layer/reg_in_vitb.csv};
\addplot[only marks, mark=*, mark size=1.9pt, mark options={fill=sizeB, draw=sizeB, line width=0.7pt}] coordinates {(11,72.61)};
\addplot[color=sizeL, line width=1pt] table[x=block, y=mass, col sep=comma] {Figures/per_layer/reg_in_vitl.csv};
\addplot[only marks, mark=*, mark size=1.9pt, mark options={fill=white, draw=sizeL, line width=0.7pt}] coordinates {(21,73.75)};
\addplot[only marks, mark=*, mark size=1.9pt, mark options={fill=sizeL, draw=sizeL, line width=0.7pt}] coordinates {(23,67.84)};
\addplot[color=sizeG, line width=1pt] table[x=block, y=mass, col sep=comma] {Figures/per_layer/reg_in_vitg.csv};
\addplot[only marks, mark=*, mark size=1.9pt, mark options={fill=white, draw=sizeG, line width=0.7pt}] coordinates {(33,71.53)};
\addplot[only marks, mark=*, mark size=1.9pt, mark options={fill=sizeG, draw=sizeG, line width=0.7pt}] coordinates {(39,67.63)};
\node[anchor=west, xshift=2.5pt, yshift=0pt, inner sep=1pt] at (axis cs:39,67.63) {\textminus4};
\node[anchor=west, xshift=2.5pt, yshift=0pt, inner sep=1pt] at (axis cs:23,67.84) {\textminus6};
\end{groupplot}
\node[anchor=south] at ([xshift=0.20in, yshift=15pt]group c1r1.north east) {\pgfplotslegendfromname{perlayer-mass-reg}};
\end{tikzpicture}%
\caption{\textbf{Per-layer CLS-to-patches attention mass for DINOv2 with register tokens~\cite{vitregister}.} Filled dots mark each model's last block (the layer used in our other figures) and open dots its peak; numbers give the drop from peak to last block, in points. ViT-S and ViT-B peak at the last block; ViT-L peaks at block 21 of 24 and ViT-G at block 32 of 40, dropping $12$--$13$ points on Waterbirds (left) and $4$--$6$ points on ImageNet val (right) by the final block. Appendix Figure~\ref{fig:per_layer_attention_mass_noreg} shows the same plot for DINOv2 \emph{without} registers; per-layer hit-rate counterparts are in Appendix Figures~\ref{fig:per_layer_attention_hit_reg} and \ref{fig:per_layer_attention_hit_noreg}.}
\label{fig:per_layer_attention_mass}
\end{figure}

\textbf{End-to-end attention training overfits; pretrained attention generalizes.} End-to-end attention training (iFAM) \cite{ifam} wins on Waterbirds, where there is no strong distribution shift between train and test, but the dataset is heavily biased toward placing the bird at the center; iFAM appears to lean on this bias: its WGA drops $18.2$ points from $100.0\%$ on the $1\%$ most-centered test images to $81.8\%$ on the $1\%$ most off-center, while $A^2$'s drop is roughly half as steep ($83.3\% \to 72.7\%$; Appendix Table \ref{tab:waterbirds_distance_tails}). As the test distribution strays further from train, $A^2$ actually outperforms iFAM despite using entirely pretrained features and a linear head: on MetaShift Animals, where train and test contexts are disjoint, $A^2$ beats iFAM by $0.7$ points in accuracy and $2$ points in worst-class accuracy. Pretrained attention offers a gentler degradation curve than fine-tuned attention, which tends to overfit train-task distribution biases.

\section{Related Work}

\subsection{Localization of SSL models}
Isolated instances of inverse scaling of localization appear in prior work, but if present, they are entangled within the method itself. To the best of our knowledge, we are the first to observe inverse scaling of localization as a general property of self-supervised model pretraining. The authors of LOST \cite{LOST} found that DINOv1 ViT-S was slightly better at localizing using their method for single object discovery than DINOv1 ViT-B (61.9/64.0/50.7 vs. 60.1/63.3/50.0) in CorLoc performance on VOC07/VOC12/COCO\_20K --- however TokenCut \cite{tokencut} mentions that the seed expansion heuristic in LOST is very sensitive to backbone, so identifying this as a scaling trend through the method is tenuous. In the vision language space, GEM \cite{gem} found that GEM-ViT-B/16 consistently outperforms GEM-ViT-L/14 on localization tasks, but this is only after using self-self attention rather than the model's own attention.

\subsection{Data augmentation}

Approaches that modify what the model attends to fall into two broad buckets: those that try to include more visual information and those that try to include less. $S^2$ \cite{scalingScales} represents the former, showing that increased performance can be achieved by constructing representations based on the merging of multiple image scales (e.g., concatenating $224^2$ and $448^2$ features), and that scaling on scales can outperform scaling on model size. In contrast, $A^2$ focuses on reducing the number of image tokens the model sees.

In the space of approaches that try to include less visual information, LookWhere \cite{fuller2025} also splits vision into where to look followed by what to extract. A selector (distilled from a self-supervised teacher) predicts patch importance and passes the top-$k$ patches (along with low-resolution selector tokens) to a deeper extractor network for processing. iFAM \cite{ifam} uses a two-stage system trained jointly on each target dataset: stage 1 learns task-specific binary attention masks via part-discovery losses, and stage 2 fine-tunes a ViT classifier on the masked input using class labels. 
TTR \cite{ttr} improves CLIP's zero-shot classification abilities by first isolating a ``core object'' at test time for each image, which is found by a combination of PCA and Kmeans clustering on the CLIP patch embeddings together with a hand-crafted prompt (``a photo of a bird''). LookWhere, iFAM and TTR are similar to $A^2$ in that they can keep the spurious feature from entering the classifier; however, iFAM and LookWhere train their selectors (LookWhere's focus is on efficiency, not robustness) and TTR uses clustering and a separate prompt to isolate the main cluster. $A^2$ uses solely frozen attention from pretrained SSL models. 

Segmentation models like SAM2 \cite{sam2} or SAM3 \cite{sam3} require concept prompts, clicks, boxes, or masks as input to the model to know what to segment. This requires the user to know the relevant object to segment at evaluation time. $A^2$ identifies the salient regions using the model's own attention.

\subsection{Regularization, reweighting, and retraining}

Assuming one has access to group labels, GroupDRO \cite{groupDRO} trains a model to minimize the maximum loss over all groups. In Deep Feature Reweighting (DFR) \cite{dfr}, the authors show that an ERM trained classifier can be modified to perform well on spurious correlation tasks by simply re-training the final layer on a small validation set where all groups are represented equally. When benchmarking against DFR, we use the same ViT backbone to ensure fair comparisons.

When one doesn't have access to group labels, JTT \cite{jtt} proposes a two-stage system, first fitting an ERM model and then training a second model that upweights the data the first model has difficulty predicting. CnC, Correct-N-Contrast \cite{cnc}, similarly uses a two-stage ERM-then-contrastive system, encouraging same-class samples from different groups to be close together in representation space using supervised contrastive learning \cite{supervisedContrastiveLearning}.

Finally in the zero-shot setting, Roboshot \cite{roboshot} uses a language model given a task description to generate spurious and helpful text descriptions, then projects these descriptions into the input embedding space to amplify helpful components and suppress spurious ones. TIE \cite{tie} uses translation instead of projection, translating the image embedding away from a spurious vector.

All these methods intervene after the spurious feature has entered the representations through loss reweighting, retraining, or embedding adjustment. $A^2$ operates upstream at the input level, preventing the spurious correlation from entering the representation.

\section{Scope, Limitations, and Future Work}
$A^2$ helps when the label is associated with the dominant foreground object and the distractor or irrelevant aspects of the image can be spatially separated. CelebA is a good example of a failure case where default attention is not task relevant, requiring a small attention adapter (documented in Appendix Section \ref{supp:adapting_attention}). Our inverse scaling finding can be potentially utilized in future self supervised model training to improve attention map quality and mitigate the effects of inverse scaling. For example, the peak localization layer's attention map can serve as a distillation target for the final layer's attention. It would be interesting to see if doing this gives large SSL models some of the robustness benefits of $A^2$ in a single forward pass.

\section{Conclusion}

Self-supervised ViTs show inverse scaling of localization: smaller models are better than larger models at identifying foreground objects. Furthermore, this trend directly translates into $A^2$, with performance decreasing as the size of the model used to select crops increases. By preventing distractors from entering the model through hard crops, $A^2$ is competitive with loss-level baselines and beats end-to-end attention training (iFAM) under stronger distribution shifts while being simple and interpretable. 

\section*{Acknowledgements}

SR and CV are partially supported by the National Science Foundation's Center for Smart Streetscapes (CS3), the National Science Foundation's AI Institute for Artificial and Natural Intelligence, the NSF ACED Award \#2435757, and the NSF RI Award \#2403016.

\bibliographystyle{plainnat}
\bibliography{main}

\begin{thebibliography}{40}
\providecommand{\natexlab}[1]{#1}
\providecommand{\url}[1]{\texttt{#1}}
\expandafter\ifx\csname urlstyle\endcsname\relax
  \providecommand{\doi}[1]{doi: #1}\else
  \providecommand{\doi}{doi: \begingroup \urlstyle{rm}\Url}\fi

\bibitem[{Adila} et~al.(2023){Adila}, {Shin}, {Cai}, and {Sala}]{roboshot}
Dyah {Adila}, Changho {Shin}, Linrong {Cai}, and Frederic {Sala}.
\newblock {Zero-Shot Robustification of Zero-Shot Models}.
\newblock \emph{arXiv e-prints}, art. arXiv:2309.04344, September 2023.
\newblock \doi{10.48550/arXiv.2309.04344}.

\bibitem[Aniraj et~al.(2026)Aniraj, Dantas, Ienco, and Marcos]{ifam}
Ananthu Aniraj, Cassio~F. Dantas, Dino Ienco, and Diego Marcos.
\newblock Two-stage vision transformers and hard masking offer robust object
  representations.
\newblock In \emph{International Conference on Pattern Recognition (ICPR)},
  2026.

\bibitem[Bousselham et~al.(2023)Bousselham, Petersen, Ferrari, and Kuehne]{gem}
Walid Bousselham, Felix Petersen, Vittorio Ferrari, and Hilde Kuehne.
\newblock Grounding everything: Emerging localization properties in
  vision-language transformers.
\newblock \emph{arXiv preprint arXiv:2312.00878}, 2023.

\bibitem[{Carion} et~al.(2025){Carion}, {Gustafson}, {Hu}, {Debnath}, {Hu},
  {Suris}, {Ryali}, {Vasudev Alwala}, {Khedr}, {Huang}, {Lei}, {Ma}, {Guo},
  {Kalla}, {Marks}, {Greer}, {Wang}, {Sun}, {R{\"a}dle}, {Afouras}, {Mavroudi},
  {Xu}, {Wu}, {Zhou}, {Momeni}, {Hazra}, {Ding}, {Vaze}, {Porcher}, {Li}, {Li},
  {Kamath}, {Cheng}, {Doll{\'a}r}, {Ravi}, {Saenko}, {Zhang}, and
  {Feichtenhofer}]{sam3}
Nicolas {Carion}, Laura {Gustafson}, Yuan-Ting {Hu}, Shoubhik {Debnath},
  Ronghang {Hu}, Didac {Suris}, Chaitanya {Ryali}, Kalyan {Vasudev Alwala},
  Haitham {Khedr}, Andrew {Huang}, Jie {Lei}, Tengyu {Ma}, Baishan {Guo}, Arpit
  {Kalla}, Markus {Marks}, Joseph {Greer}, Meng {Wang}, Peize {Sun}, Roman
  {R{\"a}dle}, Triantafyllos {Afouras}, Effrosyni {Mavroudi}, Katherine {Xu},
  Tsung-Han {Wu}, Yu~{Zhou}, Liliane {Momeni}, Rishi {Hazra}, Shuangrui {Ding},
  Sagar {Vaze}, Francois {Porcher}, Feng {Li}, Siyuan {Li}, Aishwarya {Kamath},
  Ho~Kei {Cheng}, Piotr {Doll{\'a}r}, Nikhila {Ravi}, Kate {Saenko}, Pengchuan
  {Zhang}, and Christoph {Feichtenhofer}.
\newblock {SAM 3: Segment Anything with Concepts}.
\newblock \emph{arXiv e-prints}, art. arXiv:2511.16719, November 2025.
\newblock \doi{10.48550/arXiv.2511.16719}.

\bibitem[{Caron} et~al.(2021){Caron}, {Touvron}, {Misra}, {J{\'e}gou},
  {Mairal}, {Bojanowski}, and {Joulin}]{dinov1}
Mathilde {Caron}, Hugo {Touvron}, Ishan {Misra}, Herv{\'e} {J{\'e}gou}, Julien
  {Mairal}, Piotr {Bojanowski}, and Armand {Joulin}.
\newblock {Emerging Properties in Self-Supervised Vision Transformers}.
\newblock \emph{arXiv e-prints}, art. arXiv:2104.14294, April 2021.
\newblock \doi{10.48550/arXiv.2104.14294}.

\bibitem[{Cherti} et~al.(2022){Cherti}, {Beaumont}, {Wightman}, {Wortsman},
  {Ilharco}, {Gordon}, {Schuhmann}, {Schmidt}, and {Jitsev}]{openclip}
Mehdi {Cherti}, Romain {Beaumont}, Ross {Wightman}, Mitchell {Wortsman},
  Gabriel {Ilharco}, Cade {Gordon}, Christoph {Schuhmann}, Ludwig {Schmidt},
  and Jenia {Jitsev}.
\newblock {Reproducible scaling laws for contrastive language-image learning}.
\newblock \emph{arXiv e-prints}, art. arXiv:2212.07143, December 2022.
\newblock \doi{10.48550/arXiv.2212.07143}.

\bibitem[{Chevalley} et~al.(2022){Chevalley}, {Bunne}, {Krause}, and
  {Bauer}]{causIRL}
Mathieu {Chevalley}, Charlotte {Bunne}, Andreas {Krause}, and Stefan {Bauer}.
\newblock {Invariant Causal Mechanisms through Distribution Matching}.
\newblock \emph{arXiv e-prints}, art. arXiv:2206.11646, June 2022.
\newblock \doi{10.48550/arXiv.2206.11646}.

\bibitem[Cordonnier et~al.(2021)Cordonnier, Mahendran, Dosovitskiy,
  Weissenborn, Uszkoreit, and Unterthiner]{dps}
Jean-Baptiste Cordonnier, Aravindh Mahendran, Alexey Dosovitskiy, Dirk
  Weissenborn, Jakob Uszkoreit, and Thomas Unterthiner.
\newblock Differentiable patch selection for image recognition.
\newblock In \emph{Proceedings of the IEEE/CVF Conference on Computer Vision
  and Pattern Recognition}, pages 2351--2360, 2021.

\bibitem[Darcet et~al.(2024)Darcet, Oquab, Mairal, and Bojanowski]{vitregister}
Timoth{\'e}e Darcet, Maxime Oquab, Julien Mairal, and Piotr Bojanowski.
\newblock Vision transformers need registers.
\newblock In \emph{The Twelfth International Conference on Learning
  Representations}, 2024.
\newblock URL \url{https://openreview.net/forum?id=2dnO3LLiJ1}.

\bibitem[Dehghani et~al.(2023)Dehghani, Djolonga, Mustafa, Padlewski, Heek,
  Gilmer, Steiner, Caron, Geirhos, Alabdulmohsin, Jenatton, Beyer, Tschannen,
  Arnab, Wang, Riquelme~Ruiz, Minderer, Puigcerver, Evci, Kumar, Steenkiste,
  Elsayed, Mahendran, Yu, Oliver, Huot, Bastings, Collier, Gritsenko, Birodkar,
  Vasconcelos, Tay, Mensink, Kolesnikov, Pavetic, Tran, Kipf, Lucic, Zhai,
  Keysers, Harmsen, and Houlsby]{scalingTo22B}
Mostafa Dehghani, Josip Djolonga, Basil Mustafa, Piotr Padlewski, Jonathan
  Heek, Justin Gilmer, Andreas~Peter Steiner, Mathilde Caron, Robert Geirhos,
  Ibrahim Alabdulmohsin, Rodolphe Jenatton, Lucas Beyer, Michael Tschannen,
  Anurag Arnab, Xiao Wang, Carlos Riquelme~Ruiz, Matthias Minderer, Joan
  Puigcerver, Utku Evci, Manoj Kumar, Sjoerd~Van Steenkiste, Gamaleldin~Fathy
  Elsayed, Aravindh Mahendran, Fisher Yu, Avital Oliver, Fantine Huot, Jasmijn
  Bastings, Mark Collier, Alexey~A. Gritsenko, Vighnesh Birodkar,
  Cristina~Nader Vasconcelos, Yi~Tay, Thomas Mensink, Alexander Kolesnikov,
  Filip Pavetic, Dustin Tran, Thomas Kipf, Mario Lucic, Xiaohua Zhai, Daniel
  Keysers, Jeremiah~J. Harmsen, and Neil Houlsby.
\newblock Scaling vision transformers to 22 billion parameters.
\newblock In Andreas Krause, Emma Brunskill, Kyunghyun Cho, Barbara Engelhardt,
  Sivan Sabato, and Jonathan Scarlett, editors, \emph{Proceedings of the 40th
  International Conference on Machine Learning}, volume 202 of
  \emph{Proceedings of Machine Learning Research}, pages 7480--7512. PMLR,
  23--29 Jul 2023.
\newblock URL \url{https://proceedings.mlr.press/v202/dehghani23a.html}.

\bibitem[Deng et~al.(2009)Deng, Dong, Socher, Li, Li, and Fei-Fei]{imagenet}
Jia Deng, Wei Dong, Richard Socher, Li-Jia Li, Kai Li, and Li~Fei-Fei.
\newblock Imagenet: A large-scale hierarchical image database.
\newblock In \emph{2009 IEEE Conference on Computer Vision and Pattern
  Recognition}, pages 248--255, 2009.
\newblock \doi{10.1109/CVPR.2009.5206848}.

\bibitem[{Dosovitskiy} et~al.(2020){Dosovitskiy}, {Beyer}, {Kolesnikov},
  {Weissenborn}, {Zhai}, {Unterthiner}, {Dehghani}, {Minderer}, {Heigold},
  {Gelly}, {Uszkoreit}, and {Houlsby}]{vit}
Alexey {Dosovitskiy}, Lucas {Beyer}, Alexander {Kolesnikov}, Dirk
  {Weissenborn}, Xiaohua {Zhai}, Thomas {Unterthiner}, Mostafa {Dehghani},
  Matthias {Minderer}, Georg {Heigold}, Sylvain {Gelly}, Jakob {Uszkoreit}, and
  Neil {Houlsby}.
\newblock {An Image is Worth 16x16 Words: Transformers for Image Recognition at
  Scale}.
\newblock \emph{arXiv e-prints}, art. arXiv:2010.11929, October 2020.
\newblock \doi{10.48550/arXiv.2010.11929}.

\bibitem[Fuller et~al.(2025)Fuller, Yassin, Wen, Kyrollos, Ibrahim, Green, and
  Shelhamer]{fuller2025}
Anthony Fuller, Yousef Yassin, Junfeng Wen, Daniel~G. Kyrollos, Tarek Ibrahim,
  James~R. Green, and Evan Shelhamer.
\newblock Lookwhere? efficient visual recognition by learning where to look and
  what to see from self-supervision, 2025.
\newblock URL \url{https://arxiv.org/abs/2505.18051}.

\bibitem[{Geirhos} et~al.(2020){Geirhos}, {Jacobsen}, {Michaelis}, {Zemel},
  {Brendel}, {Bethge}, and {Wichmann}]{shortcutLearning}
Robert {Geirhos}, J{\"o}rn-Henrik {Jacobsen}, Claudio {Michaelis}, Richard
  {Zemel}, Wieland {Brendel}, Matthias {Bethge}, and Felix~A. {Wichmann}.
\newblock {Shortcut Learning in Deep Neural Networks}.
\newblock \emph{arXiv e-prints}, art. arXiv:2004.07780, April 2020.
\newblock \doi{10.48550/arXiv.2004.07780}.

\bibitem[He et~al.(2022)He, Chen, Xie, Li, Doll{\'a}r, and Girshick]{mae}
Kaiming He, Xinlei Chen, Saining Xie, Yanghao Li, Piotr Doll{\'a}r, and Ross
  Girshick.
\newblock Masked autoencoders are scalable vision learners.
\newblock In \emph{Proceedings of the IEEE/CVF conference on computer vision
  and pattern recognition}, pages 16000--16009, 2022.

\bibitem[{Khosla} et~al.(2020){Khosla}, {Teterwak}, {Wang}, {Sarna}, {Tian},
  {Isola}, {Maschinot}, {Liu}, and {Krishnan}]{supervisedContrastiveLearning}
Prannay {Khosla}, Piotr {Teterwak}, Chen {Wang}, Aaron {Sarna}, Yonglong
  {Tian}, Phillip {Isola}, Aaron {Maschinot}, Ce~{Liu}, and Dilip {Krishnan}.
\newblock {Supervised Contrastive Learning}.
\newblock \emph{arXiv e-prints}, art. arXiv:2004.11362, April 2020.
\newblock \doi{10.48550/arXiv.2004.11362}.

\bibitem[Kirichenko et~al.(2022)Kirichenko, Izmailov, and Wilson]{dfr}
Polina Kirichenko, Pavel Izmailov, and Andrew~Gordon Wilson.
\newblock Last layer re-training is sufficient for robustness to spurious
  correlations.
\newblock \emph{arXiv preprint arXiv:2204.02937}, 2022.

\bibitem[Lappe and Giese(2025)]{lappe}
Alexander Lappe and Martin Giese.
\newblock Register and [cls] tokens induce a decoupling of local and global
  features in large vits.
\newblock In D.~Belgrave, C.~Zhang, H.~Lin, R.~Pascanu, P.~Koniusz,
  M.~Ghassemi, and N.~Chen, editors, \emph{Advances in Neural Information
  Processing Systems}, volume~38, pages 1009--1029. Curran Associates, Inc.,
  2025.
\newblock URL
  \url{https://proceedings.neurips.cc/paper_files/paper/2025/file/01bcbd34b02f3da9700a3ddd0480c156-Paper-Conference.pdf}.

\bibitem[{Liang} and {Zou}(2022)]{MetaShift}
Weixin {Liang} and James {Zou}.
\newblock {MetaShift: A Dataset of Datasets for Evaluating Contextual
  Distribution Shifts and Training Conflicts}.
\newblock \emph{arXiv e-prints}, art. arXiv:2202.06523, February 2022.
\newblock \doi{10.48550/arXiv.2202.06523}.

\bibitem[Liu et~al.(2021)Liu, Haghgoo, Chen, Raghunathan, Koh, Sagawa, Liang,
  and Finn]{jtt}
Evan~Z Liu, Behzad Haghgoo, Annie~S Chen, Aditi Raghunathan, Pang~Wei Koh,
  Shiori Sagawa, Percy Liang, and Chelsea Finn.
\newblock Just train twice: Improving group robustness without training group
  information.
\newblock In Marina Meila and Tong Zhang, editors, \emph{Proceedings of the
  38th International Conference on Machine Learning}, volume 139 of
  \emph{Proceedings of Machine Learning Research}, pages 6781--6792. PMLR,
  18--24 Jul 2021.
\newblock URL \url{https://proceedings.mlr.press/v139/liu21f.html}.

\bibitem[Liu et~al.(2015)Liu, Luo, Wang, and Tang]{celebA}
Ziwei Liu, Ping Luo, Xiaogang Wang, and Xiaoou Tang.
\newblock Deep learning face attributes in the wild.
\newblock In \emph{Proceedings of International Conference on Computer Vision
  (ICCV)}, December 2015.

\bibitem[Lu et~al.(2025)Lu, Pan, and Wang]{ttr}
Shenyu Lu, Zhaoying Pan, and Xiaoqian Wang.
\newblock Think twice: Test-time reasoning for robust clip zero-shot
  classification.
\newblock In \emph{Proceedings of the IEEE/CVF International Conference on
  Computer Vision (ICCV)}, pages 2919--2929, October 2025.

\bibitem[Lynch et~al.(2023)Lynch, Dovonon, Kaddour, and
  Silva]{lynch2023spawrious}
Aengus Lynch, Gbètondji J-S Dovonon, Jean Kaddour, and Ricardo Silva.
\newblock Spawrious: A benchmark for fine control of spurious correlation
  biases, 2023.

\bibitem[Oquab et~al.(2023)Oquab, Darcet, Moutakanni, Vo, Szafraniec, Khalidov,
  Fernandez, Haziza, Massa, El-Nouby, et~al.]{dinov2}
Maxime Oquab, Timoth{\'e}e Darcet, Th{\'e}o Moutakanni, Huy Vo, Marc
  Szafraniec, Vasil Khalidov, Pierre Fernandez, Daniel Haziza, Francisco Massa,
  Alaaeldin El-Nouby, et~al.
\newblock Dinov2: Learning robust visual features without supervision.
\newblock \emph{arXiv preprint arXiv:2304.07193}, 2023.

\bibitem[{Radford} et~al.(2021){Radford}, {Kim}, {Hallacy}, {Ramesh}, {Goh},
  {Agarwal}, {Sastry}, {Askell}, {Mishkin}, {Clark}, {Krueger}, and
  {Sutskever}]{clip}
Alec {Radford}, Jong~Wook {Kim}, Chris {Hallacy}, Aditya {Ramesh}, Gabriel
  {Goh}, Sandhini {Agarwal}, Girish {Sastry}, Amanda {Askell}, Pamela
  {Mishkin}, Jack {Clark}, Gretchen {Krueger}, and Ilya {Sutskever}.
\newblock {Learning Transferable Visual Models From Natural Language
  Supervision}.
\newblock \emph{arXiv e-prints}, art. arXiv:2103.00020, February 2021.
\newblock \doi{10.48550/arXiv.2103.00020}.

\bibitem[Ravi et~al.(2024)Ravi, Gabeur, Hu, Hu, Ryali, Ma, Khedr, R{\"a}dle,
  Rolland, Gustafson, Mintun, Pan, Alwala, Carion, Wu, Girshick, Doll{\'a}r,
  and Feichtenhofer]{sam2}
Nikhila Ravi, Valentin Gabeur, Yuan-Ting Hu, Ronghang Hu, Chaitanya Ryali,
  Tengyu Ma, Haitham Khedr, Roman R{\"a}dle, Chloe Rolland, Laura Gustafson,
  Eric Mintun, Junting Pan, Kalyan~Vasudev Alwala, Nicolas Carion, Chao-Yuan
  Wu, Ross Girshick, Piotr Doll{\'a}r, and Christoph Feichtenhofer.
\newblock Sam 2: Segment anything in images and videos.
\newblock \emph{arXiv preprint arXiv:2408.00714}, 2024.
\newblock URL \url{https://arxiv.org/abs/2408.00714}.

\bibitem[{Sagawa} et~al.(2019){Sagawa}, {Koh}, {Hashimoto}, and
  {Liang}]{groupDRO}
Shiori {Sagawa}, Pang~Wei {Koh}, Tatsunori~B. {Hashimoto}, and Percy {Liang}.
\newblock {Distributionally Robust Neural Networks for Group Shifts: On the
  Importance of Regularization for Worst-Case Generalization}.
\newblock \emph{arXiv e-prints}, art. arXiv:1911.08731, November 2019.
\newblock \doi{10.48550/arXiv.1911.08731}.

\bibitem[Selvaraju et~al.(2017)Selvaraju, Cogswell, Das, Vedantam, Parikh, and
  Batra]{gradcam}
Ramprasaath~R. Selvaraju, Michael Cogswell, Abhishek Das, Ramakrishna Vedantam,
  Devi Parikh, and Dhruv Batra.
\newblock Grad-cam: Visual explanations from deep networks via gradient-based
  localization.
\newblock In \emph{Proceedings of the IEEE International Conference on Computer
  Vision (ICCV)}, Oct 2017.

\bibitem[{Shi} et~al.(2024){Shi}, {Wu}, {Mao}, {Wang}, and
  {Darrell}]{scalingScales}
Baifeng {Shi}, Ziyang {Wu}, Maolin {Mao}, Xin {Wang}, and Trevor {Darrell}.
\newblock {When Do We Not Need Larger Vision Models?}
\newblock \emph{arXiv e-prints}, art. arXiv:2403.13043, March 2024.
\newblock \doi{10.48550/arXiv.2403.13043}.

\bibitem[{Sim{\'e}oni} et~al.(2021){Sim{\'e}oni}, {Puy}, {Vo}, {Roburin},
  {Gidaris}, {Bursuc}, {P{\'e}rez}, {Marlet}, and {Ponce}]{LOST}
Oriane {Sim{\'e}oni}, Gilles {Puy}, Huy~V. {Vo}, Simon {Roburin}, Spyros
  {Gidaris}, Andrei {Bursuc}, Patrick {P{\'e}rez}, Renaud {Marlet}, and Jean
  {Ponce}.
\newblock {Localizing Objects with Self-Supervised Transformers and no Labels}.
\newblock \emph{arXiv e-prints}, art. arXiv:2109.14279, September 2021.
\newblock \doi{10.48550/arXiv.2109.14279}.

\bibitem[{Sim{\'e}oni} et~al.(2025){Sim{\'e}oni}, {Vo}, {Seitzer},
  {Baldassarre}, {Oquab}, {Jose}, {Khalidov}, {Szafraniec}, {Yi},
  {Ramamonjisoa}, {Massa}, {Haziza}, {Wehrstedt}, {Wang}, {Darcet},
  {Moutakanni}, {Sentana}, {Roberts}, {Vedaldi}, {Tolan}, {Brandt}, {Couprie},
  {Mairal}, {J{\'e}gou}, {Labatut}, and {Bojanowski}]{dinov3}
Oriane {Sim{\'e}oni}, Huy~V. {Vo}, Maximilian {Seitzer}, Federico
  {Baldassarre}, Maxime {Oquab}, Cijo {Jose}, Vasil {Khalidov}, Marc
  {Szafraniec}, Seungeun {Yi}, Micha{\"e}l {Ramamonjisoa}, Francisco {Massa},
  Daniel {Haziza}, Luca {Wehrstedt}, Jianyuan {Wang}, Timoth{\'e}e {Darcet},
  Th{\'e}o {Moutakanni}, Leonel {Sentana}, Claire {Roberts}, Andrea {Vedaldi},
  Jamie {Tolan}, John {Brandt}, Camille {Couprie}, Julien {Mairal}, Herv{\'e}
  {J{\'e}gou}, Patrick {Labatut}, and Piotr {Bojanowski}.
\newblock {DINOv3}.
\newblock \emph{arXiv e-prints}, art. arXiv:2508.10104, August 2025.
\newblock \doi{10.48550/arXiv.2508.10104}.

\bibitem[{Sun} and {Saenko}(2016)]{coral}
Baochen {Sun} and Kate {Saenko}.
\newblock {Deep CORAL: Correlation Alignment for Deep Domain Adaptation}.
\newblock \emph{arXiv e-prints}, art. arXiv:1607.01719, July 2016.
\newblock \doi{10.48550/arXiv.1607.01719}.

\bibitem[Sun et~al.(2017)Sun, Shrivastava, Singh, and
  Gupta]{unreasonableEffectivenessData}
Chen Sun, Abhinav Shrivastava, Saurabh Singh, and Abhinav Gupta.
\newblock { Revisiting Unreasonable Effectiveness of Data in Deep Learning Era
  }.
\newblock In \emph{2017 IEEE International Conference on Computer Vision
  (ICCV)}, pages 843--852, Los Alamitos, CA, USA, October 2017. IEEE Computer
  Society.
\newblock \doi{10.1109/ICCV.2017.97}.
\newblock URL \url{https://doi.ieeecomputersociety.org/10.1109/ICCV.2017.97}.

\bibitem[{Vaswani} et~al.(2017){Vaswani}, {Shazeer}, {Parmar}, {Uszkoreit},
  {Jones}, {Gomez}, {Kaiser}, and {Polosukhin}]{attentionAllYouNeed}
Ashish {Vaswani}, Noam {Shazeer}, Niki {Parmar}, Jakob {Uszkoreit}, Llion
  {Jones}, Aidan~N. {Gomez}, Lukasz {Kaiser}, and Illia {Polosukhin}.
\newblock {Attention Is All You Need}.
\newblock \emph{arXiv e-prints}, art. arXiv:1706.03762, June 2017.
\newblock \doi{10.48550/arXiv.1706.03762}.

\bibitem[{Wang} et~al.(2022){Wang}, {Shen}, {Yuan}, {Du}, {Li}, {Hu},
  {Crowley}, and {Vaufreydaz}]{tokencut}
Yangtao {Wang}, Xi~{Shen}, Yuan {Yuan}, Yuming {Du}, Maomao {Li}, Shell~Xu
  {Hu}, James~L {Crowley}, and Dominique {Vaufreydaz}.
\newblock {TokenCut: Segmenting Objects in Images and Videos with
  Self-supervised Transformer and Normalized Cut}.
\newblock \emph{arXiv e-prints}, art. arXiv:2209.00383, September 2022.
\newblock \doi{10.48550/arXiv.2209.00383}.

\bibitem[{Xu} et~al.(2015){Xu}, {Ba}, {Kiros}, {Cho}, {Courville},
  {Salakhutdinov}, {Zemel}, and {Bengio}]{hardAttention}
Kelvin {Xu}, Jimmy {Ba}, Ryan {Kiros}, Kyunghyun {Cho}, Aaron {Courville},
  Ruslan {Salakhutdinov}, Richard {Zemel}, and Yoshua {Bengio}.
\newblock {Show, Attend and Tell: Neural Image Caption Generation with Visual
  Attention}.
\newblock \emph{arXiv e-prints}, art. arXiv:1502.03044, February 2015.
\newblock \doi{10.48550/arXiv.1502.03044}.

\bibitem[{Yang} et~al.(2023){Yang}, {Nushi}, {Palangi}, and
  {Mirzasoleiman}]{tie}
Yu~{Yang}, Besmira {Nushi}, Hamid {Palangi}, and Baharan {Mirzasoleiman}.
\newblock {Mitigating Spurious Correlations in Multi-modal Models during
  Fine-tuning}.
\newblock \emph{arXiv e-prints}, art. arXiv:2304.03916, April 2023.
\newblock \doi{10.48550/arXiv.2304.03916}.

\bibitem[Zhai et~al.(2022)Zhai, Kolesnikov, Houlsby, and Beyer]{scalingVITs}
Xiaohua Zhai, Alexander Kolesnikov, Neil Houlsby, and Lucas Beyer.
\newblock Scaling vision transformers.
\newblock In \emph{Proceedings of the IEEE/CVF Conference on Computer Vision
  and Pattern Recognition (CVPR)}, pages 12104--12113, June 2022.

\bibitem[{Zhang} et~al.(2022){Zhang}, {Sohoni}, {Zhang}, {Finn}, and
  {R{\'e}}]{cnc}
Michael {Zhang}, Nimit~S. {Sohoni}, Hongyang~R. {Zhang}, Chelsea {Finn}, and
  Christopher {R{\'e}}.
\newblock {Correct-N-Contrast: A Contrastive Approach for Improving Robustness
  to Spurious Correlations}.
\newblock \emph{arXiv e-prints}, art. arXiv:2203.01517, March 2022.
\newblock \doi{10.48550/arXiv.2203.01517}.

\bibitem[Zhou et~al.(2022)Zhou, Wei, Wang, Shen, Xie, Yuille, and Kong]{ibot}
Jinghao Zhou, Chen Wei, Huiyu Wang, Wei Shen, Cihang Xie, Alan Yuille, and Tao
  Kong.
\newblock ibot: Image bert pre-training with online tokenizer.
\newblock \emph{International Conference on Learning Representations (ICLR)},
  2022.

\end{thebibliography}

\appendix

\clearpage
\raggedbottom
\makeatletter
\setlength{\@fptop}{0pt}\setlength{\@fpsep}{\floatsep}\setlength{\@fpbot}{0pt plus 1fil}
\makeatother
\section{Attention Localization and Hit Rate across model families}

\definecolor{trendcell}{RGB}{220, 232, 245}    %
\definecolor{boundarycell}{RGB}{250, 232, 215}  %

\begin{table}[H]
\centering
\small
\setlength{\fboxsep}{2pt}
\caption{Attention localization vs.\ ground-truth bounding boxes across six pretraining families. Mass is the fraction of attention mass inside the ground truth bounding boxes; Hit is whether the $\arg\max$ of attention lies inside any ground truth box. Bold = best per family per column. \colorbox{trendcell}{Blue rows} mark the smallest model under self-supervised pretraining, where the smallest ViT beats the largest on every column. \colorbox{boundarycell}{Peach row} marks the smallest model under contrastive image-text pretraining, where the trend breaks (smallest is not the best).}
\label{tab:attention_iou_all}
\begin{tabular}{@{}lccccc@{}}
\toprule
& & \multicolumn{2}{c}{\textbf{Waterbirds}} & \multicolumn{2}{c}{\textbf{ImageNet val}} \\
\cmidrule(lr){3-4} \cmidrule(lr){5-6}
\textbf{Model} & \textbf{Params} & \textbf{Mass} $\uparrow$ & \textbf{Hit} $\uparrow$ & \textbf{Mass} $\uparrow$ & \textbf{Hit} $\uparrow$ \\
\midrule
\multicolumn{6}{l}{\textbf{DINOv1 (self-distillation, original)}} \\
\rowcolor{trendcell}
\quad ViT-S/16 &  21M  & $\mathbf{0.689}$ & $\mathbf{0.939}$ & $\mathbf{0.681}$ & $\mathbf{0.849}$ \\
\quad ViT-B/16 &  86M  & $0.649$ & $0.934$ & $0.659$ & $0.829$ \\
\addlinespace[3pt]
\multicolumn{6}{l}{\textbf{DINOv2 (self-distillation)}} \\
\rowcolor{trendcell}
\quad ViT-S    &  21M  & $\mathbf{0.820}$ & $\mathbf{0.947}$ & $\mathbf{0.745}$ & $\mathbf{0.834}$ \\
\quad ViT-B    &  86M  & $0.764$ & $0.875$ & $0.726$ & $0.818$ \\
\quad ViT-L    & 300M  & $0.650$ & $0.823$ & $0.678$ & $0.767$ \\
\quad ViT-G    & 1.1B  & $0.608$ & $0.819$ & $0.676$ & $0.765$ \\
\addlinespace[3pt]
\multicolumn{6}{l}{\textbf{DINOv3 (self-distillation)}} \\
\rowcolor{trendcell}
\quad ViT-S/16   &   21M  & $\mathbf{0.767}$ & $\mathbf{0.913}$ & $0.723$ & $0.821$ \\
\quad ViT-S+/16  &   29M  & $0.758$ & $0.900$ & $0.729$ & $0.821$ \\
\quad ViT-B/16   &   86M  & $0.735$ & $0.877$ & $\mathbf{0.740}$ & $\mathbf{0.840}$ \\
\quad ViT-L/16   &  300M  & $0.690$ & $0.888$ & $0.705$ & $0.824$ \\
\quad ViT-H+/16  &  840M  & $0.662$ & $0.870$ & $0.705$ & $0.808$ \\
\quad ViT-7B/16  & 6.7B   & $0.577$ & $0.798$ & $0.649$ & $0.756$ \\
\addlinespace[3pt]
\multicolumn{6}{l}{\textbf{MAE (reconstruction)}} \\
\rowcolor{trendcell}
\quad ViT-B/16 &  86M  & $\mathbf{0.474}$ & $\mathbf{0.879}$ & $\mathbf{0.551}$ & $\mathbf{0.658}$ \\
\quad ViT-L/16 & 307M  & $0.376$ & $0.370$ & $0.507$ & $0.499$ \\
\quad ViT-H/14 & 632M  & $0.377$ & $0.451$ & $0.503$ & $0.554$ \\
\addlinespace[3pt]
\multicolumn{6}{l}{\textbf{iBOT (masked self-distillation)}} \\
\rowcolor{trendcell}
\quad ViT-S/16 &  21M  & $\mathbf{0.696}$ & $0.909$ & $\mathbf{0.675}$ & $\mathbf{0.824}$ \\
\quad ViT-B/16 &  85M  & $0.639$ & $\mathbf{0.949}$ & $0.635$ & $0.809$ \\
\quad ViT-L/16 & 307M  & $0.557$ & $0.902$ & $0.583$ & $0.750$ \\
\addlinespace[3pt]
\multicolumn{6}{l}{\textbf{OpenCLIP (contrastive image-text)}} \\
\rowcolor{boundarycell}
\quad ViT-B/16 &  86M  & $0.480$ & $0.166$ & $0.525$ & $0.291$ \\
\quad ViT-L/14 & 304M  & $\mathbf{0.561}$ & $\mathbf{0.815}$ & $\mathbf{0.567}$ & $0.543$ \\
\quad ViT-H/14 & 632M  & $0.526$ & $0.431$ & $0.554$ & $0.457$ \\
\quad ViT-g/14 & 1.0B  & $0.522$ & $0.585$ & $0.556$ & $\mathbf{0.555}$ \\
\bottomrule
\end{tabular}
\end{table}

\begin{figure}[H]
\centering
\definecolor{famDinoOne}{HTML}{7C4607}%
\definecolor{famDinoThree}{HTML}{005586}%
\definecolor{famIbot}{HTML}{0E8E4F}%
\definecolor{famMae}{HTML}{7E5CB7}%
\begin{tikzpicture}[font=\fontsize{8}{9.5}\selectfont]
\begin{groupplot}[
  group style={group size=2 by 1, horizontal sep=0.40in},
  scale only axis, width=2.16in, height=1.05in,
  xmode=log, xmin=15, xmax=9500,
  xtick={21,86,300,1100,6716}, xticklabels={21M,86M,300M,1.1B,6.7B},
  axis lines*=left, axis line style={line width=0.5pt, -},
  tick align=outside, major tick length=2pt, minor tick length=0pt,
  tick style={line width=0.5pt, black},
  tick label style={font=\fontsize{8}{9.5}\selectfont, /pgf/number format/assume math mode=true},
  label style={font=\fontsize{8}{9.5}\selectfont},
  xlabel={Attention model size (parameters)}, xlabel style={yshift=2pt},
  title style={font=\fontsize{8}{9.5}\selectfont\bfseries, at={(0,1)}, anchor=south west, inner sep=0pt, yshift=4pt},
  clip=false,
]
\nextgroupplot[title={Waterbirds}, ylabel={Hit rate (\%)}, ylabel style={yshift=-2pt},
  ymin=8, ymax=102, ytick={20,40,60,80,100},
  legend to name=loclegend-hit,
  legend style={draw=none, fill=none, legend columns=6, font=\fontsize{8}{9.5}\selectfont,
    /tikz/every even column/.append style={column sep=9pt}}]
\addlegendimage{color=accent, line width=1.2pt, mark=*, mark size=2.4pt, mark options={fill=accent, draw=accent, line width=0pt}}\addlegendentry{\textbf{DINOv2}}
\addlegendimage{color=famDinoOne, line width=0.8pt, mark=square*, mark size=1.9pt, mark options={fill=famDinoOne!35, draw=famDinoOne, line width=0.6pt}}\addlegendentry{DINOv1}
\addlegendimage{color=famDinoThree, line width=0.8pt, mark=diamond*, mark size=1.9pt, mark options={fill=famDinoThree!35, draw=famDinoThree, line width=0.6pt}}\addlegendentry{DINOv3}
\addlegendimage{color=famIbot, line width=0.8pt, mark=triangle*, mark size=1.9pt, mark options={fill=famIbot!35, draw=famIbot, line width=0.6pt}}\addlegendentry{iBOT}
\addlegendimage{color=famMae, line width=0.8pt, mark=*, mark size=1.9pt, mark options={fill=famMae!35, draw=famMae, line width=0.6pt}}\addlegendentry{MAE}
\addlegendimage{color=refgray, line width=0.8pt, mark=pentagon*, mark size=1.9pt, mark options={fill=refgray!35, draw=refgray, line width=0.6pt}}\addlegendentry{OpenCLIP (image--text)}
\addplot[color=refgray, line width=0.8pt, mark=pentagon*, mark size=1.9pt, mark options={fill=refgray!35, draw=refgray, line width=0.6pt}] table[x=params, y=wb_hit, col sep=comma] {Figures/attention_scaling/openclip.csv};
\addplot[color=famDinoOne, line width=0.8pt, mark=square*, mark size=1.9pt, mark options={fill=famDinoOne!35, draw=famDinoOne, line width=0.6pt}] table[x=params, y=wb_hit, col sep=comma] {Figures/attention_scaling/dinov1.csv};
\addplot[color=famDinoThree, line width=0.8pt, mark=diamond*, mark size=1.9pt, mark options={fill=famDinoThree!35, draw=famDinoThree, line width=0.6pt}] table[x=params, y=wb_hit, col sep=comma] {Figures/attention_scaling/dinov3.csv};
\addplot[color=famIbot, line width=0.8pt, mark=triangle*, mark size=1.9pt, mark options={fill=famIbot!35, draw=famIbot, line width=0.6pt}] table[x=params, y=wb_hit, col sep=comma] {Figures/attention_scaling/ibot.csv};
\addplot[color=famMae, line width=0.8pt, mark=*, mark size=1.9pt, mark options={fill=famMae!35, draw=famMae, line width=0.6pt}] table[x=params, y=wb_hit, col sep=comma] {Figures/attention_scaling/mae.csv};
\addplot[color=accent, line width=1.2pt, mark=*, mark size=2.4pt, mark options={fill=accent, draw=accent, line width=0pt}] table[x=params, y=wb_hit, col sep=comma] {Figures/attention_scaling/dinov2.csv};
\nextgroupplot[title={ImageNet val}, ymin=24, ymax=92, ytick={30,50,70,90}]
\addplot[color=refgray, line width=0.8pt, mark=pentagon*, mark size=1.9pt, mark options={fill=refgray!35, draw=refgray, line width=0.6pt}] table[x=params, y=in_hit, col sep=comma] {Figures/attention_scaling/openclip.csv};
\addplot[color=famDinoOne, line width=0.8pt, mark=square*, mark size=1.9pt, mark options={fill=famDinoOne!35, draw=famDinoOne, line width=0.6pt}] table[x=params, y=in_hit, col sep=comma] {Figures/attention_scaling/dinov1.csv};
\addplot[color=famDinoThree, line width=0.8pt, mark=diamond*, mark size=1.9pt, mark options={fill=famDinoThree!35, draw=famDinoThree, line width=0.6pt}] table[x=params, y=in_hit, col sep=comma] {Figures/attention_scaling/dinov3.csv};
\addplot[color=famIbot, line width=0.8pt, mark=triangle*, mark size=1.9pt, mark options={fill=famIbot!35, draw=famIbot, line width=0.6pt}] table[x=params, y=in_hit, col sep=comma] {Figures/attention_scaling/ibot.csv};
\addplot[color=famMae, line width=0.8pt, mark=*, mark size=1.9pt, mark options={fill=famMae!35, draw=famMae, line width=0.6pt}] table[x=params, y=in_hit, col sep=comma] {Figures/attention_scaling/mae.csv};
\addplot[color=accent, line width=1.2pt, mark=*, mark size=2.4pt, mark options={fill=accent, draw=accent, line width=0pt}] table[x=params, y=in_hit, col sep=comma] {Figures/attention_scaling/dinov2.csv};
\end{groupplot}
\node[anchor=south] at ([xshift=0.20in, yshift=15pt]group c1r1.north east) {\pgfplotslegendfromname{loclegend-hit}};
\end{tikzpicture}%
\caption{Attention hit rate vs.\ model size across six pretraining families, on Waterbirds (left) and ImageNet val (right). Hit rate is the fraction of images where the $\arg\max$ of the attention map falls inside the ground-truth bounding box. The DINOv2 and DINOv3 families show the cleanest inverse-scaling trend; MAE drops sharply from ViT-B/16 to ViT-L/16 on Waterbirds; OpenCLIP (gray) breaks the trend, with ViT-B/16 at $16.6\%$ on Waterbirds (well below the next size). Companion to Figure~\ref{fig:attention_mass_scaling}; see Table~\ref{tab:attention_iou_all} for precise values.}
\label{fig:attention_hit_scaling}
\end{figure}

\begin{figure}[H]
\centering
\definecolor{sizeS}{HTML}{0FA49D}%
\definecolor{sizeB}{HTML}{C65B2D}%
\definecolor{sizeL}{HTML}{943F69}%
\definecolor{sizeG}{HTML}{363FB4}%
\begin{tikzpicture}[font=\fontsize{8}{9.5}\selectfont]
\begin{groupplot}[
  group style={group size=2 by 1, horizontal sep=0.40in},
  scale only axis, width=2.18in, height=1.05in,
  xmin=-1.5, xmax=41.5, xtick={0,11,23,39}, xlabel={Block (0-indexed)}, xlabel style={yshift=2pt},
  axis lines*=left, axis line style={line width=0.5pt, -},
  tick align=outside, major tick length=2pt, minor tick length=0pt,
  tick style={line width=0.5pt, black},
  tick label style={font=\fontsize{8}{9.5}\selectfont, /pgf/number format/assume math mode=true},
  label style={font=\fontsize{8}{9.5}\selectfont},
  title style={font=\fontsize{8}{9.5}\selectfont\bfseries, at={(0,1)}, anchor=south west, inner sep=0pt, yshift=4pt},
  clip=false,
]
\nextgroupplot[title={Waterbirds}, ymin=12.8, ymax=112.7, ytick={20,40,60,80,100}, ylabel={Hit rate (\%)}, ylabel style={yshift=-2pt},
  legend to name=perlayer-hit-reg,
  legend style={draw=none, fill=none, legend columns=6, font=\fontsize{8}{9.5}\selectfont,
    /tikz/every even column/.append style={column sep=8pt}}]
\addlegendimage{color=sizeS, line width=1pt}\addlegendentry{ViT-S}
\addlegendimage{color=sizeB, line width=1pt}\addlegendentry{ViT-B}
\addlegendimage{color=sizeL, line width=1pt}\addlegendentry{ViT-L}
\addlegendimage{color=sizeG, line width=1pt}\addlegendentry{ViT-G}
\addlegendimage{only marks, mark=*, mark size=1.9pt, mark options={fill=white, draw=black, line width=0.7pt}}\addlegendentry{peak block}
\addlegendimage{only marks, mark=*, mark size=1.9pt, mark options={fill=black, draw=black, line width=0.7pt}}\addlegendentry{last block}
\addplot[color=sizeS, line width=1pt] table[x=block, y=hit, col sep=comma] {Figures/per_layer/reg_wb_vits.csv};
\addplot[only marks, mark=*, mark size=1.9pt, mark options={fill=sizeS, draw=sizeS, line width=0.7pt}] coordinates {(11,94.68)};
\addplot[color=sizeB, line width=1pt] table[x=block, y=hit, col sep=comma] {Figures/per_layer/reg_wb_vitb.csv};
\addplot[only marks, mark=*, mark size=1.9pt, mark options={fill=white, draw=sizeB, line width=0.7pt}] coordinates {(10,98.19)};
\addplot[only marks, mark=*, mark size=1.9pt, mark options={fill=sizeB, draw=sizeB, line width=0.7pt}] coordinates {(11,87.49)};
\addplot[color=sizeL, line width=1pt] table[x=block, y=hit, col sep=comma] {Figures/per_layer/reg_wb_vitl.csv};
\addplot[only marks, mark=*, mark size=1.9pt, mark options={fill=white, draw=sizeL, line width=0.7pt}] coordinates {(18,97.29)};
\addplot[only marks, mark=*, mark size=1.9pt, mark options={fill=sizeL, draw=sizeL, line width=0.7pt}] coordinates {(23,82.27)};
\addplot[color=sizeG, line width=1pt] table[x=block, y=hit, col sep=comma] {Figures/per_layer/reg_wb_vitg.csv};
\addplot[only marks, mark=*, mark size=1.9pt, mark options={fill=white, draw=sizeG, line width=0.7pt}] coordinates {(27,97.27)};
\addplot[only marks, mark=*, mark size=1.9pt, mark options={fill=sizeG, draw=sizeG, line width=0.7pt}] coordinates {(39,81.90)};
\node[anchor=west, xshift=2.5pt, yshift=0pt, inner sep=1pt] at (axis cs:39,81.90) {\textminus15};
\node[anchor=west, xshift=2.5pt, yshift=0pt, inner sep=1pt] at (axis cs:23,82.27) {\textminus15};
\node[anchor=south, xshift=0pt, yshift=10pt, inner sep=1pt] at (axis cs:11,87.49) {ViT-B \textminus11};
\nextgroupplot[title={ImageNet val}, ymin=22.5, ymax=88.0, ytick={40,60,80}]
\addplot[color=sizeS, line width=1pt] table[x=block, y=hit, col sep=comma] {Figures/per_layer/reg_in_vits.csv};
\addplot[only marks, mark=*, mark size=1.9pt, mark options={fill=sizeS, draw=sizeS, line width=0.7pt}] coordinates {(11,83.43)};
\addplot[color=sizeB, line width=1pt] table[x=block, y=hit, col sep=comma] {Figures/per_layer/reg_in_vitb.csv};
\addplot[only marks, mark=*, mark size=1.9pt, mark options={fill=white, draw=sizeB, line width=0.7pt}] coordinates {(10,82.60)};
\addplot[only marks, mark=*, mark size=1.9pt, mark options={fill=sizeB, draw=sizeB, line width=0.7pt}] coordinates {(11,81.79)};
\addplot[color=sizeL, line width=1pt] table[x=block, y=hit, col sep=comma] {Figures/per_layer/reg_in_vitl.csv};
\addplot[only marks, mark=*, mark size=1.9pt, mark options={fill=white, draw=sizeL, line width=0.7pt}] coordinates {(18,84.47)};
\addplot[only marks, mark=*, mark size=1.9pt, mark options={fill=sizeL, draw=sizeL, line width=0.7pt}] coordinates {(23,76.67)};
\addplot[color=sizeG, line width=1pt] table[x=block, y=hit, col sep=comma] {Figures/per_layer/reg_in_vitg.csv};
\addplot[only marks, mark=*, mark size=1.9pt, mark options={fill=white, draw=sizeG, line width=0.7pt}] coordinates {(32,83.77)};
\addplot[only marks, mark=*, mark size=1.9pt, mark options={fill=sizeG, draw=sizeG, line width=0.7pt}] coordinates {(39,76.52)};
\node[anchor=west, xshift=2.5pt, yshift=0pt, inner sep=1pt] at (axis cs:39,76.52) {\textminus7};
\node[anchor=north, xshift=0pt, yshift=-7pt, inner sep=1pt] at (axis cs:23,76.67) {\textminus8};
\end{groupplot}
\node[anchor=south] at ([xshift=0.20in, yshift=15pt]group c1r1.north east) {\pgfplotslegendfromname{perlayer-hit-reg}};
\end{tikzpicture}%
\caption{\textbf{Per-layer attention hit rate for DINOv2 with register tokens~\cite{vitregister}.} Hit-rate counterpart to Figure~\ref{fig:per_layer_attention_mass} in the main paper. Filled dots mark each model's last block and open dots its peak; numbers give the drop from peak to last block, in points (drops under 2 points are not labeled). Hit rate stays high across late layers for all sizes, reflecting that registers absorb the high-norm artifact tokens that would otherwise capture the CLS attention argmax.}
\label{fig:per_layer_attention_hit_reg}
\end{figure}

\begin{figure}[H]
\centering
\definecolor{sizeS}{HTML}{0FA49D}%
\definecolor{sizeB}{HTML}{C65B2D}%
\definecolor{sizeL}{HTML}{943F69}%
\definecolor{sizeG}{HTML}{363FB4}%
\begin{tikzpicture}[font=\fontsize{8}{9.5}\selectfont]
\begin{groupplot}[
  group style={group size=2 by 1, horizontal sep=0.40in},
  scale only axis, width=2.18in, height=1.05in,
  xmin=-1.5, xmax=41.5, xtick={0,11,23,39}, xlabel={Block (0-indexed)}, xlabel style={yshift=2pt},
  axis lines*=left, axis line style={line width=0.5pt, -},
  tick align=outside, major tick length=2pt, minor tick length=0pt,
  tick style={line width=0.5pt, black},
  tick label style={font=\fontsize{8}{9.5}\selectfont, /pgf/number format/assume math mode=true},
  label style={font=\fontsize{8}{9.5}\selectfont},
  title style={font=\fontsize{8}{9.5}\selectfont\bfseries, at={(0,1)}, anchor=south west, inner sep=0pt, yshift=4pt},
  clip=false,
]
\nextgroupplot[title={Waterbirds}, ymin=32.3, ymax=85.3, ytick={40,50,60,70,80}, ylabel={Attention mass in GT box (\%)}, ylabel style={yshift=-2pt},
  legend to name=perlayer-mass-noreg,
  legend style={draw=none, fill=none, legend columns=6, font=\fontsize{8}{9.5}\selectfont,
    /tikz/every even column/.append style={column sep=8pt}}]
\addlegendimage{color=sizeS, line width=1pt}\addlegendentry{ViT-S}
\addlegendimage{color=sizeB, line width=1pt}\addlegendentry{ViT-B}
\addlegendimage{color=sizeL, line width=1pt}\addlegendentry{ViT-L}
\addlegendimage{color=sizeG, line width=1pt}\addlegendentry{ViT-G}
\addlegendimage{only marks, mark=*, mark size=1.9pt, mark options={fill=white, draw=black, line width=0.7pt}}\addlegendentry{peak block}
\addlegendimage{only marks, mark=*, mark size=1.9pt, mark options={fill=black, draw=black, line width=0.7pt}}\addlegendentry{last block}
\addplot[color=sizeS, line width=1pt] table[x=block, y=mass, col sep=comma] {Figures/per_layer/noreg_wb_vits.csv};
\addplot[only marks, mark=*, mark size=1.9pt, mark options={fill=white, draw=sizeS, line width=0.7pt}] coordinates {(10,82.49)};
\addplot[only marks, mark=*, mark size=1.9pt, mark options={fill=sizeS, draw=sizeS, line width=0.7pt}] coordinates {(11,79.52)};
\addplot[color=sizeB, line width=1pt] table[x=block, y=mass, col sep=comma] {Figures/per_layer/noreg_wb_vitb.csv};
\addplot[only marks, mark=*, mark size=1.9pt, mark options={fill=white, draw=sizeB, line width=0.7pt}] coordinates {(10,69.35)};
\addplot[only marks, mark=*, mark size=1.9pt, mark options={fill=sizeB, draw=sizeB, line width=0.7pt}] coordinates {(11,47.47)};
\addplot[color=sizeL, line width=1pt] table[x=block, y=mass, col sep=comma] {Figures/per_layer/noreg_wb_vitl.csv};
\addplot[only marks, mark=*, mark size=1.9pt, mark options={fill=white, draw=sizeL, line width=0.7pt}] coordinates {(21,59.65)};
\addplot[only marks, mark=*, mark size=1.9pt, mark options={fill=sizeL, draw=sizeL, line width=0.7pt}] coordinates {(23,46.84)};
\addplot[color=sizeG, line width=1pt] table[x=block, y=mass, col sep=comma] {Figures/per_layer/noreg_wb_vitg.csv};
\addplot[only marks, mark=*, mark size=1.9pt, mark options={fill=white, draw=sizeG, line width=0.7pt}] coordinates {(29,53.92)};
\addplot[only marks, mark=*, mark size=1.9pt, mark options={fill=sizeG, draw=sizeG, line width=0.7pt}] coordinates {(39,42.32)};
\node[anchor=west, xshift=2.5pt, yshift=0pt, inner sep=1pt] at (axis cs:39,42.32) {\textminus12};
\node[anchor=north, xshift=0pt, yshift=-7pt, inner sep=1pt] at (axis cs:23,46.84) {\textminus13};
\node[anchor=west, xshift=2.5pt, yshift=0pt, inner sep=1pt] at (axis cs:11,79.52) {\textminus3};
\node[anchor=south west, xshift=4pt, yshift=5pt, inner sep=1pt] at (axis cs:11,47.47) {\textminus22};
\nextgroupplot[title={ImageNet val}, ymin=45.1, ymax=74.0, ytick={50,60,70}]
\addplot[color=sizeS, line width=1pt] table[x=block, y=mass, col sep=comma] {Figures/per_layer/noreg_in_vits.csv};
\addplot[only marks, mark=*, mark size=1.9pt, mark options={fill=white, draw=sizeS, line width=0.7pt}] coordinates {(10,72.45)};
\addplot[only marks, mark=*, mark size=1.9pt, mark options={fill=sizeS, draw=sizeS, line width=0.7pt}] coordinates {(11,72.02)};
\addplot[color=sizeB, line width=1pt] table[x=block, y=mass, col sep=comma] {Figures/per_layer/noreg_in_vitb.csv};
\addplot[only marks, mark=*, mark size=1.9pt, mark options={fill=white, draw=sizeB, line width=0.7pt}] coordinates {(10,61.24)};
\addplot[only marks, mark=*, mark size=1.9pt, mark options={fill=sizeB, draw=sizeB, line width=0.7pt}] coordinates {(11,54.08)};
\addplot[color=sizeL, line width=1pt] table[x=block, y=mass, col sep=comma] {Figures/per_layer/noreg_in_vitl.csv};
\addplot[only marks, mark=*, mark size=1.9pt, mark options={fill=white, draw=sizeL, line width=0.7pt}] coordinates {(21,56.83)};
\addplot[only marks, mark=*, mark size=1.9pt, mark options={fill=sizeL, draw=sizeL, line width=0.7pt}] coordinates {(23,51.47)};
\addplot[color=sizeG, line width=1pt] table[x=block, y=mass, col sep=comma] {Figures/per_layer/noreg_in_vitg.csv};
\addplot[only marks, mark=*, mark size=1.9pt, mark options={fill=white, draw=sizeG, line width=0.7pt}] coordinates {(32,57.62)};
\addplot[only marks, mark=*, mark size=1.9pt, mark options={fill=sizeG, draw=sizeG, line width=0.7pt}] coordinates {(39,46.67)};
\node[anchor=west, xshift=2.5pt, yshift=0pt, inner sep=1pt] at (axis cs:39,46.67) {\textminus11};
\node[anchor=north west, xshift=1.5pt, yshift=-1.5pt, inner sep=1pt] at (axis cs:23,51.47) {\textminus5};
\node[anchor=south west, xshift=4pt, yshift=5pt, inner sep=1pt] at (axis cs:11,54.08) {\textminus7};
\end{groupplot}
\node[anchor=south] at ([xshift=0.20in, yshift=15pt]group c1r1.north east) {\pgfplotslegendfromname{perlayer-mass-noreg}};
\end{tikzpicture}%
\caption{\textbf{Per-layer CLS-to-patches attention mass for DINOv2 \emph{without} register tokens.} Without registers to absorb the late-layer high-norm ``artifact'' tokens identified by Darcet et al.~\cite{vitregister}, the inverse scaling trend in Figure~\ref{fig:per_layer_attention_mass} (with registers, main paper) becomes substantially sharper: the last-layer mass drop from ViT-S to ViT-G widens from $21$ points to $37$ points on Waterbirds and from $7$ to $25$ points on ImageNet val. ViT-S is largely unaffected (within $2.5$ points of its registered counterpart at the last layer), while ViT-B/L/G's last-layer attention drifts off-object in late blocks. The per-layer hit-rate version (Figure~\ref{fig:per_layer_attention_hit_noreg}) shows the effect even more starkly: for large ViTs without registers, the attention $\arg\max$ falls \emph{outside} the GT bounding box on more than $80\%$ of images at the final block. Registers do not eliminate the inverse scaling property we observe. Filled dots mark each model's last block and open dots its peak; numbers give the drop from peak to last block, in points.}
\label{fig:per_layer_attention_mass_noreg}
\end{figure}

\begin{figure}[H]
\centering
\definecolor{sizeS}{HTML}{0FA49D}%
\definecolor{sizeB}{HTML}{C65B2D}%
\definecolor{sizeL}{HTML}{943F69}%
\definecolor{sizeG}{HTML}{363FB4}%
\begin{tikzpicture}[font=\fontsize{8}{9.5}\selectfont]
\begin{groupplot}[
  group style={group size=2 by 1, horizontal sep=0.40in},
  scale only axis, width=2.18in, height=1.05in,
  xmin=-1.5, xmax=41.5, xtick={0,11,23,39}, xlabel={Block (0-indexed)}, xlabel style={yshift=2pt},
  axis lines*=left, axis line style={line width=0.5pt, -},
  tick align=outside, major tick length=2pt, minor tick length=0pt,
  tick style={line width=0.5pt, black},
  tick label style={font=\fontsize{8}{9.5}\selectfont, /pgf/number format/assume math mode=true},
  label style={font=\fontsize{8}{9.5}\selectfont},
  title style={font=\fontsize{8}{9.5}\selectfont\bfseries, at={(0,1)}, anchor=south west, inner sep=0pt, yshift=4pt},
  clip=false,
]
\nextgroupplot[title={Waterbirds}, ymin=-5.7, ymax=101.3, ytick={20,40,60,80}, ylabel={Hit rate (\%)}, ylabel style={yshift=-2pt},
  legend to name=perlayer-hit-noreg,
  legend style={draw=none, fill=none, legend columns=6, font=\fontsize{8}{9.5}\selectfont,
    /tikz/every even column/.append style={column sep=8pt}}]
\addlegendimage{color=sizeS, line width=1pt}\addlegendentry{ViT-S}
\addlegendimage{color=sizeB, line width=1pt}\addlegendentry{ViT-B}
\addlegendimage{color=sizeL, line width=1pt}\addlegendentry{ViT-L}
\addlegendimage{color=sizeG, line width=1pt}\addlegendentry{ViT-G}
\addlegendimage{only marks, mark=*, mark size=1.9pt, mark options={fill=white, draw=black, line width=0.7pt}}\addlegendentry{peak block}
\addlegendimage{only marks, mark=*, mark size=1.9pt, mark options={fill=black, draw=black, line width=0.7pt}}\addlegendentry{last block}
\addplot[color=sizeS, line width=1pt] table[x=block, y=hit, col sep=comma] {Figures/per_layer/noreg_wb_vits.csv};
\addplot[only marks, mark=*, mark size=1.9pt, mark options={fill=sizeS, draw=sizeS, line width=0.7pt}] coordinates {(11,95.58)};
\addplot[color=sizeB, line width=1pt] table[x=block, y=hit, col sep=comma] {Figures/per_layer/noreg_wb_vitb.csv};
\addplot[only marks, mark=*, mark size=1.9pt, mark options={fill=white, draw=sizeB, line width=0.7pt}] coordinates {(7,80.39)};
\addplot[only marks, mark=*, mark size=1.9pt, mark options={fill=sizeB, draw=sizeB, line width=0.7pt}] coordinates {(11,0.17)};
\addplot[color=sizeL, line width=1pt] table[x=block, y=hit, col sep=comma] {Figures/per_layer/noreg_wb_vitl.csv};
\addplot[only marks, mark=*, mark size=1.9pt, mark options={fill=white, draw=sizeL, line width=0.7pt}] coordinates {(17,88.97)};
\addplot[only marks, mark=*, mark size=1.9pt, mark options={fill=sizeL, draw=sizeL, line width=0.7pt}] coordinates {(23,10.30)};
\addplot[color=sizeG, line width=1pt] table[x=block, y=hit, col sep=comma] {Figures/per_layer/noreg_wb_vitg.csv};
\addplot[only marks, mark=*, mark size=1.9pt, mark options={fill=white, draw=sizeG, line width=0.7pt}] coordinates {(16,71.45)};
\addplot[only marks, mark=*, mark size=1.9pt, mark options={fill=sizeG, draw=sizeG, line width=0.7pt}] coordinates {(39,18.23)};
\node[anchor=west, xshift=2.5pt, yshift=0pt, inner sep=1pt] at (axis cs:39,18.23) {\textminus53};
\node[anchor=west, xshift=2.5pt, yshift=0pt, inner sep=1pt] at (axis cs:23,10.30) {\textminus79};
\node[anchor=west, xshift=2.5pt, yshift=0pt, inner sep=1pt] at (axis cs:11,0.17) {\textminus80};
\nextgroupplot[title={ImageNet val}, ymin=-0.4, ymax=87.8, ytick={20,40,60,80}]
\addplot[color=sizeS, line width=1pt] table[x=block, y=hit, col sep=comma] {Figures/per_layer/noreg_in_vits.csv};
\addplot[only marks, mark=*, mark size=1.9pt, mark options={fill=sizeS, draw=sizeS, line width=0.7pt}] coordinates {(11,83.04)};
\addplot[color=sizeB, line width=1pt] table[x=block, y=hit, col sep=comma] {Figures/per_layer/noreg_in_vitb.csv};
\addplot[only marks, mark=*, mark size=1.9pt, mark options={fill=white, draw=sizeB, line width=0.7pt}] coordinates {(8,61.01)};
\addplot[only marks, mark=*, mark size=1.9pt, mark options={fill=sizeB, draw=sizeB, line width=0.7pt}] coordinates {(11,5.08)};
\addplot[color=sizeL, line width=1pt] table[x=block, y=hit, col sep=comma] {Figures/per_layer/noreg_in_vitl.csv};
\addplot[only marks, mark=*, mark size=1.9pt, mark options={fill=white, draw=sizeL, line width=0.7pt}] coordinates {(17,66.90)};
\addplot[only marks, mark=*, mark size=1.9pt, mark options={fill=sizeL, draw=sizeL, line width=0.7pt}] coordinates {(23,19.57)};
\addplot[color=sizeG, line width=1pt] table[x=block, y=hit, col sep=comma] {Figures/per_layer/noreg_in_vitg.csv};
\addplot[only marks, mark=*, mark size=1.9pt, mark options={fill=white, draw=sizeG, line width=0.7pt}] coordinates {(0,58.87)};
\addplot[only marks, mark=*, mark size=1.9pt, mark options={fill=sizeG, draw=sizeG, line width=0.7pt}] coordinates {(39,29.58)};
\node[anchor=west, xshift=2.5pt, yshift=0pt, inner sep=1pt] at (axis cs:39,29.58) {\textminus29};
\node[anchor=west, xshift=2.5pt, yshift=0pt, inner sep=1pt] at (axis cs:23,19.57) {\textminus47};
\node[anchor=west, xshift=2.5pt, yshift=0pt, inner sep=1pt] at (axis cs:11,5.08) {\textminus56};
\end{groupplot}
\node[anchor=south] at ([xshift=0.20in, yshift=15pt]group c1r1.north east) {\pgfplotslegendfromname{perlayer-hit-noreg}};
\end{tikzpicture}%
\caption{\textbf{Per-layer attention hit rate for DINOv2 \emph{without} register tokens.} Hit-rate counterpart to Figure~\ref{fig:per_layer_attention_mass_noreg}. Without registers, the hit rate of ViT-B/L/G collapses sharply partway through the network (ViT-B layer 9, ViT-L layer 18, ViT-G layer 18), consistent with attention being captured by high-norm background tokens in late blocks. ViT-S is unaffected and remains the most reliable localizer at the last block. Numbers give the drop from each model's peak (open dot) to its last block (filled dot), in points.}
\label{fig:per_layer_attention_hit_noreg}
\end{figure}

\begin{figure}[H]
\centering
\definecolor{regblue}{HTML}{3E82BA}%
\begin{tikzpicture}[font=\fontsize{8}{9.5}\selectfont]
\begin{groupplot}[
  group style={group size=4 by 2, horizontal sep=0.13in, vertical sep=0.12in},
  scale only axis, width=1.12in, height=0.82in,
  ymin=-3, ymax=103, ytick={0,50,100},
  axis lines*=left, axis line style={line width=0.5pt, -},
  tick align=outside, major tick length=2pt, minor tick length=0pt,
  tick style={line width=0.5pt, black},
  tick label style={font=\fontsize{8}{9.5}\selectfont, /pgf/number format/assume math mode=true},
  label style={font=\fontsize{8}{9.5}\selectfont},
  title style={font=\fontsize{8}{9.5}\selectfont\bfseries, at={(0,1)}, anchor=south west, inner sep=0pt, yshift=4pt},
  clip=false,
]
\nextgroupplot[xmin=-0.94, xmax=11.94, xtick={0,5,11}, title={ViT-S}, ylabel={Waterbirds (\%)}, xticklabels={},
  legend to name=regtakeover,
  legend style={draw=none, fill=none, legend columns=5, font=\fontsize{8}{9.5}\selectfont,
    /tikz/every even column/.append style={column sep=7pt}},
  legend image code/.code={\draw[mark repeat=2, mark phase=2] plot coordinates {(0cm,0cm) (0.25cm,0cm) (0.5cm,0cm)};}]
\addlegendimage{color=accent, line width=1.1pt}\addlegendentry{\textbf{mass in GT box}}
\addlegendimage{black, line width=0.6pt}\addlegendentry{share on patches}
\addlegendimage{color=regblue, line width=0.8pt, dash pattern=on 3pt off 1.5pt}\addlegendentry{share on registers}
\addlegendimage{only marks, mark=*, mark size=1.7pt, mark options={fill=white, draw=accent, line width=0.7pt}}\addlegendentry{peak block}
\addlegendimage{only marks, mark=*, mark size=1.7pt, mark options={fill=accent, draw=accent, line width=0.7pt}}\addlegendentry{last block}
\addplot[black, line width=0.6pt] table[x=block, y=patches, col sep=comma] {Figures/register_takeover/wb_vits.csv};
\addplot[color=regblue, line width=0.8pt, dash pattern=on 3pt off 1.5pt] table[x=block, y=registers, col sep=comma] {Figures/register_takeover/wb_vits.csv};
\addplot[color=accent, line width=1.1pt] table[x=block, y=mass, col sep=comma] {Figures/register_takeover/wb_vits.csv};
\addplot[only marks, mark=*, mark size=1.7pt, mark options={fill=accent, draw=accent, line width=0.7pt}] coordinates {(11,82.03)};
\nextgroupplot[xmin=-0.94, xmax=11.94, xtick={0,5,11}, title={ViT-B}, xticklabels={}, yticklabels={}]
\addplot[black, line width=0.6pt] table[x=block, y=patches, col sep=comma] {Figures/register_takeover/wb_vitb.csv};
\addplot[color=regblue, line width=0.8pt, dash pattern=on 3pt off 1.5pt] table[x=block, y=registers, col sep=comma] {Figures/register_takeover/wb_vitb.csv};
\addplot[color=accent, line width=1.1pt] table[x=block, y=mass, col sep=comma] {Figures/register_takeover/wb_vitb.csv};
\addplot[only marks, mark=*, mark size=1.7pt, mark options={fill=accent, draw=accent, line width=0.7pt}] coordinates {(11,76.39)};
\nextgroupplot[xmin=-1.42, xmax=24.42, xtick={0,10,23}, title={ViT-L}, xticklabels={}, yticklabels={}]
\addplot[black, line width=0.6pt] table[x=block, y=patches, col sep=comma] {Figures/register_takeover/wb_vitl.csv};
\addplot[color=regblue, line width=0.8pt, dash pattern=on 3pt off 1.5pt] table[x=block, y=registers, col sep=comma] {Figures/register_takeover/wb_vitl.csv};
\addplot[color=accent, line width=1.1pt] table[x=block, y=mass, col sep=comma] {Figures/register_takeover/wb_vitl.csv};
\addplot[only marks, mark=*, mark size=1.7pt, mark options={fill=white, draw=accent, line width=0.7pt}] coordinates {(21,76.77)};
\addplot[only marks, mark=*, mark size=1.7pt, mark options={fill=accent, draw=accent, line width=0.7pt}] coordinates {(23,65.01)};
\nextgroupplot[xmin=-2.06, xmax=41.06, xtick={0,20,39}, title={ViT-G}, xticklabels={}, yticklabels={}]
\addplot[black, line width=0.6pt] table[x=block, y=patches, col sep=comma] {Figures/register_takeover/wb_vitg.csv};
\addplot[color=regblue, line width=0.8pt, dash pattern=on 3pt off 1.5pt] table[x=block, y=registers, col sep=comma] {Figures/register_takeover/wb_vitg.csv};
\addplot[color=accent, line width=1.1pt] table[x=block, y=mass, col sep=comma] {Figures/register_takeover/wb_vitg.csv};
\addplot[only marks, mark=*, mark size=1.7pt, mark options={fill=white, draw=accent, line width=0.7pt}] coordinates {(32,73.59)};
\addplot[only marks, mark=*, mark size=1.7pt, mark options={fill=accent, draw=accent, line width=0.7pt}] coordinates {(39,60.84)};
\nextgroupplot[xmin=-0.94, xmax=11.94, xtick={0,5,11}, ylabel={ImageNet val (\%)}]
\addplot[black, line width=0.6pt] table[x=block, y=patches, col sep=comma] {Figures/register_takeover/in_vits.csv};
\addplot[color=regblue, line width=0.8pt, dash pattern=on 3pt off 1.5pt] table[x=block, y=registers, col sep=comma] {Figures/register_takeover/in_vits.csv};
\addplot[color=accent, line width=1.1pt] table[x=block, y=mass, col sep=comma] {Figures/register_takeover/in_vits.csv};
\addplot[only marks, mark=*, mark size=1.7pt, mark options={fill=accent, draw=accent, line width=0.7pt}] coordinates {(11,74.55)};
\nextgroupplot[xmin=-0.94, xmax=11.94, xtick={0,5,11}, yticklabels={}]
\addplot[black, line width=0.6pt] table[x=block, y=patches, col sep=comma] {Figures/register_takeover/in_vitb.csv};
\addplot[color=regblue, line width=0.8pt, dash pattern=on 3pt off 1.5pt] table[x=block, y=registers, col sep=comma] {Figures/register_takeover/in_vitb.csv};
\addplot[color=accent, line width=1.1pt] table[x=block, y=mass, col sep=comma] {Figures/register_takeover/in_vitb.csv};
\addplot[only marks, mark=*, mark size=1.7pt, mark options={fill=accent, draw=accent, line width=0.7pt}] coordinates {(11,72.61)};
\nextgroupplot[xmin=-1.42, xmax=24.42, xtick={0,10,23}, yticklabels={}]
\addplot[black, line width=0.6pt] table[x=block, y=patches, col sep=comma] {Figures/register_takeover/in_vitl.csv};
\addplot[color=regblue, line width=0.8pt, dash pattern=on 3pt off 1.5pt] table[x=block, y=registers, col sep=comma] {Figures/register_takeover/in_vitl.csv};
\addplot[color=accent, line width=1.1pt] table[x=block, y=mass, col sep=comma] {Figures/register_takeover/in_vitl.csv};
\addplot[only marks, mark=*, mark size=1.7pt, mark options={fill=white, draw=accent, line width=0.7pt}] coordinates {(21,73.75)};
\addplot[only marks, mark=*, mark size=1.7pt, mark options={fill=accent, draw=accent, line width=0.7pt}] coordinates {(23,67.84)};
\nextgroupplot[xmin=-2.06, xmax=41.06, xtick={0,20,39}, yticklabels={}]
\addplot[black, line width=0.6pt] table[x=block, y=patches, col sep=comma] {Figures/register_takeover/in_vitg.csv};
\addplot[color=regblue, line width=0.8pt, dash pattern=on 3pt off 1.5pt] table[x=block, y=registers, col sep=comma] {Figures/register_takeover/in_vitg.csv};
\addplot[color=accent, line width=1.1pt] table[x=block, y=mass, col sep=comma] {Figures/register_takeover/in_vitg.csv};
\addplot[only marks, mark=*, mark size=1.7pt, mark options={fill=white, draw=accent, line width=0.7pt}] coordinates {(33,71.53)};
\addplot[only marks, mark=*, mark size=1.7pt, mark options={fill=accent, draw=accent, line width=0.7pt}] coordinates {(39,67.63)};
\end{groupplot}
\node[anchor=north] at ([xshift=2.435in, yshift=-16pt]group c1r2.south west) {Block (0-indexed)};
\node[anchor=south] at ([xshift=2.435in, yshift=17pt]group c1r1.north west) {\pgfplotslegendfromname{regtakeover}};
\end{tikzpicture}%
\caption{\textbf{The dip in patch-side localization coincides with attention shifting to register tokens.} For DINOv2 with registers on Waterbirds (top) and ImageNet val (bottom), we plot the per-layer fraction of CLS softmax mass on patch tokens (black) vs.\ register tokens (blue, dashed), alongside the patch-side mass-in-GT-bbox metric from Figure~\ref{fig:per_layer_attention_mass} (red). For ViT-L and ViT-G, register tokens absorb a growing share of CLS attention in late layers, eventually \emph{exceeding} patch attention in ViT-G's final block ($55\%$ registers vs.\ $40\%$ patches on Waterbirds); the patch-side localization metric drops correspondingly ($\sim\!13$ points from peak-localization layer to last block in ViT-G). ViT-S and ViT-B never experience this register takeover and continue to concentrate attention on patch tokens through their last block.}
\label{fig:register_attention_vs_localization}
\end{figure}

\section{Patch Size, Pretraining Objective, and Other Confounders}
\label{sec:patch_objective_controls}

\paragraph{The impact of patch size on localization.} We compare DINOv1 ViT-S and ViT-B at the /8 and /16 patch size. 

\begin{figure}[H]
\centering
\definecolor{famDinoOne}{HTML}{7C4607}%
\begin{tikzpicture}[font=\fontsize{8}{9.5}\selectfont]
\begin{groupplot}[
  group style={group size=2 by 1, horizontal sep=0.40in},
  scale only axis, width=2.22in, height=1.05in,
  xmin=-0.25, xmax=1.25, xtick={0,1}, xticklabels={ViT-S, ViT-B},
  axis lines*=left, axis line style={line width=0.5pt, -},
  tick align=outside, major tick length=2pt, minor tick length=0pt,
  tick style={line width=0.5pt, black},
  tick label style={font=\fontsize{8}{9.5}\selectfont, /pgf/number format/assume math mode=true},
  label style={font=\fontsize{8}{9.5}\selectfont},
  title style={font=\fontsize{8}{9.5}\selectfont\bfseries, at={(0,1)}, anchor=south west, inner sep=0pt, yshift=4pt},
  clip=false,
]
\nextgroupplot[title={Waterbirds}, ylabel={Attention mass in GT box (\%)}, ylabel style={yshift=-2pt}, ymin=63.3, ymax=77.3, ytick={65,70,75},
  legend to name=patchsize,
  legend style={draw=none, fill=none, legend columns=2, font=\fontsize{8}{9.5}\selectfont,
    /tikz/every even column/.append style={column sep=9pt}}]
\addlegendimage{color=famDinoOne, line width=0.8pt, dash pattern=on 3pt off 1.5pt, mark=square*, mark size=1.9pt, mark options={fill=famDinoOne, draw=famDinoOne, line width=0.6pt, solid}}\addlegendentry{patch size 8}
\addlegendimage{color=famDinoOne, line width=0.8pt, mark=square*, mark size=1.9pt, mark options={fill=famDinoOne!35, draw=famDinoOne, line width=0.6pt}}\addlegendentry{patch size 16}
\addplot[color=famDinoOne, line width=0.8pt, mark=square*, mark size=1.9pt, mark options={fill=famDinoOne!35, draw=famDinoOne, line width=0.6pt}] table[x=x, y=wb_mass, col sep=comma] {Figures/patch_size/patch16.csv};
\addplot[color=famDinoOne, line width=0.8pt, dash pattern=on 3pt off 1.5pt, mark=square*, mark size=1.9pt, mark options={fill=famDinoOne, draw=famDinoOne, line width=0.6pt, solid}] table[x=x, y=wb_mass, col sep=comma] {Figures/patch_size/patch8.csv};
\nextgroupplot[title={ImageNet val}, ymin=65.0, ymax=72.9, ytick={66,68,70,72}]
\addplot[color=famDinoOne, line width=0.8pt, mark=square*, mark size=1.9pt, mark options={fill=famDinoOne!35, draw=famDinoOne, line width=0.6pt}] table[x=x, y=in_mass, col sep=comma] {Figures/patch_size/patch16.csv};
\addplot[color=famDinoOne, line width=0.8pt, dash pattern=on 3pt off 1.5pt, mark=square*, mark size=1.9pt, mark options={fill=famDinoOne, draw=famDinoOne, line width=0.6pt, solid}] table[x=x, y=in_mass, col sep=comma] {Figures/patch_size/patch8.csv};
\end{groupplot}
\node[anchor=south] at ([xshift=0.20in, yshift=15pt]group c1r1.north east) {\pgfplotslegendfromname{patchsize}};
\end{tikzpicture}%
\caption{Finer patch sizes improve localization, but within each patch size family, we see inverse scaling with model size. Table~\ref{tab:dino_patch_size_control} reports precise values.}
\label{fig:dino_patch_size_control}
\end{figure}

\begin{table}[H]
\centering
\small
\caption{\textbf{DINOv1 patch-size vs. localization.}}
\label{tab:dino_patch_size_control}
\begin{tabular}{@{}lcccc@{}}
\toprule
& \multicolumn{2}{c}{\textbf{Waterbirds}} & \multicolumn{2}{c}{\textbf{ImageNet val}} \\
\cmidrule(lr){2-3} \cmidrule(lr){4-5}
\textbf{DINOv1 model} & \textbf{Mass} $\uparrow$ & \textbf{Hit} $\uparrow$ & \textbf{Mass} $\uparrow$ & \textbf{Hit} $\uparrow$ \\
\midrule
ViT-S/16 & 0.689 & 0.939 & 0.681 & 0.849 \\
ViT-S/8  & 0.757 & 0.978 & 0.720 & 0.868 \\
ViT-B/16 & 0.649 & 0.934 & 0.659 & 0.829 \\
ViT-B/8  & 0.671 & 0.950 & 0.677 & 0.842 \\
\bottomrule
\end{tabular}
\end{table}

\paragraph{Pretraining objective.} We compare DeiT-S/16 (supervised) and DINOv1 ViT-S/16 (self-distilled), matching architecture, patch size, and ImageNet-1K pretraining corpus. 

\begin{table}[H]
\centering
\small
\caption{\textbf{DINOv1 vs. DeiT localization.} Both models use ViT-S/16 and ImageNet-1K pretraining. }
\label{tab:dino_deit_control}
\begin{tabular}{@{}lcccc@{}}
\toprule
& \multicolumn{2}{c}{\textbf{Waterbirds}} & \multicolumn{2}{c}{\textbf{ImageNet val}} \\
\cmidrule(lr){2-3} \cmidrule(lr){4-5}
\textbf{Model / pretraining} & \textbf{Mass} $\uparrow$ & \textbf{Hit} $\uparrow$ & \textbf{Mass} $\uparrow$ & \textbf{Hit} $\uparrow$ \\
\midrule
DeiT-S/16 (supervised) & 0.394 & 0.157 & 0.442 & 0.301 \\
DINOv1 ViT-S/16 (self-distilled) & \textbf{0.689} & \textbf{0.939} & \textbf{0.681} & \textbf{0.849} \\
\bottomrule
\end{tabular}
\end{table}

\begin{table}[H]
\centering
\caption{Fixed-patch-size comparisons across SSL families. Mass is the fraction of last-layer patch attention within ground-truth bounding boxes; $\Delta$ is larger minus smaller model mass. WB denotes Waterbirds and IN denotes ImageNet validation. MAE uses B/16 versus L/16 to hold patch size fixed, rather than the larger H/14 checkpoint.}
\label{tab:pretraining_corpus_endpoints}
\begingroup
\small
\setlength{\tabcolsep}{4pt}
\begin{tabular}{@{}llclrr@{}}
\toprule
Family & Corpus & Size pair & WB mass & $\Delta$ WB & $\Delta$ IN \\
\midrule
DINOv1 (/16) & IN-1k & S $\to$ B & $0.689 \to 0.649$ & $-0.040$ & $-0.021$ \\
DINOv2 (/14) & LVD-142M & S $\to$ g & $0.820 \to 0.608$ & $-0.212$ & $-0.069$ \\
DINOv3 (/16) & LVD-1689M & S $\to$ 7B & $0.767 \to 0.577$ & $-0.190$ & $-0.074$ \\
iBOT (/16) & IN-1k & S $\to$ L & $0.696 \to 0.557$ & $-0.139$ & $-0.092$ \\
MAE (/16) & IN-1k & B $\to$ L & $0.474 \to 0.376$ & $-0.098$ & $-0.044$ \\
\bottomrule
\end{tabular}
\endgroup
\end{table}

\begin{table}[H]
\centering
\caption{Waterbirds attention mass / hit rate for image--text pretrained ViTs. Parameter counts refer to the vision encoder.}
\label{tab:clip_corpus_localization}
\begingroup
\small
\setlength{\tabcolsep}{4pt}
\begin{tabular}{@{}lcccc@{}}
\toprule
Model family / corpus & \shortstack{B/16\\86M} & \shortstack{L/14\\304M} & \shortstack{H/14\\632M} & \shortstack{g/14\\1.0B} \\
\midrule
OpenCLIP / LAION-2B & 0.480 / 0.166 & 0.561 / 0.815 & 0.526 / 0.431 & 0.522 / 0.585 \\
OpenAI CLIP / WIT-400M & 0.426 / 0.166 & 0.483 / 0.172 & --- & --- \\
\bottomrule
\end{tabular}
\endgroup
\end{table}

\begin{table}[H]
\centering
\caption{Reported downstream comparison with and without registers for the DINOv2 ViT-S selector / ViT-G embedder cross model result. Our method does not rely on registers. Waterbirds uses one $64\times64$ crop; Spawrious O2O Hard uses two crops of sizes $64\times64$ and $32\times32$. Cells are test accuracy / worst-group accuracy (\%).}
\label{tab:register_downstream_control}
\begingroup
\small
\setlength{\tabcolsep}{5pt}
\begin{tabular}{@{}lcc@{}}
\toprule
Register setting & Waterbirds & Spawrious O2O Hard \\
\midrule
With registers & 98.2 / 94.4 & 89.2 / 79.9 \\
Without registers & 98.2 / 94.4 & 91.7 / 82.1 \\
\bottomrule
\end{tabular}
\endgroup
\end{table}

\section{Attention Aggregation and Crop Selection}
\label{sec:method_details}

\paragraph{Head aggregation.} The CLS token's attention in the last block of an $H$-head ViT yields per-head maps $a^h \in \mathbb{R}^N$ given the query $q^h_{\text{CLS}}$ and key matrix $K^h \in \mathbb{R}^{N \times d_k}$ (with $d_k$ the per-head embedding dimension):
\[
a^h = \text{softmax}\!\left(\frac{q^h_{\text{CLS}} \cdot (K^h)^\intercal}{\sqrt{d_k}}\right) \in \mathbb{R}^N.
\]
We aggregate by mean over heads, $a = \frac{1}{H} \sum_{h=1}^{H} a^h \in \mathbb{R}^N$, before bilinearly upsampling to the input resolution. Alternative aggregators over heads (entropy-weighted, lowest-entropy $k=6$ mean, and max) are ablated in Appendix Tables \ref{tab:agg_ablation_spawrious_o2o}-\ref{tab:agg_ablation_imagenet_vitg}, and none perform better than the mean over all heads. Max lowers Spawrious O2O downstream accuracy by $1.3$ points.

\paragraph{Greedy non-overlapping crop selection.} Given the upsampled $M \times M$ attention map $a$ and a list of crop sizes $s_1, \dots, s_n$ ordered largest-to-smallest, we sequentially place crops $c_1, \dots, c_n$. Each crop $c_k = (x_k, y_k, s_k)$ is the location that maximizes the sum of attention mass within it, subject to non-overlap with previously placed crops. Letting $\mathrm{BoxSum}(a, x, y, x+s, y+s)$ denote the sum of attention values inside the crop starting at $(x,y)$ with size $s$:
\[
c_k = \underset{\substack{c=(x,y,s_k)\\ 0 \le x,y \le M-s_k\\ c \cap c_j = \varnothing,\ \forall j<k}}{\arg\max} \mathrm{BoxSum}(a, x, y, x+s_k, y+s_k), \qquad k=1,\dots,n.
\]
For a new dataset, a user can start from these defaults (k=2, sizes [64, 32]). This configuration is used across all five datasets we tried. 

\section{MetaShift Animals Dataset Card}
\label{sec:metashift_animals_data_card}

MetaShift \cite{MetaShift} is a dataset created from the Visual Genome in order to study learning under distribution shift. The dataset is grouped into subsets, where a subset is a co-occurrence of a particular class with another spurious feature. For example, an elephant with a person, an elephant with a bike, etc.

We created a subset of the MetaShift dataset focused on $8$-class animal classification under distribution shift. Our dataset includes the classes \texttt{bear, bird, cow, elephant, giraffe, horse, sheep, zebra} in various contexts. For example, elephant train contexts include \{\texttt{people, rope, wall, tree,...}\}, while elephant test contexts include \{\texttt{water, umbrella, camera, dirt,...}\}. Within each class, the train and test set context tags are disjoint; we split them equally (the context count) between the two sets. We remove any image IDs present in the test set from the train set to enforce zero overlap between train and test (this deduplication step is why the train set ends up being far smaller than the test set).

Per-class train/test image counts are summarized in Table~\ref{tab:metashift_animals_data_card_stats}, and visual examples across context tags for four representative classes are shown in Figure~\ref{fig:metashift_animals_context_examples}. The dataset can be found at \url{https://github.com/SreehariRamMohan/MetaShift_Animals}.

\begin{table}[H]
\centering
\small
\caption{MetaShift Animals (8-class benchmark) class counts.}
\label{tab:metashift_animals_data_card_stats}
\begin{tabular}{lcc}
\toprule
\textbf{Class} & \textbf{Train} & \textbf{Test} \\
\midrule
bear & 503 & 978 \\
bird & 1,122 & 977 \\
cow & 125 & 968 \\
elephant & 367 & 1,580 \\
giraffe & 238 & 2,368 \\
horse & 1,188 & 586 \\
sheep & 90 & 1,142 \\
zebra & 57 & 2,230 \\
\midrule
Total images & 3,690 & 10,829 \\
Class-context groups & 73 & 77 \\
Unique context tags & 54 & 47 \\
Unique image IDs & 2,385 & 6,043 \\
\bottomrule
\end{tabular}
\end{table}

\begin{figure}[H]
\centering
\input{Figures/metashift_animals/giraffe_appendix.tex}\\[6pt]
\input{Figures/metashift_animals/horse_appendix.tex}\\[6pt]
\input{Figures/metashift_animals/sheep_appendix.tex}\\[6pt]
\input{Figures/metashift_animals/zebra_appendix.tex}
\caption{\textbf{MetaShift Animals: full context visualization.} Companion to Figure~\ref{fig:metashift_animals_main}. Each panel shows one class (giraffe, horse, sheep, zebra) across multiple Visual Genome context tags. Within each panel, the top row contains training images and the bottom row contains test images; the per-image label is the context tag, which acts as the spurious feature. Train and test context tags are disjoint by construction.}
\label{fig:metashift_animals_context_examples}
\end{figure}

\section{Crop Count Ablations, Sensitivity, Configurations, and Computational Cost}
\label{sec:crop_count_ablation}

\begin{figure}[H]
\centering
\input{Figures/crop_count/figure15.tex}
\caption{\textbf{Qualitative example of $A^2$ attention-guided crops.} One to four $64\!\times\!64$ crops selected by ViT-S attention on a MetaShift Animals bear. Crop regions are outlined in red. $A^2$ focuses on the bear and excludes the surrounding water and aquarium context.}
\label{fig:crop_count_single}
\end{figure}

We run with a fixed crop configuration of two crops, the first $64 \times 64$ and the second $32 \times 32$. 

\begin{table}[H]
\centering
\small
\caption{Ablating the number of crops ($64\times64$) using ViT-S for both attention and embedding. Coverage \% is the proportion of the image seen by the classifier.}
\label{tab:ablation_crop_count}
\begin{tabular}{@{}ccccc@{}}
\toprule
& & \multicolumn{1}{c}{\textbf{Spawrious O2O}} & \multicolumn{2}{c}{\textbf{MetaShift Animals}} \\
\cmidrule(lr){3-3} \cmidrule(lr){4-5}
\textbf{\# Crops} & \textbf{Coverage (\%)} & \textbf{Test Acc (\%)} & \textbf{Test Acc (\%)} & \textbf{Worst-Class (\%)} \\
\midrule
1 & 8.2 & \textbf{91.5} & 83.1 & 61.8 \\
2 & 16.3 & 88.3 & \textbf{85.5} & 67.0 \\
3 & 24.5 & 87.0 & 84.7 & 66.2 \\
4 & 32.7 & 86.0 & 85.0 & \textbf{68.3} \\
\bottomrule
\end{tabular}
\end{table}

\begin{figure}[tbp]
\centering
\input{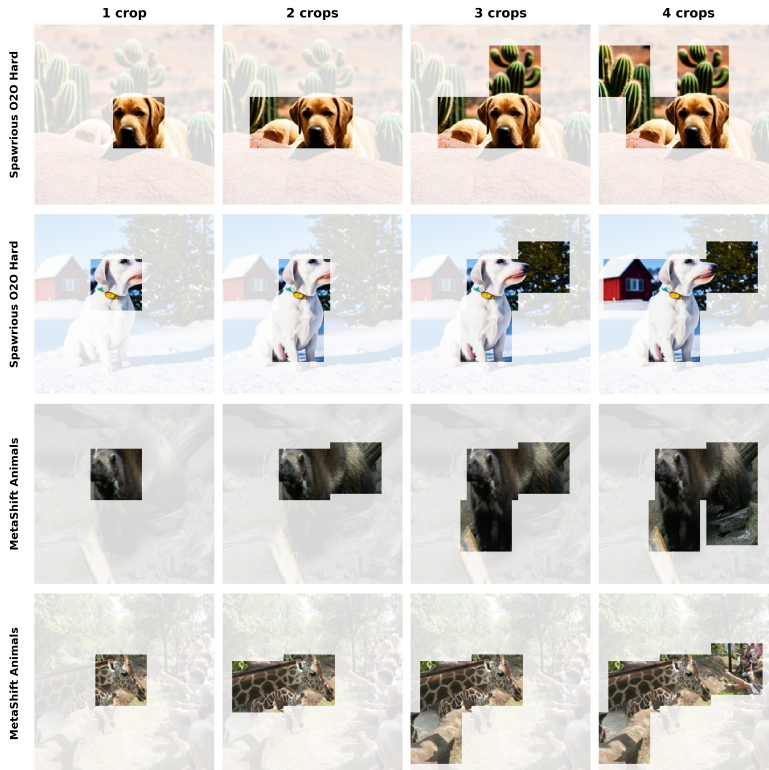}
\caption{Crop count ablation visualization. Each column shows 1--4 attention-guided $64\!\times\!64$ crops selected by ViT-S. Rows 1--2: Spawrious O2O Hard, Rows 3--4: MetaShift Animals. Crop regions are outlined in red.}
\label{fig:crop_count_ablation}
\end{figure}

\subsection{Inference cost}
\label{sec:inference_cost}

We report FLOPs and inference latency on one NVIDIA A100 GPU using FP32 and batch size one. $A^2$ requires one selector forward pass followed by one embedding forward pass per crop. Tables~\ref{tab:compute_cross_model}--\ref{tab:compute_same_backbone}
compare cross-model, zero-shot, and same-backbone configurations.

\begin{table}[H]
\centering
\caption{\textbf{Cross-model inference cost.} DINOv2 ViT-S selects crops and
DINOv2 ViT-G embeds them at $224\!\times\!224$. The $k=2$ row corresponds
to the shared main-paper crop count; $k=1$ is the one-crop variant.
Latencies include crop selection.}
\label{tab:compute_cross_model}
\begingroup
\small
\setlength{\tabcolsep}{4pt}
\begin{tabular}{@{}lrrr@{}}
\toprule
Configuration & GFLOPs & FLOPs ratio & Latency (ms) \\
\midrule
ViT-G, full image ($224^2$) & 304 & $1.00\times$ & 54 \\
$A^2$, $k=1$ & 311 & $1.02\times$ & 98 \\
$A^2$, $k=2$ & 615 & $2.02\times$ & 153 \\
ViT-G, full image ($518^2$) & 1790 & $5.9\times$ & 251 \\
\bottomrule
\end{tabular}
\endgroup
\end{table}

\begin{table}[H]
\centering
\caption{\textbf{Zero-shot inference cost.} OpenCLIP ViT-L/14 embeds images
or crops; $A^2_{\mathrm{ZS}}$ uses a ViT-S selector and one crop. TTR's
reported latency covers two CLIP passes and excludes clustering, so its
full end-to-end latency is not reported here.}
\label{tab:compute_zero_shot}
\begingroup
\small
\setlength{\tabcolsep}{4pt}
\begin{tabular}{@{}lrrr@{}}
\toprule
Configuration & GFLOPs & FLOPs ratio & Latency (ms) \\
\midrule
Full-image CLIP zero-shot & 81 & $1.00\times$ & 16 \\
$A^2_{\mathrm{ZS}}$, $k=1$ & 87 & $1.07\times$ & 60 \\
TTR (two CLIP passes) & 162 & $2.0\times$ & $31^{\dagger}$ \\
\bottomrule
\end{tabular}
\par\smallskip
{\footnotesize $\dagger$ Excludes clustering time.}
\endgroup
\end{table}

\begin{table}[H]
\centering
\caption{\textbf{Same-backbone inference cost.} DINOv2 ViT-S is used for
both selection and embedding. $S^2$ uses five $224\!\times\!224$ views
through \texttt{s2wrapper}. Latencies for $A^2$ include crop selection.}
\label{tab:compute_same_backbone}
\begingroup
\small
\setlength{\tabcolsep}{4pt}
\begin{tabular}{@{}lrrr@{}}
\toprule
Configuration & GFLOPs & FLOPs ratio & Latency (ms) \\
\midrule
ViT-S, full image ($224^2$) & 6.2 & $1.0\times$ & 5 \\
$A^2$, $k=1$ & 12 & $2.0\times$ & 48 \\
$A^2$, $k=2$ & 19 & $3.0\times$ & 53 \\
$S^2$ (five $224^2$ views) & 31 & $5.0\times$ & 12 \\
ViT-S, full image ($518^2$) & 47 & $7.5\times$ & 11 \\
\bottomrule
\end{tabular}
\endgroup
\end{table}

\subsection{Crop Configuration Options and TopK}

We ablate the design choices taken in $A^2$ for cropping. In particular, we test our decision to resize each crop to $224\times224$ rather than keeping its cropped size fed into the embedding model --- also exploring other intermediate sizes like $112\times112$. This answers the question of whether classification performance benefits from having the small crop resized into a larger (blurry) image. The blurriness of the crop only marginally hurts performance, costing 0.5 WGA on Waterbirds. We note that on our other benchmarks, cropping from native resolution shows no consistent benefit over the existing paper pipeline. Upsampling the crop to $224^2$ is helpful over processing in the crop native size.

We compare square crops to instead taking the patches with the highest attention and feeding them to the embedding model (with positional encodings preserved). We find that square cropping is an inductive bias that forces the model to focus its attention at a continuous unimodal, bimodal, or k-modal region(s) of the image. With top-$k$ selection, the model runs the risk of selecting artifacts or disjoint regions of the image which can lead it astray (as it does in Spawrious M2M). Square crops are also nice because they do not require any special hacking to the embedder and can be used off the shelf with $A^2_{\text{ZS}}$ (CLIP). The top-$k$ option has the efficiency argument of avoiding multiple forward embedding passes.

\begin{table}[H]
\centering
\caption{\textbf{Crop pixel source and embedder input resolution on Waterbirds.}
DINOv2 ViT-S selects one crop and the frozen DINOv2 ViT-G embeds it.
The crop width is $64$ in the $224\times224$ attention frame.
``Near-native'' rounds each original-resolution crop dimension to the nearest multiple of $14$ pixels for patch compatibility. }
\label{tab:crop_ablation_pixel_source}
\begingroup
\small
\setlength{\tabcolsep}{4pt}
\begin{tabular}{@{}llcrr@{}}
\toprule
Crop pixel source & Embedder input & Tokens & Acc. & WGA \\
\midrule
$224\times224$ frame & $224\times224$ & $256$ & $98.2$ & $94.4$ \\
$224\times224$ frame & $70\times70$ & $\sim25$ & $97.3$ & $92.5$ \\
$224\times224$ frame & $112\times112$ & $64$ & $97.9$ & $93.1$ \\
Original resolution & $224\times224$ & $256$ & $98.4$ & $94.9$ \\
Original resolution & Near-native & $\sim80$ & $98.0$ & $93.8$ \\
\bottomrule
\end{tabular}
\endgroup
\end{table}

\begin{table}[H]
\centering
\caption{\textbf{Crop size and resize-target sensitivity.}
Worst-group accuracy (\%) with a DINOv2 ViT-S selector and ViT-G embedder.
All crop pixels come from the resized $224\times224$ frame.
Waterbirds uses one crop of width $s$; Spawrious O2O Hard uses two crops
of widths $[s,s/2]$. The patch-aligned condition maps $s=32,64,128$ to
input widths $28,70,126$, respectively. }
\label{tab:crop_ablation_size_resize}
\begingroup
\small
\setlength{\tabcolsep}{5pt}
\begin{tabular}{@{}llrrr@{}}
\toprule
Benchmark & Embedder input & $s=32$ & $s=64$ & $s=128$ \\
\midrule
Waterbirds & $112\times112$ & $87.5$ & $93.1$ & $94.2$ \\
 & $224\times224$ & $89.9$ & $\mathbf{94.4}$ & $93.8$ \\
 & Patch-aligned & $54.4$ & $92.5$ & $93.9$ \\
\midrule
Spawrious O2O Hard & $112\times112$ & $76.6$ & $78.4$ & $68.9$ \\
 & $224\times224$ & $76.5$ & $\mathbf{80.0}$ & $76.6$ \\
 & Patch-aligned & $64.8$ & $74.8$ & $64.7$ \\
\bottomrule
\end{tabular}
\endgroup
\end{table}

\begin{table}[H]
\centering
\caption{\textbf{Square crops versus top-$K$ attention-ranked patch tokens.} The selector is DINOv2 ViT-S and the frozen embedder is DINOv2 ViT-G. The token budget is chosen to approximately match the image area covered by the crops before resizing. Square crops use one $64\times64$ crop for
Waterbirds and two crops of widths $[64,32]$ for the other benchmarks; each crop is resized to $224\times224$ before embedding. }
\label{tab:crop_ablation_topk}
\begingroup
\small
\setlength{\tabcolsep}{4pt}
\begin{tabular}{@{}lcrrrr@{}}
\toprule
 & & \multicolumn{2}{c}{Square crops} & \multicolumn{2}{c}{Top-$K$ tokens} \\
\cmidrule(lr){3-4}\cmidrule(lr){5-6}
Benchmark & $K$ & Acc. & WGA/WC & Acc. & WGA/WC \\
\midrule
Waterbirds & $21$ & $98.2$ & $94.4$ & $97.7$ & $94.9$ \\
Spawrious O2O Hard & $26$ & $89.2$ & $79.9$ & $86.7$ & $79.7$ \\
Spawrious M2M Hard & $26$ & $86.0$ & $64.6$ & $82.3$ & $50.9$ \\
MetaShift Cat vs.\ Dog & $26$ & $77.3$ & $76.6$ & $77.0$ & $75.8$ \\
MetaShift Animals & $26$ & $88.9$ & $70.8$ & $89.1$ & $70.1$ \\
\bottomrule
\end{tabular}
\endgroup
\end{table}

\section{Full Results for Spawrious, Waterbirds, and MetaShift}

\subsection{Spawrious}

\begin{table}[H]
\centering
\small
\caption{Spawrious Benchmark (O2O Hard + M2M Hard). We report the out of domain test accuracy. Note, by default the native resolution of all images is $224^2$. In cases where we use a larger image, we have bilinearly interpolated the image to be bigger. $A^2$ uses two square crops of width [64, 32] for Spawrious.}
\label{tab:spawrious}
\begin{tabular}{@{}llcc@{}}
\toprule
\textbf{Method} & \textbf{Backbone} & \textbf{O2O-Hard} & \textbf{M2M-Hard} \\
\midrule
\multicolumn{4}{l}{\textit{Baselines using DINOv2 backbone}} \\
Random Crops & ViT-S (21M) & 38.2 & 22.1 \\
Center Crop & ViT-S (21M) & 80.2 & 68.4 \\
Center Crop & ViT-G (1.1B) & 72.6 & 61.4 \\
\midrule
\multicolumn{4}{l}{\textit{Prior work (trained on Spawrious) \cite{lynch2023spawrious}:}} \\
ERM~\cite{lynch2023spawrious, jtt} & ResNet-50 & 71.32 & 58.70 \\
GroupDRO$^\dagger$~\cite{lynch2023spawrious,jtt, groupDRO} & ResNet-50 & 76.99 & 60.86 \\
IRM$^\dagger$ & ResNet-50 & 74.90 & 60.93 \\
CORAL~\cite{lynch2023spawrious, coral} & ResNet-50 & 79.65 & 67.97 \\
CausIRL~\cite{lynch2023spawrious, causIRL} & ResNet-50 & 80.40 & 68.93 \\
DFR$^\dagger$ & ViT-S (21M) & $86.1 \pm 0.4$ & $73.4 \pm 1.2$ \\
DFR$^\dagger$ & ViT-B (86M) & $90.1 \pm 0.3$ & $84.9 \pm 1.2$ \\
DFR$^\dagger$ & ViT-L (300M) & $94.3 \pm 0.2$ & $86.1 \pm 0.5$ \\
DFR$^\dagger$ & ViT-G (1.1B) & $95.7 \pm 0.3$ & $88.0 \pm 0.5$ \\
DFR$^\dagger$ + $A^2$ & ViT-S (21M) & $95.2 \pm 0.1$ & $86.2 \pm 0.4$ \\
\midrule
\multicolumn{4}{l}{\textit{DINOv2} backbone} \\
Full only & ViT-S (21M) & 80.5 & 66.6 \\
Full only ($518^2$ image) & ViT-S (21M) & 86.2 & 72.6 \\

Full only & ViT-B (86M) & 78.8 & 73.4 \\
Full only ($518^2$ image) & ViT-B (86M) & 82.5 & 82.4 \\
Full only & ViT-L (300M) & 80.8 & 81.9 \\
Full only ($518^2$ image) & ViT-L (300M) & 86.3 & 86.7 \\

Full only & ViT-G (1.1B) & 87.8 & 79.7 \\
Attending on Attention ($A^2$)\textsuperscript{$\ddagger$} & ViT-S (21M) & 91.0 & 82.4 \\
Attending on Attention ($A^2$)\textsuperscript{$\ddagger$} & ViT-B (86M) & 88.4 & 83.8 \\
Attending on Attention ($A^2$)\textsuperscript{$\ddagger$} & ViT-L (300M) & 82.1 & 83.5 \\
Attending on Attention ($A^2$)\textsuperscript{$\ddagger$} & ViT-G (1.1B) & 83.1 & 80.6 \\
$A^2$ cross-model (Table \ref{tab:cross_m2m})\textsuperscript{$\ddagger$} & ViT-S $\times$ ViT-L & 89.7 & 87.1\\
Scaling on Scales \cite{scalingScales} (scales=[1, 2] $224^2$) & ViT-S (21M) & 82.2 & 60.0 \\
\midrule
\multicolumn{4}{l}{\textit{OpenCLIP} backbone} \\
Full only & ViT-L-14 & 71.0 & 50.6 \\
OpenCLIP ZS prompt (full) & ViT-L-14 & 89.8 & 92.4 \\
\bottomrule
\end{tabular}
\par\smallskip
{\footnotesize $^\dagger$ Requires group labels during training.} \\
{\footnotesize $^\ddagger$ Key rows were rerun across 5 seeds. Values are reported as means and show no variation (std $=0.0$) since we use the lbfgs solver with sklearn Logistic Regression.}\\
\end{table}

\begin{table}[H]
\centering
\small
\caption{Worst-group accuracy on the test set of Spawrious O2O Hard and M2M Hard. DFR rows are means $\pm$ standard deviation over 5 seeds (DFR's group-balanced retraining is stochastic). The $A^2_\text{LR}$ cross-model row uses lbfgs Logistic Regression on the full training set and is therefore deterministic; we report the single fit. Groups are (breed, location) pairs; WGA is the worst-performing group on the test split.}
\label{tab:spawrious_wga_results}
\begin{tabular}{@{}lccc@{}}
\toprule
\textbf{Configuration} & \textbf{Backbone} & \textbf{O2O-Hard WGA} & \textbf{M2M-Hard WGA} \\
\midrule

DFR & ViT-S (21M) & $74.9 \pm 0.4$ & $49.9 \pm 1.7$ \\
\textbf{DFR} + \textbf{$A^2$} & ViT-S (21M) & $88.8 \pm 1.2$ & $64.6 \pm 0.8$ \\
DFR & ViT-B (86M) & $84.5 \pm 1.0$ & $69.7 \pm 1.5$ \\
DFR & ViT-L (300M) & $89.1 \pm 0.4$ & $72.3 \pm 0.8$ \\
DFR & ViT-G (1.1B) & $89.9 \pm 0.5$ & $73.9 \pm 0.7$ \\
\midrule
$A^2_\text{LR}$ cross-model & ViT-S $\times$ ViT-G & 79.9 & 64.6 \\
\bottomrule
\end{tabular}
\end{table}

\subsection{Waterbirds}

\begin{table}[H]
\centering
\small
\caption{Waterbirds Benchmark. We report the out of domain test average accuracy and worst-group accuracy (\%). Image size is $224^2$ by default; rows tagged ($518^2$ image) use bilinear upsampling. $A^2_{\text{LR}}$ (non cross model) uses one $64\!\times\!64$ attention-guided crop. $A^2_{\text{LR}}$ cross model uses our shared 2 crop configuration of one $64\!\times\!64$ crop and one $32\!\times\!32$ crop. $A^2_{\text{ZS}}$ uses ViT-S to select a $128\!\times\!128$ crop. Key rows ($\ddagger$) averaged over 5 seeds.}
\label{tab:waterbirds}
\begin{tabular}{@{}llcc@{}}
\toprule
\textbf{Method} & \textbf{Backbone} & \textbf{Avg Acc} & \textbf{Worst-Group} \\
\midrule

\multicolumn{4}{l}{\textit{Baselines using DINOv2 backbone}} \\
Random Crops & ViT-S (21M) & 70.7 & 32.4 \\
Center Crop & ViT-S (21M) & 90.9 & 78.7 \\
Center Crop & ViT-G (1.1B) & 98.0\textsuperscript{$\S$} & 94.9 \\

\midrule

\multicolumn{4}{l}{\textit{Prior work (trained on Waterbirds) \cite{jtt}:}} \\
ERM~\cite{lynch2023spawrious, jtt} & ResNet-50 & 97.3 & 72.6 \\
CVaR DRO & ResNet-50 & 96.0 & 75.9 \\
LfF & ResNet-50 & 91.2 & 78.0 \\
JTT~\cite{jtt} & ResNet-50 & 93.3 & 86.7 \\
Group DRO$^\dagger$~\cite{lynch2023spawrious,jtt, groupDRO} & ResNet-50 & 93.5 & 91.4 \\
DFR$^\dagger$ & ResNet-50 \cite{dfr} & $94.2 \pm 0.4$ & $92.9 \pm 0.2$  \\
DFR$^\dagger$ & ViT-S (21M) & $92.4 \pm 0.4$ & $90.4 \pm 0.6$ \\
DFR$^\dagger$ & ViT-B (86M) & $95.5 \pm 0.2$ & $94.0 \pm 0.5$ \\
DFR$^\dagger$ & ViT-L (300M) & $97.7 \pm 0.1$ & $97.1 \pm 0.1$ \\
DFR$^\dagger$ & ViT-G (1.1B) & $98.1 \pm 0.1$ & $96.4 \pm 0.1$ \\
DFR$^\dagger$ + $A^2$ & ViT-S (21M) & $91.9 \pm 0.5$ & $91.0 \pm 0.6$ \\
\midrule
\multicolumn{4}{l}{\textit{DINOv2 backbone}} \\
Full only & ViT-S (21M) & 88.7 & 72.7 \\
Full only ($518^2$ image) & ViT-S (21M) & 91.9 & 81.6 \\
Full only & ViT-B (86M) & 94.4 & 84.0 \\
Full only ($518^2$ image) & ViT-B (86M) & 96.3 & 89.9 \\
Full only & ViT-L (300M) & 96.9 & 89.3 \\
Full only ($518^2$ image) & ViT-L (300M) & 97.7 & 90.3 \\
Full only & ViT-G (1.1B) & 97.8 & 92.8 \\

Attending on Attention ($A^2$)\textsuperscript{$\ddagger$} & ViT-S (21M) & 93.0 & 80.4 \\
Attending on Attention ($A^2$)\textsuperscript{$\ddagger$} & ViT-B (86M) & 96.1 & 88.6 \\
Attending on Attention ($A^2$)\textsuperscript{$\ddagger$} & ViT-L (300M) & 93.8 & 85.2 \\
Attending on Attention ($A^2$)\textsuperscript{$\ddagger$} & ViT-G (1.1B) & 95.8 & 89.9 \\
$A^2$ cross-model \textsuperscript{$\ddagger$} & ViT-S $\times$ ViT-G & 98.3 & 95.3\\
Scaling on Scales \cite{scalingScales} (scales=[1, 2] $224^2$) & ViT-S (21M) & 88.3 & 73.1 \\
\midrule
\multicolumn{4}{l}{\textit{OpenCLIP backbone:}} \\
Full only & ViT-L-14 & 87.5 & 69.2 \\
OpenCLIP ZS prompt (full) & ViT-L-14 & 73.5 & 46.9 \\
\bottomrule
\end{tabular}
\par\smallskip
{\footnotesize $^\dagger$ Requires group labels during training.} \\
{\footnotesize $^\ddagger$ Key rows were rerun across 5 seeds. Values are reported as means and show no variation (std $=0.0$) since we use the lbfgs solver with sklearn Logistic Regression.}\\
{\footnotesize $^\S$ See Section \ref{sec:waterbirds_center_crop_baseline_strength} for explanation as to why center crop does so well in Waterbirds (dataset center bias)}
\end{table}

\subsection{MetaShift Cat vs.\ Dog}

\begin{table}[H]
\centering
\small
\caption{MetaShift Cat vs.\ Dog~\cite{MetaShift}. Test accuracy and worst-group accuracy (\%) reported.
Training has 100\% spurious correlation (cat=indoor, dog=outdoor); test has flipped correlation. Image size is $224^2$ by default; rows tagged ($518^2$ image) use bilinear upsampling. $A^2_{\text{LR}}$ uses two attention-guided crops of width $[64, 32]$; $A^2_{\text{ZS}}$ uses ViT-S to select a $128\!\times\!128$ crop. Key rows are reported over $5$ seeds. }
\label{tab:metashift_catdog}
\begin{tabular}{@{}llcc@{}}
\toprule
\textbf{Method} & \textbf{Backbone} & \textbf{Test Acc} & \textbf{Worst-Group} \\
\midrule

\multicolumn{4}{l}{\textit{Baselines using DINOv2 backbone}} \\
Random Crops & ViT-S (21M) & 44.0 & 40.6 \\
Center Crop & ViT-S (21M) & 64.0 & 61.3 \\
Center Crop & ViT-G (1.1B) & 76.9 & 74.8 \\

\midrule

\multicolumn{4}{l}{\textit{DINOv2 backbone}} \\
Full only & ViT-S (21M) & 59.8 & 56.0 \\
Full only ($518^2$ image) & ViT-S (21M) & 65.2 & 60.9 \\
Full only & ViT-B (86M) & 68.5 & 68.2 \\
Full only ($518^2$ image) & ViT-B (86M) & 75.3 & 73.9 \\
Full only & ViT-L (300M) & 80.0 & 79.2 \\
Full only ($518^2$ image) & ViT-L (300M) & 83.2 & 82.4 \\
Full only & ViT-G (1.1B) & 78.6 & 76.8 \\
DFR$^\dagger$ & ViT-S (21M) & $58.7 \pm 2.0$ & $54.5 \pm 3.9$ \\
DFR$^\dagger$ + $A^2$ & ViT-S (21M) & $69.4 \pm 0.8$ & $67.7 \pm 1.5$ \\

DFR$^\dagger$ & ViT-B (86M) & $68.6 \pm 3.4$ & $65.9 \pm 5.1$ \\
DFR$^\dagger$ & ViT-L (300M) & $73.6 \pm 1.9$ & $70.4 \pm 3.5$ \\
DFR$^\dagger$ & ViT-G (1.1B) & $76.0 \pm 2.0$ & $74.2 \pm 3.3$ \\
Attending on Attention ($A^2$)\textsuperscript{$\ddagger$} & ViT-S (21M) & 69.8 & 68.3 \\
Attending on Attention ($A^2$)\textsuperscript{$\ddagger$} & ViT-B (86M) & 73.9 & 71.8 \\
Attending on Attention ($A^2$)\textsuperscript{$\ddagger$} & ViT-L (300M) & 72.5 & 70.6 \\
Attending on Attention ($A^2$)\textsuperscript{$\ddagger$} & ViT-G (1.1B) & 76.1 & 73.7 \\
$A^2$ cross-model (Table \ref{tab:cross_catdog})\textsuperscript{$\ddagger$} & ViT-S $\times$ ViT-G & 77.3 & 76.6 \\
Scaling on Scales \cite{scalingScales} (scales=[1, 2] $224^2$) & ViT-S (21M) & 59.7 & 56.9 \\
OpenCLIP ZS prompt (full) & ViT-L-14 & 92.4 & 88.8 \\
\bottomrule
\end{tabular}
\par\smallskip
{\footnotesize $^\dagger$ Requires group labels during training.} \\
{\footnotesize $^\ddagger$ Key rows were rerun across 5 seeds. Values are reported as means and show no variation (std $=0.0$) since we use the lbfgs solver with sklearn Logistic Regression.}\\
\end{table}

\subsection{MetaShift Animals}

\begin{table}[H]
\centering
\small
\caption{MetaShift Animals (8-class Domain Generalization). Test accuracy and worst-class accuracy (\%) reported.
Within each class, training and test contexts are disjoint. Image size is $224^2$ by default; rows tagged ($518^2$ image) use bilinear upsampling. $A^2_{\text{LR}}$ uses two attention-guided crops of width $[64, 32]$; $A^2_{\text{ZS}}$ uses ViT-S to select a $128\!\times\!128$ crop. Key rows ($\ddagger$) averaged over 5 seeds.}
\label{tab:metashift_animals}
\begin{tabular}{@{}llcc@{}}
\toprule
\textbf{Method} & \textbf{Backbone} & \textbf{Test Acc} & \textbf{Worst-Class} \\
\midrule

\multicolumn{4}{l}{\textit{Baselines using DINOv2 backbone}} \\
Random Crops & ViT-S (21M) & 58.3 & 34.6 \\
Center Crop & ViT-S (21M) & 82.6 & 62.7 \\
Center Crop & ViT-G (1.1B) & 89.2 & 72.4 \\

\midrule

\multicolumn{4}{l}{\textit{DINOv2 backbone}} \\
Full only & ViT-S (21M) & 83.2 & 63.2 \\
Full only ($518^2$ image) & ViT-S (21M) & 87.5 & 68.3 \\
Full only & ViT-B (86M) & 88.3 & 70.3 \\
Full only ($518^2$ image) & ViT-B (86M) & 89.6 & 72.7 \\
Full only & ViT-L (300M) & 89.7 & 75.9 \\
Full only ($518^2$ image) & ViT-L (300M) & 90.1 & 77.6 \\
Full only & ViT-G (1.1B) & 90.6 & 75.9 \\

DFR$^\dagger$ & ViT-S (21M) & $88.3 \pm 0.1$ & $72.0 \pm 2.0$ \\
DFR$^\dagger$ + $A^2$ & ViT-S (21M) & $88.9 \pm 0.1$ & $72.9 \pm 1.3$ \\
DFR$^\dagger$ & ViT-B (86M) & $90.3 \pm 0.3$ & $73.5 \pm 2.2$ \\
DFR$^\dagger$ & ViT-L (300M) & $91.4 \pm 0.0$ & $77.9 \pm 0.8$ \\
DFR$^\dagger$ & ViT-G (1.1B) & $91.4 \pm 0.3$ & $76.6 \pm 1.9$ \\

Attending on Attention ($A^2$)\textsuperscript{$\ddagger$} & ViT-S (21M) & 85.1 & 67.0 \\
Attending on Attention ($A^2$)\textsuperscript{$\ddagger$} & ViT-B (86M) & 87.5 & 67.9 \\
Attending on Attention ($A^2$)\textsuperscript{$\ddagger$} & ViT-L (300M) & 85.2 & 69.1 \\
Attending on Attention ($A^2$)\textsuperscript{$\ddagger$} & ViT-G (1.1B) & 87.7 & 69.2 \\
$A^2$ cross-model (Table \ref{tab:cross_animals})\textsuperscript{$\ddagger$} & ViT-S $\times$ ViT-G & 88.9 & 70.8 \\

Scaling on Scales \cite{scalingScales} (scales=[1, 2] $224^2$) & ViT-S (21M) & 84.7 & 65.5 \\

\midrule
\multicolumn{4}{l}{\textit{OpenCLIP backbone}} \\

OpenCLIP ZS prompt (full) & ViT-L-14 & 90.4 & 58.8 \\
\bottomrule
\end{tabular}
\par\smallskip
{\footnotesize $^\dagger$ Requires group labels during training.} \\
{\footnotesize $^\ddagger$ Key rows were rerun across 5 seeds. Values are reported as means and show no variation (std $=0.0$) since we use the lbfgs solver with sklearn Logistic Regression.}\\
\end{table}

\section{Cross Model Matrices}
Cross model matrices across all $5$ datasets are shown below. In each matrix, \textbf{Bold} = best in column (best attention for each embedding) and \textit{Italic} = diagonal entries which use the same model for embedding and attention computation. Each matrix reports that dataset's standard metric: test accuracy for Spawrious O2O/M2M, worst-group accuracy for Waterbirds and Cat vs.\ Dog, and worst-class accuracy for Animals.

\subsection{Spawrious}

\begin{table}[H]
\centering
\small
\caption{Spawrious O2O Hard: Cross-model attention $\times$ embedding matrix. Rows = attention source, columns = embedding model. Values are test accuracy.}
\label{tab:cross_o2o}
\begin{tabular}{lcccc}
\toprule
\textbf{Attn $\backslash$ Emb} & \textbf{ViT-S} & \textbf{ViT-B} & \textbf{ViT-L} & \textbf{ViT-G} \\
\midrule
ViT-S attn & \textbf{\textit{91.0}} & \textbf{89.5} & \textbf{89.7} & \textbf{89.2} \\
ViT-B attn & 89.6 & \textit{88.4} & 88.2 & 87.9 \\
ViT-L attn & 85.2 & 82.7 & \textit{82.1} & 84.6 \\
ViT-G attn & 86.3 & 84.3 & 84.6 & \textit{83.1} \\
\bottomrule
\end{tabular}
\end{table}

\begin{table}[H]
\centering
\small
\caption{Spawrious M2M Hard: Cross-model attention $\times$ embedding matrix. \textbf{ViT-S attn $\rightarrow$ ViT-L emb (87.1) is the overall best}, beating all diagonal entries. Values are test accuracy.}
\label{tab:cross_m2m}
\begin{tabular}{lcccc}
\toprule
\textbf{Attn $\backslash$ Emb} & \textbf{ViT-S} & \textbf{ViT-B} & \textbf{ViT-L} & \textbf{ViT-G} \\
\midrule
ViT-S attn & \textbf{\textit{82.4}} & \textbf{84.6} & \textbf{87.1} & \textbf{86.0} \\
ViT-B attn & 81.0 & \textit{83.8} & 85.3 & 84.8 \\
ViT-L attn & 76.5 & 80.2 & \textit{83.5} & 82.3 \\
ViT-G attn & 74.7 & 79.9 & 81.2 & \textit{80.6} \\
\bottomrule
\end{tabular}
\end{table}

\subsection{Waterbirds}

\begin{table}[H]
\centering
\small
\caption{Waterbirds: Cross-model attention $\times$ embedding matrix. \textbf{ViT-S attn $\rightarrow$ ViT-G emb (94.4 WGA)} beats all diagonal entries including ViT-G $\rightarrow$ ViT-G (89.9). Values are worst-group accuracy. All runs in this matrix use a single $64 \times 64$ crop.}
\label{tab:cross_waterbirds}
\begin{tabular}{lcccc}
\toprule
\textbf{Attn $\backslash$ Emb} & \textbf{ViT-S} & \textbf{ViT-B} & \textbf{ViT-L} & \textbf{ViT-G} \\
\midrule
ViT-S attn & \textbf{\textit{80.4}} & \textbf{90.5} & \textbf{93.1} & \textbf{94.4} \\
ViT-B attn & 77.9 & \textit{88.6} & 91.0 & 93.5 \\
ViT-L attn & 67.9 & 75.5 & \textit{85.2} & 86.0 \\
ViT-G attn & 71.7 & 81.6 & 86.4 & \textit{89.9} \\
\bottomrule
\end{tabular}
\end{table}

\subsection{MetaShift Cat vs.\ Dog}

\begin{table}[H]
\centering
\small
\caption{MetaShift Cat vs.\ Dog: Cross-model attention $\times$ embedding matrix. ViT-S attention is best for 3/4 embedding columns. Values are worst-group accuracy.}
\label{tab:cross_catdog}
\begin{tabular}{lcccc}
\toprule
\textbf{Attn $\backslash$ Emb} & \textbf{ViT-S} & \textbf{ViT-B} & \textbf{ViT-L} & \textbf{ViT-G} \\
\midrule
ViT-S attn & \textit{68.3} & \textbf{76.2} & \textbf{75.5} & \textbf{76.6} \\
ViT-B attn & \textbf{69.3} & \textit{71.8} & 74.0 & 75.8 \\
ViT-L attn & 64.4 & 70.8 & \textit{70.6} & 73.6 \\
ViT-G attn & 66.2 & 70.6 & 73.2 & \textit{73.7} \\
\bottomrule
\end{tabular}
\end{table}

\subsection{MetaShift Animals}

\begin{table}[H]
\centering
\small
\caption{MetaShift Animals: Cross-model attention $\times$ embedding matrix. ViT-S attention is best for $3/4$ embedding columns. Values are worst-class accuracy.}
\label{tab:cross_animals}
\begin{tabular}{lcccc}
\toprule
\textbf{Attn $\backslash$ Emb} & \textbf{ViT-S} & \textbf{ViT-B} & \textbf{ViT-L} & \textbf{ViT-G} \\
\midrule
ViT-S attn & \textbf{\textit{67.0}} & \textbf{68.2} & 70.3 & \textbf{70.8} \\
ViT-B attn & 64.6 & \textit{67.9} & 69.6 & 69.0 \\
ViT-L attn & 62.1 & 65.8 & \textit{69.1} & 70.0 \\
ViT-G attn & 64.0 & 64.8 & \textbf{70.4} & \textit{69.2} \\
\bottomrule
\end{tabular}
\end{table}

\section{Adapting Attention}
\label{supp:adapting_attention}

\begin{figure}[t!]
    \centering
    \input{Figures/celeba_adapter/celeba_adapter.tex}
    \caption{\textbf{Visualizing the MLP Adapter.} The MLP Adapter changes the attention map for the {blond male} class in the CelebA dataset. Default attention focuses on the face or lower body; adapted attention focuses on the hair, which is the label-defining feature.}
    \label{fig:mlp_adapter_celebA}
\end{figure}

$A^2$ assumes the attention map naturally concentrates attention on the task-relevant features of the image. One prominent counter-example is CelebA \cite{celebA}, where the task is to predict whether a person has blond hair or not given a photo of their face, with the spurious feature being gender (since blond men are a rarity). The default attention maps will focus attention on the face or lower body (Figure \ref{fig:mlp_adapter_celebA} shows this). To address this limitation, we initialize a small MLP adapter network ($3{,}000$ params) whose input is the latent activations of the ViT, and whose output is added to the attention map from the ViT. The MLP is trained end-to-end, by backpropagating through the logistic regression fit and patch selection process (using DPS \cite{dps}). This proof-of-concept approach allows us to refocus attention to the hair in CelebA and improves worst-group accuracy (blond men) from $23.33\%$ to $66.11\%$, outperforming a baseline logistic regression fit on ViT-S embeddings of the full image by $35.01\%$ (baseline gets $31.1\%$ WGA). See Appendix Tables \ref{tab:celeba_eval_adapter}, \ref{tab:celeba_groups_adapter} for full results.

\begin{figure}[H]
    \centering
    \input{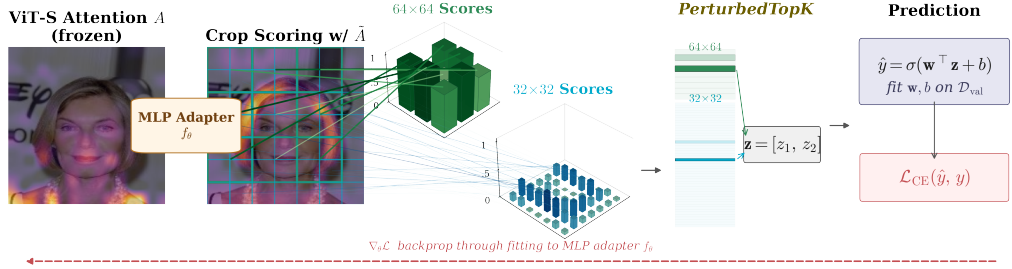}
    \caption{Learning the MLP attention adapter by differentiating through classifier fit using differentiable patch selection. The adapter learns to focus on the hair.}
    \label{fig:dps_adapter_pipeline}
\end{figure}

The zero-shot method assumes that the features the attention map focuses on are the ones relevant to the task; this assumption holds for the 5 benchmarks we test, but it may not always be the case. In this scenario, we can learn a simple MLP adapter network ($\approx 3{,}000$ parameters), which operates on the patch features from the ViT to produce an augmented $\tilde{A}$ which allows us to $\mathrm{Crop}_i(x;\;\tilde{A})$, embed with $\mathcal{T}(c_i)$, and fit a classifier using crops from a better attention map. Figure~\ref{fig:dps_adapter_pipeline} illustrates the full pipeline.

We use differentiable patch selection (DPS) \cite{dps} with a fixed set of 58 possible crops (9 $64\times64$ and 49 $32\times32$ crops in the $224\times224$ image) to get $k$ weighted crops $c_1(x), \ldots, c_k(x)$ from the attention map, $\tilde{A}$. We compute the embedding $z$ by concatenating the embedded crops using the embedding model, $\mathcal{T}$:

\[
z(x) = \left[\,\mathcal{T}(c_1(x)),\;\ldots,\;\mathcal{T}(c_k(x))\,\right]
  \in \mathbb{R}^{k \cdot d}
\]

We then fit a logistic regression classifier by taking a few gradient steps on $w, b$, the classifier weights using cross entropy loss $\ell$.

\[
  (w^{t+1}, b^{t+1}) = (w^t, b^t) - \eta \nabla_{w,b}\ell\left(z(x) w^t + b^t, y\right), \quad t=0,\dots,T-1.
\]

We evaluate the classifier on a held out partition of the training set, compute the cross entropy loss, and backpropagate through the classifier fitting and patch selection to optimize the MLP adapter weights.

\begin{table}[H]
\centering
\small
\caption{MLP Adapter Evaluation on CelebA Dataset. Adapter and default attention results are reported as mean $\pm$ std over 5 seeds.
}
\label{tab:celeba_eval_adapter}
\begin{tabular}{lcc}
\toprule
\textbf{Configuration} & \textbf{Test} & \textbf{Worst-Group Acc.} \\
\midrule
$A^2$ with MLP Adapter (384-dim) & $91.07 \pm 0.16$\% & $\textbf{66.11} \pm \textbf{1.57}$\% \\
$A^2$ default attention (384-dim) & $78.41 \pm 0.19$\% & $23.33 \pm 0.93$\% \\
ViT-S classifier (full image $224 \times 224$) & \textbf{94.2}\% & 31.1\% \\
\bottomrule
\end{tabular}
\end{table}

\begin{table}[H]
\centering
\small
\caption{Per-Group Accuracy Breakdown on CelebA Test Set. The MLP adapter increases performance on the minority class (blond men). Adapter and default attention results are reported as mean $\pm$ std over 5 seeds.}
\label{tab:celeba_groups_adapter}
\begin{tabular}{lccc}
\toprule
\textbf{Group} & \textbf{MLP Adapter} & \textbf{Default Attn} &
\textbf{ViT-S Classifier} \\
\midrule
Non-blond female (n=9767) & $86.96 \pm 0.37$\% & $67.52 \pm 0.37$\% & 95.1\%\\
Non-blond male (n=7535) & $96.01 \pm 0.18$\% & $92.90 \pm 0.25$\% & 99.4\%\\
Blond female (n=2480) & $94.12 \pm 0.38$\% & $81.31 \pm 0.39$\% & 78.8\%\\
Blond male (n=180) & $66.11 \pm 1.57$\% & $23.33 \pm 0.93$\% & 31.1\%\\
\bottomrule
\end{tabular}
\end{table}

\section{On center crop as a strong baseline for Waterbirds (Table \ref{tab:waterbirds})}
\label{sec:waterbirds_center_crop_baseline_strength}

\begin{figure}[h]
\centering
\begin{subfigure}[b]{0.155\textwidth}
    \centering
    \includegraphics[width=\linewidth]{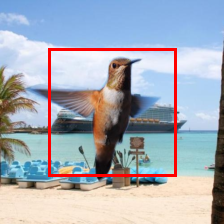}
\end{subfigure}
\begin{subfigure}[b]{0.155\textwidth}
    \centering
    \includegraphics[width=\linewidth]{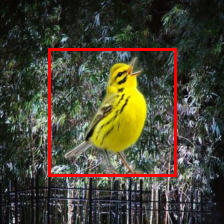}
\end{subfigure}
\begin{subfigure}[b]{0.155\textwidth}
    \centering
    \includegraphics[width=\linewidth]{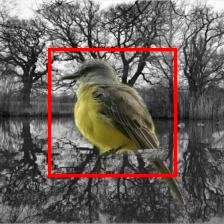}
\end{subfigure}
\hspace{0.02\textwidth}
\begin{subfigure}[b]{0.155\textwidth}
    \centering
    \includegraphics[width=\linewidth]{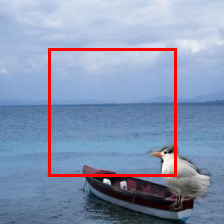}
\end{subfigure}
\begin{subfigure}[b]{0.155\textwidth}
    \centering
    \includegraphics[width=\linewidth]{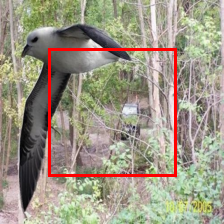}
\end{subfigure}
\begin{subfigure}[b]{0.155\textwidth}
    \centering
    \includegraphics[width=\linewidth]{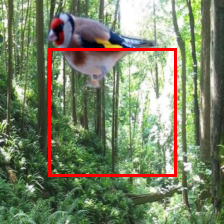}
\end{subfigure}
\caption{Examples from the bottom 1\% (left three) subset and top 1\% (right three) subsets. Red boxes show the center crop. Notice how in the top 1\%, the center crop completely misses the bird, which explains the large drop off in Table \ref{tab:waterbirds_distance_tails}.}
\label{fig:waterbirds_top_bottom_1pct}
\end{figure}

Simply center cropping the image to $128 \times 128$, embedding the crop, and fitting a logistic regression classifier seems like a competitive baseline. This is because Waterbirds is heavily biased toward placing the main subject (bird) at the center of the image. To illustrate this point, we sorted the test set birds by distance to the image center, and showed the performance of a center crop classifier on various percentiles (of bird distance to center). The results (Table \ref{tab:waterbirds_distance_tails}) show a clear trend that as the main subject drifts away from the center, the $A^2$ fitted classifier outperforms the center crop classifier. In the $58$ test set images with the bird furthest away from the image centroid, $A^2$ outperforms center crop by $6.9$ points, while for the $58$ test set images closest to the image centroid, $A^2$ actually underperforms the center crop classifier by $1.7$ points (the bird is exactly centered here so center cropping is the most optimal crop to take). For reference, we also include iFAM $K{=}8$ results in the same table; iFAM relies on the center bias as well, and performance drops when going from subset bottom1 to top1 ($100.0\% \to 81.8\%$), consistent with iFAM exploiting Waterbirds' center bias more than $A^2$ does. We show some images from the top 1\% and bottom 1\% in Figure \ref{fig:waterbirds_top_bottom_1pct}.

\begin{table}[H]
\centering
\caption{Distance stratification on Waterbirds test set. Center crop uses a fixed $128\times128$ center crop with a ViT-S logistic regression classifier; $A^2$ uses one $64\times64$ attention-guided crop with a ViT-S logistic regression classifier; iFAM K=8 is trained on the full Waterbirds train set (see Appendix~\ref{sec:ifam_protocol}). Models are trained once on the full Waterbirds training set and evaluated on each test-set subset. Test set is sorted from the bird being closest to the image center (``bottom'') to furthest from the image center (``top''); the ``top1'' subset is the $58$ test birds furthest from the image centroid. WGA results for the top and bottom 1st and 5th percentiles may have higher noise due to minority groups having $<20$ samples in those subsets, but are included for reference. }
\label{tab:waterbirds_distance_tails}
\setlength{\tabcolsep}{4pt}
\renewcommand{\arraystretch}{1.1}
\small
\begin{tabular}{@{}lccccccc@{}}
\toprule
\textbf{Subset} & \textbf{N} & \textbf{\shortstack{Center Crop\\Acc}} & \textbf{\shortstack{Center Crop\\WGA}} & \textbf{\shortstack{$A^2$\\Acc}} & \textbf{\shortstack{$A^2$\\WGA}} & \textbf{\shortstack{iFAM $K{=}8$\\Acc}} & \textbf{\shortstack{iFAM $K{=}8$\\WGA}} \\
\midrule
top1     & 58   & 84.5 & 63.6 & 91.4 & 72.7 & 98.3 & 81.8 \\
top5     & 290  & 89.0 & 75.6 & 93.8 & 87.8 & 98.3 & 92.7 \\
top10    & 580  & 89.8 & 78.4 & 93.1 & 84.6 & 98.4 & 92.3 \\
bottom10 & 580  & 91.6 & 84.5 & 92.2 & 78.0 & 98.6 & 96.6 \\
bottom5  & 290  & 92.4 & 85.8 & 92.4 & 74.1 & 99.7 & 100.0 \\
bottom1  & 58   & 93.1 & 86.7 & 91.4 & 83.3 & 100.0 & 100.0 \\
\midrule
full     & 5794 & 90.9 & 78.7 & 93.0 & 80.4 & 98.8 & 96.1 \\
\bottomrule
\end{tabular}
\renewcommand{\arraystretch}{1.0}
\setlength{\tabcolsep}{6pt}
\end{table}

\section{DFR Baseline Implementation Details}

All DFR rows run only Stage 2 of \cite{dfr}, last-layer retraining on a group-balanced subset, applied to frozen pretrained DINOv2 features. We do not fine-tune the backbone (Stage 1 ERM) on the biased train set; this is the natural baseline because $A^2$ also operates on frozen DINOv2 features, so comparing against balanced last-layer retraining on the same backbone isolates the effect of group balancing versus attention-guided cropping. We base our code on the public DFR repo \url{https://github.com/izmailovpavel/spurious_feature_learning/blob/main/dfr_evaluate_spurious.py} (modified to use frozen DINOv2 embeddings) and fit L1-regularized logistic regression (sklearn \texttt{liblinear}) on standardized features, grid-searching $C \in \{1.0, 0.7, 0.3, 0.1, 0.07, 0.03, 0.01\}$ and (for binary tasks) per-class weights to maximize worst-group accuracy on a tuning split, then averaging weights over $20$ retrains on independently subsampled group-balanced sets. \textbf{Waterbirds} uses DFR(Val) (official val split 50/50 for tuning vs.\ retraining). \textbf{Spawrious O2O/M2M Hard}, \textbf{MetaShift Cat vs.\ Dog}, and \textbf{MetaShift Animals} have no provided val, so we use DFR(Train): $20\%$ of train held out as pseudo-val for hyperparameter tuning, group-balanced retraining on the remaining $80\%$.

\section{Attention Concentration and Different Head Aggregators}

\begin{table}[H]
\centering
\small
\caption{Attention concentration across DINOv2 model sizes (ImageNet validation, 50k images). Interpretation, 3rd column, first row: 48\% of attention goes to 10\% of patches.}
\label{tab:attention_entropy}
\begin{tabular}{@{}lccc@{}}
\toprule
\textbf{Model} & \textbf{Entropy (norm)} & \textbf{Top-10\%} & \textbf{Top-25\%} \\
\midrule
ViT-S (21M) & $\mathbf{0.9219 \pm 0.0382}$ & $\mathbf{0.4814 \pm 0.1504}$ & $\mathbf{0.7430 \pm 0.1259}$ \\
ViT-B (86M) & $0.9292 \pm 0.0316$ & $0.4653 \pm 0.1305$ & $0.7207 \pm 0.1098$ \\
ViT-L (300M) & $0.9343 \pm 0.0271$ & $0.4528 \pm 0.1090$ & $0.7052 \pm 0.0991$ \\
ViT-G (1.1B) & $0.9351 \pm 0.0226$ & $0.4522 \pm 0.0941$ & $0.7043 \pm 0.0840$ \\
\bottomrule
\end{tabular}
\end{table}

\begin{table}[H]
\centering
\small
\caption{Attention concentration on Waterbirds validation set.}
\label{tab:attention_entropy_waterbirds}
\begin{tabular}{@{}lccc@{}}
\toprule
\textbf{Model} & \textbf{Entropy (norm)} & \textbf{Top-10\%} & \textbf{Top-25\%} \\
\midrule
ViT-S (21M) & $\mathbf{0.8763 \pm 0.0303}$ & $\mathbf{0.6742 \pm 0.1173}$ & $\mathbf{0.8742 \pm 0.0612}$ \\
ViT-B (86M) & $0.8980 \pm 0.0245$ & $0.6089 \pm 0.1049$ & $0.8246 \pm 0.0606$ \\
ViT-L (300M) & $0.9187 \pm 0.0271$ & $0.5195 \pm 0.1119$ & $0.7636 \pm 0.0850$ \\
ViT-G (1.1B) & $0.9307 \pm 0.0196$ & $0.4738 \pm 0.0842$ & $0.7234 \pm 0.0688$ \\
\bottomrule
\end{tabular}
\end{table}

\begin{table}[H]
\centering
\small
\caption{Different attention head aggregation methods on Spawrious O2O Hard (cross-model ViT-G attention and ViT-S embeddings). We report $A^2$ crops-only OOD test accuracy, where crops are chosen using the attention map resulting when using the following aggregation method (on the last-layer self-attention across all $24$ ViT-G heads).}
\label{tab:agg_ablation_spawrious_o2o}
\begin{tabular}{lc}
\toprule
\textbf{Aggregation Method} & \textbf{$A^2$ Test OOD (\%)} \\
\midrule
Mean & \textbf{86.3} \\
Entropy-weighted & \textbf{86.3} \\
Lowest-k mean (k=6) & 85.7 \\
Max & 85.0 \\
\bottomrule
\end{tabular}
\end{table}

\begin{table}[H]
\centering
\small
\caption{Attention concentration on Spawrious O2O Hard test split for ViT-G using different head aggregation methods}
\label{tab:agg_ablation_spawrious_concentration}
\begin{tabular}{lccc}
\toprule
\textbf{Aggregation Method} & \textbf{Entropy (norm)} & \textbf{Top-10\%} & \textbf{Top-25\%} \\
\midrule
Mean & $0.9202 \pm 0.0210$ & $0.5029 \pm 0.0914$ & $0.7698 \pm 0.0717$ \\
Entropy-weighted & $0.9170 \pm 0.0224$ & $0.5142 \pm 0.0950$ & $0.7780 \pm 0.0728$ \\
Lowest-entropy $k=6$ mean & $\mathbf{0.8561 \pm 0.0338}$ & $\mathbf{0.7185 \pm 0.1056}$ & $\mathbf{0.9111 \pm 0.0513}$ \\
Max & $0.9000 \pm 0.0268$ & $0.5817 \pm 0.0969$ & $0.8117 \pm 0.0680$ \\
\bottomrule
\end{tabular}
\end{table}

\begin{table}[H]
\centering
\small
\caption{Attention concentration on ImageNet (50k validation split) for ViT-G using different head aggregation methods}
\label{tab:agg_ablation_imagenet_vitg}
\begin{tabular}{lccc}
\toprule
\textbf{Aggregation Method} & \textbf{Entropy (norm)} & \textbf{Top-10\%} & \textbf{Top-25\%} \\
\midrule
Mean & $0.9351 \pm 0.0226$ & $0.4522 \pm 0.0941$ & $0.7043 \pm 0.0840$ \\
Entropy-weighted & $0.9325 \pm 0.0241$ & $0.4619 \pm 0.0983$ & $0.7117 \pm 0.0857$ \\
Lowest-entropy $k=6$ mean & $\mathbf{0.8731 \pm 0.0411}$ & $\mathbf{0.6624 \pm 0.1271}$ & $\mathbf{0.8654 \pm 0.0764}$ \\
Max & $0.9218 \pm 0.0273$ & $0.5050 \pm 0.0998$ & $0.7383 \pm 0.0794$ \\
\bottomrule
\end{tabular}
\end{table}

\section{Adapter Learning Curves}

\begin{figure}[H]
    \centering
    \begin{tikzpicture}[font=\fontsize{8}{9.5}\selectfont]
\begin{groupplot}[
  group style={group size=2 by 1, horizontal sep=0.45in},
  scale only axis, width=2.15in, height=1.15in,
  xmin=0.19999999999999996, xmax=20.8, xtick={1,5,10,15,20}, xlabel={Epoch}, xlabel style={yshift=2pt},
  axis lines*=left, axis line style={line width=0.5pt, -},
  tick align=outside, major tick length=2pt, minor tick length=0pt,
  tick style={line width=0.5pt, black},
  tick label style={font=\fontsize{8}{9.5}\selectfont, /pgf/number format/assume math mode=true},
  label style={font=\fontsize{8}{9.5}\selectfont},
  title style={font=\fontsize{8}{9.5}\selectfont\bfseries, at={(0,1)}, anchor=south west, inner sep=0pt, yshift=4pt},
  clip=false,
]
\nextgroupplot[title={Test accuracy}, ylabel={Test accuracy (\%)}, ylabel style={yshift=-2pt}, ymin=77.64, ymax=91.90, ytick={78,80,82,84,86,88,90},
  legend to name=adaptercurves,
  legend style={draw=none, fill=none, legend columns=2, font=\fontsize{8}{9.5}\selectfont, /tikz/every even column/.append style={column sep=9pt}}]
\addlegendimage{color=accent, line width=1.1pt, mark=*, mark size=1.7pt, mark options={fill=accent, draw=accent}}\addlegendentry{\textbf{$A^2$ with MLP adapter}}
\addlegendimage{color=refgray, line width=0.8pt, dash pattern=on 2pt off 2pt}\addlegendentry{default attention}
\addplot[color=refgray, line width=0.8pt, dash pattern=on 2pt off 2pt, forget plot] coordinates {(0.4,78.4085) (20.6,78.4085)};
\addplot[color=accent, line width=1.1pt, mark=*, mark size=1.7pt, mark options={fill=accent, draw=accent}, error bars/.cd, y dir=both, y explicit, error bar style={line width=0.5pt, accent}, error mark options={rotate=90, line width=0.5pt, mark size=1.25pt, accent}] table[x=epoch, y=acc_mean, y error=acc_se, col sep=comma] {Figures/adapter_curves/curves.csv};
\node[anchor=south east, inner sep=1pt] at (rel axis cs:0.98,0.11) {mean $\pm$ s.e., 5 seeds};
\nextgroupplot[title={Worst-group accuracy}, ylabel={Worst-group accuracy (\%)}, ylabel style={yshift=-2pt}, ymin=20.65, ymax=70.77, ytick={30,40,50,60,70}]
\addplot[color=refgray, line width=0.8pt, dash pattern=on 2pt off 2pt, forget plot] coordinates {(0.4,23.3333) (20.6,23.3333)};
\addplot[color=accent, line width=1.1pt, mark=*, mark size=1.7pt, mark options={fill=accent, draw=accent}, error bars/.cd, y dir=both, y explicit, error bar style={line width=0.5pt, accent}, error mark options={rotate=90, line width=0.5pt, mark size=1.25pt, accent}] table[x=epoch, y=wga_mean, y error=wga_se, col sep=comma] {Figures/adapter_curves/curves.csv};
\node[anchor=south east, inner sep=1pt] at (rel axis cs:0.98,0.11) {mean $\pm$ s.e., 5 seeds};
\end{groupplot}
\node[anchor=south] at ([xshift=0.225in, yshift=15pt]group c1r1.north east) {\pgfplotslegendfromname{adaptercurves}};
\end{tikzpicture}%
    \caption{MLP Adapter Learning curves on CelebA. Results averaged across 5 seeds.}
    \label{fig:mlp_adapter_learning_curves_celebA}
\end{figure}
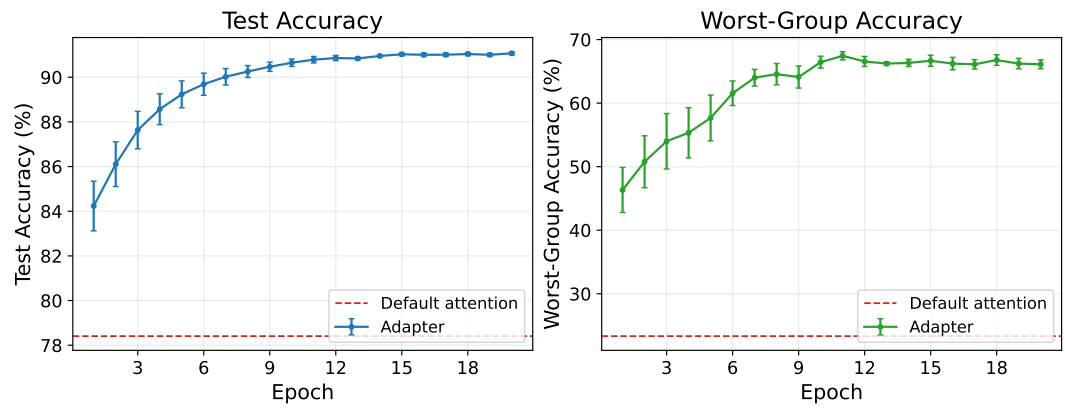

\section{DINOv3 Attention Scaling Graphs and Results}

DINOv3 Attention Scaling. The Attention model is varied from DINOv3 ViT-S/16 (21M) to DINOv3 ViT-7B/16 (6.7B) with the embedding model fixed to DINOv2 ViT-G (1.1B). We show the general trend of decreasing $A^2$ performance (Test Acc and WGA) as the attention model size grows also holds for DINOv3.

\begin{table}[H]
\centering
\small
\caption{Mean $A^2_{\text{LR}}$ performance across all evaluated datasets, ordered from smallest to largest DINOv3 attention model.}
\label{tab:dinov3_where_to_look_mean}
\begin{tabular}{lccc}
\toprule
Attention model & Params & Test Acc (\%) & WGA (\%) \\
\midrule
ViT-S/16 & 21M & 86.76 & 76.16 \\
ViT-S+/16 & 29M & 86.38 & 75.40 \\
ViT-B/16 & 86M & 86.66 & 75.52 \\
ViT-L/16 & 300M & 84.98 & 72.12 \\
ViT-H+/16 & 840M & 84.38 & 72.44 \\
ViT-7B/16 & 6{,}716M & 84.76 & 73.38 \\
\bottomrule
\end{tabular}
\end{table}

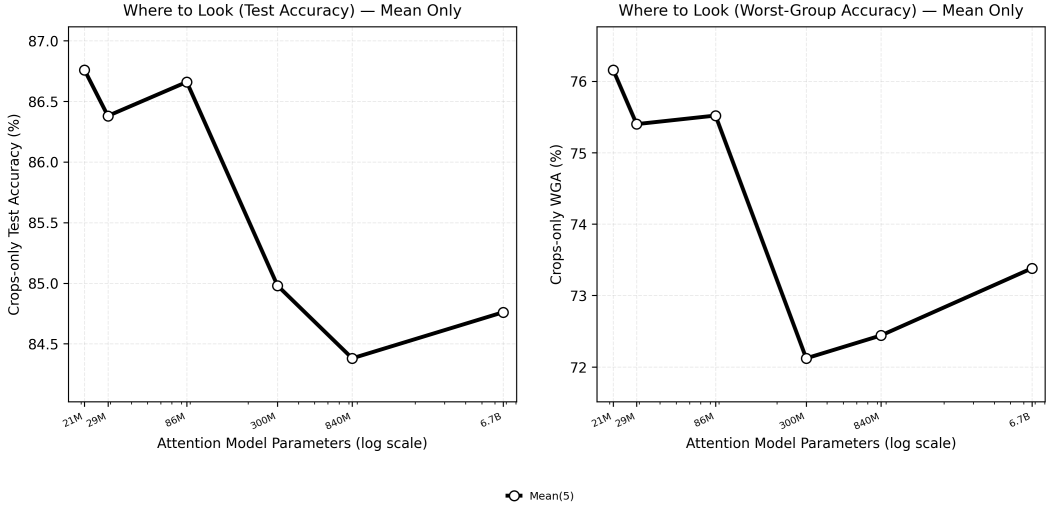
\begin{figure}[H]
\centering
\definecolor{dsSky}{HTML}{1C90DD}%
\definecolor{dsIndigo}{HTML}{4B4589}%
\definecolor{dsGreen}{HTML}{168E4D}%
\definecolor{dsOrchid}{HTML}{9F57A5}%
\definecolor{dsSienna}{HTML}{844500}%
\begin{tikzpicture}[font=\fontsize{8}{9.5}\selectfont]
\begin{groupplot}[
  group style={group size=2 by 1, horizontal sep=0.50in},
  scale only axis, width=2.10in, height=1.15in,
  xmode=log, xmin=15, xmax=9500, xtick={21,29,86,300,840,6716}, xticklabels={21M,,86M,300M,840M,6.7B},
  xlabel={DINOv3 selector size (embedder: DINOv2 ViT-G)}, xlabel style={yshift=2pt},
  axis lines*=left, axis line style={line width=0.5pt, -},
  tick align=outside, major tick length=2pt, minor tick length=0pt,
  tick style={line width=0.5pt, black},
  tick label style={font=\fontsize{8}{9.5}\selectfont, /pgf/number format/assume math mode=true},
  label style={font=\fontsize{8}{9.5}\selectfont},
  title style={font=\fontsize{8}{9.5}\selectfont\bfseries, at={(0,1)}, anchor=south west, inner sep=0pt, yshift=4pt},
  clip=false,
]
\nextgroupplot[title={Where to look (test accuracy)}, ylabel={Crops-only test accuracy (\%)}, ylabel style={yshift=-2pt}, ymin=84.19, ymax=86.95, ytick={84.5,85,85.5,86,86.5}, yticklabels={84.5,85.0,85.5,86.0,86.5}]

\addplot[color=accent, line width=1.2pt, mark=*, mark size=2.4pt, mark options={fill=accent, draw=accent, line width=0pt}] table[x=params, y=mean_acc, col sep=comma] {Figures/dinov3_where/values.csv};
\node[anchor=north east, inner sep=1pt] at (rel axis cs:0.98,0.98) {mean over the five datasets};
\nextgroupplot[title={Where to look (worst-group accuracy)}, ylabel={Crops-only WGA (\%)}, ylabel style={yshift=-2pt}, ymin=71.80, ymax=76.48, ytick={72,73,74,75,76}, yticklabels={72,73,74,75,76}]
\addplot[color=accent, line width=1.2pt, mark=*, mark size=2.4pt, mark options={fill=accent, draw=accent, line width=0pt}] table[x=params, y=mean_wga, col sep=comma] {Figures/dinov3_where/values.csv};
\node[anchor=north east, inner sep=1pt] at (rel axis cs:0.98,0.98) {mean over the five datasets};
\end{groupplot}

\end{tikzpicture}%
\caption{DINOv3 where-to-look scaling (mean across datasets) with fixed DINOv2 ViT-G embeddings. Absolute values are reported in Table~\ref{tab:dinov3_where_to_look_mean}.}
\label{fig:dinov3_where_to_look_scaling_mean}
\end{figure}
\section{Additional DINOv3 Where-to-Look Scaling Results}
\label{sec:supp_dinov3_where_to_look}

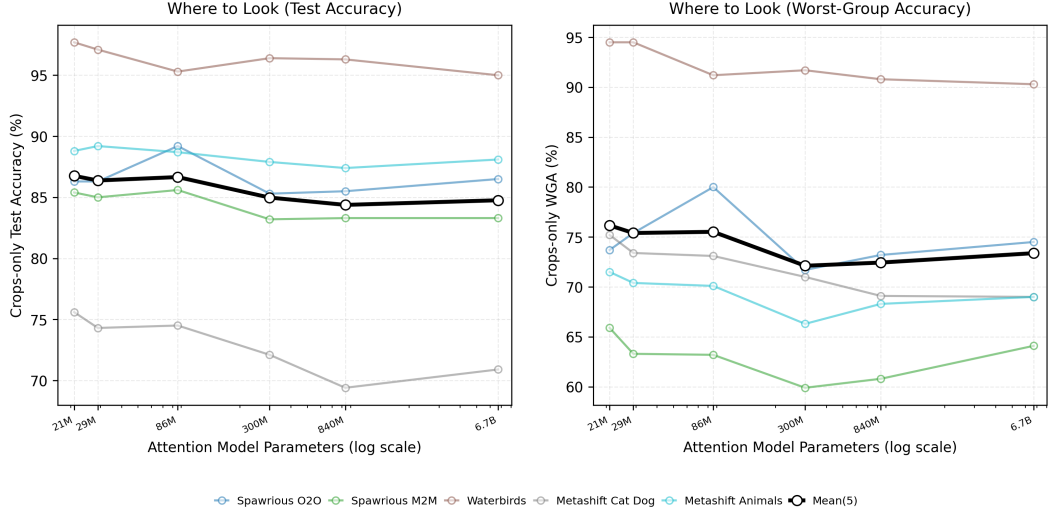
\begin{figure}[H]
\centering
\definecolor{dsSky}{HTML}{1C90DD}%
\definecolor{dsIndigo}{HTML}{4B4589}%
\definecolor{dsGreen}{HTML}{168E4D}%
\definecolor{dsOrchid}{HTML}{9F57A5}%
\definecolor{dsSienna}{HTML}{844500}%
\begin{tikzpicture}[font=\fontsize{8}{9.5}\selectfont]
\begin{groupplot}[
  group style={group size=2 by 1, horizontal sep=0.50in},
  scale only axis, width=2.10in, height=1.15in,
  xmode=log, xmin=15, xmax=9500, xtick={21,29,86,300,840,6716}, xticklabels={21M,,86M,300M,840M,6.7B},
  xlabel={DINOv3 selector size (embedder: DINOv2 ViT-G)}, xlabel style={yshift=2pt},
  axis lines*=left, axis line style={line width=0.5pt, -},
  tick align=outside, major tick length=2pt, minor tick length=0pt,
  tick style={line width=0.5pt, black},
  tick label style={font=\fontsize{8}{9.5}\selectfont, /pgf/number format/assume math mode=true},
  label style={font=\fontsize{8}{9.5}\selectfont},
  title style={font=\fontsize{8}{9.5}\selectfont\bfseries, at={(0,1)}, anchor=south west, inner sep=0pt, yshift=4pt},
  clip=false,
]
\nextgroupplot[title={Where to look (test accuracy)}, ylabel={Crops-only test accuracy (\%)}, ylabel style={yshift=-2pt}, ymin=67.14, ymax=99.96, ytick={70,80,90}, yticklabels={70,80,90},
  legend to name=dinov3-full,
  legend style={draw=none, fill=none, legend columns=6, font=\fontsize{8}{9.5}\selectfont,
    /tikz/every even column/.append style={column sep=6pt}},
  legend image code/.code={\draw[mark repeat=2, mark phase=2] plot coordinates {(0cm,0cm) (0.25cm,0cm) (0.5cm,0cm)};}]
\addlegendimage{color=accent, line width=1.2pt, mark=*, mark size=2.4pt, mark options={fill=accent, draw=accent, line width=0pt}}\addlegendentry{\textbf{Mean}}
\addlegendimage{color=dsSky, line width=0.8pt, mark=*, mark size=1.9pt, mark options={fill=dsSky!35, draw=dsSky, line width=0.6pt}}\addlegendentry{Spawrious O2O}
\addlegendimage{color=dsIndigo, line width=0.8pt, mark=square*, mark size=1.9pt, mark options={fill=dsIndigo!35, draw=dsIndigo, line width=0.6pt}}\addlegendentry{Spawrious M2M}
\addlegendimage{color=dsGreen, line width=0.8pt, mark=triangle*, mark size=1.9pt, mark options={fill=dsGreen!35, draw=dsGreen, line width=0.6pt}}\addlegendentry{Waterbirds}
\addlegendimage{color=dsOrchid, line width=0.8pt, mark=diamond*, mark size=1.9pt, mark options={fill=dsOrchid!35, draw=dsOrchid, line width=0.6pt}}\addlegendentry{Cat vs.\ Dog}
\addlegendimage{color=dsSienna, line width=0.8pt, mark=pentagon*, mark size=1.9pt, mark options={fill=dsSienna!35, draw=dsSienna, line width=0.6pt}}\addlegendentry{Animals}
\addplot[color=dsSky, line width=0.8pt, mark=*, mark size=1.9pt, mark options={fill=dsSky!35, draw=dsSky, line width=0.6pt}] table[x=params, y=o2o_acc, col sep=comma] {Figures/dinov3_where/values.csv};
\addplot[color=dsIndigo, line width=0.8pt, mark=square*, mark size=1.9pt, mark options={fill=dsIndigo!35, draw=dsIndigo, line width=0.6pt}] table[x=params, y=m2m_acc, col sep=comma] {Figures/dinov3_where/values.csv};
\addplot[color=dsGreen, line width=0.8pt, mark=triangle*, mark size=1.9pt, mark options={fill=dsGreen!35, draw=dsGreen, line width=0.6pt}] table[x=params, y=wb_acc, col sep=comma] {Figures/dinov3_where/values.csv};
\addplot[color=dsOrchid, line width=0.8pt, mark=diamond*, mark size=1.9pt, mark options={fill=dsOrchid!35, draw=dsOrchid, line width=0.6pt}] table[x=params, y=catdog_acc, col sep=comma] {Figures/dinov3_where/values.csv};
\addplot[color=dsSienna, line width=0.8pt, mark=pentagon*, mark size=1.9pt, mark options={fill=dsSienna!35, draw=dsSienna, line width=0.6pt}] table[x=params, y=animals_acc, col sep=comma] {Figures/dinov3_where/values.csv};
\addplot[color=accent, line width=1.2pt, mark=*, mark size=2.4pt, mark options={fill=accent, draw=accent, line width=0pt}] table[x=params, y=mean_acc, col sep=comma] {Figures/dinov3_where/values.csv};
\nextgroupplot[title={Where to look (worst-group accuracy)}, ylabel={Crops-only WGA (\%)}, ylabel style={yshift=-2pt}, ymin=57.13, ymax=97.27, ytick={60,70,80,90}, yticklabels={60,70,80,90}]
\addplot[color=dsSky, line width=0.8pt, mark=*, mark size=1.9pt, mark options={fill=dsSky!35, draw=dsSky, line width=0.6pt}] table[x=params, y=o2o_wga, col sep=comma] {Figures/dinov3_where/values.csv};
\addplot[color=dsIndigo, line width=0.8pt, mark=square*, mark size=1.9pt, mark options={fill=dsIndigo!35, draw=dsIndigo, line width=0.6pt}] table[x=params, y=m2m_wga, col sep=comma] {Figures/dinov3_where/values.csv};
\addplot[color=dsGreen, line width=0.8pt, mark=triangle*, mark size=1.9pt, mark options={fill=dsGreen!35, draw=dsGreen, line width=0.6pt}] table[x=params, y=wb_wga, col sep=comma] {Figures/dinov3_where/values.csv};
\addplot[color=dsOrchid, line width=0.8pt, mark=diamond*, mark size=1.9pt, mark options={fill=dsOrchid!35, draw=dsOrchid, line width=0.6pt}] table[x=params, y=catdog_wga, col sep=comma] {Figures/dinov3_where/values.csv};
\addplot[color=dsSienna, line width=0.8pt, mark=pentagon*, mark size=1.9pt, mark options={fill=dsSienna!35, draw=dsSienna, line width=0.6pt}] table[x=params, y=animals_wga, col sep=comma] {Figures/dinov3_where/values.csv};
\addplot[color=accent, line width=1.2pt, mark=*, mark size=2.4pt, mark options={fill=accent, draw=accent, line width=0pt}] table[x=params, y=mean_wga, col sep=comma] {Figures/dinov3_where/values.csv};
\end{groupplot}
\node[anchor=south] at ([xshift=0.25in, yshift=15pt]group c1r1.north east) {\pgfplotslegendfromname{dinov3-full}};
\end{tikzpicture}%
\caption{DINOv3 where-to-look scaling per dataset (colored lines) with the mean trend (red), using fixed DINOv2 ViT-G embeddings. Per-dataset absolute values in Tables~\ref{tab:dinov3_where_to_look_metashift_animals}, \ref{tab:dinov3_where_to_look_metashift_catdog}, \ref{tab:dinov3_where_to_look_spawrious_o2o}, \ref{tab:dinov3_where_to_look_spawrious_m2m}, and \ref{tab:dinov3_where_to_look_waterbirds}.}
\label{fig:dinov3_where_to_look_scaling_full}
\end{figure}

\subsection{Per-dataset DINOv3 Where-to-Look Results}

\begin{table}[H]
\centering
\small
\caption{MetaShift Animals ($A^2_{\text{LR}}$) with fixed DINOv2 ViT-G embeddings.}
\label{tab:dinov3_where_to_look_metashift_animals}
\begin{tabular}{lccc}
\toprule
Attention model & Params & Test Acc (\%) & WC (\%) \\
\midrule
ViT-S/16 & 21M & 88.8 & 71.5 \\
ViT-S+/16 & 29M & 89.2 & 70.4 \\
ViT-B/16 & 86M & 88.7 & 70.1 \\
ViT-L/16 & 300M & 87.9 & 66.3 \\
ViT-H+/16 & 840M & 87.4 & 68.3 \\
ViT-7B/16 & 6{,}716M & 88.1 & 69.0 \\
\bottomrule
\end{tabular}
\end{table}

\begin{table}[H]
\centering
\small
\caption{MetaShift Cat vs.\ Dog ($A^2_{\text{LR}}$) with fixed DINOv2 ViT-G embeddings.}
\label{tab:dinov3_where_to_look_metashift_catdog}
\begin{tabular}{lccc}
\toprule
Attention model & Params & Test Acc (\%) & WGA (\%) \\
\midrule
ViT-S/16 & 21M & 75.6 & 75.2 \\
ViT-S+/16 & 29M & 74.3 & 73.4 \\
ViT-B/16 & 86M & 74.5 & 73.1 \\
ViT-L/16 & 300M & 72.1 & 71.0 \\
ViT-H+/16 & 840M & 69.4 & 69.1 \\
ViT-7B/16 & 6{,}716M & 70.9 & 69.0 \\
\bottomrule
\end{tabular}
\end{table}

\begin{table}[H]
\centering
\small
\caption{Spawrious O2O Hard ($A^2_{\text{LR}}$) with fixed DINOv2 ViT-G embeddings.}
\label{tab:dinov3_where_to_look_spawrious_o2o}
\begin{tabular}{lccc}
\toprule
Attention model & Params & Test Acc (\%) & WGA (\%) \\
\midrule
ViT-S/16 & 21M & 86.3 & 73.7 \\
ViT-S+/16 & 29M & 86.3 & 75.4 \\
ViT-B/16 & 86M & 89.2 & 80.0 \\
ViT-L/16 & 300M & 85.3 & 71.7 \\
ViT-H+/16 & 840M & 85.5 & 73.2 \\
ViT-7B/16 & 6{,}716M & 86.5 & 74.5 \\
\bottomrule
\end{tabular}
\end{table}

\begin{table}[H]
\centering
\small
\caption{Spawrious M2M Hard ($A^2_{\text{LR}}$) with fixed DINOv2 ViT-G embeddings.}
\label{tab:dinov3_where_to_look_spawrious_m2m}
\begin{tabular}{lccc}
\toprule
Attention model & Params & Test Acc (\%) & WGA (\%) \\
\midrule
ViT-S/16 & 21M & 85.4 & 65.9 \\
ViT-S+/16 & 29M & 85.0 & 63.3 \\
ViT-B/16 & 86M & 85.6 & 63.2 \\
ViT-L/16 & 300M & 83.2 & 59.9 \\
ViT-H+/16 & 840M & 83.3 & 60.8 \\
ViT-7B/16 & 6{,}716M & 83.3 & 64.1 \\
\bottomrule
\end{tabular}
\end{table}

\begin{table}[H]
\centering
\small
\caption{Waterbirds ($A^2_{\text{LR}}$) with fixed DINOv2 ViT-G embeddings.}
\label{tab:dinov3_where_to_look_waterbirds}
\begin{tabular}{lccc}
\toprule
Attention model & Params & Test Acc (\%) & WGA (\%) \\
\midrule
ViT-S/16 & 21M & 97.7 & 94.5 \\
ViT-S+/16 & 29M & 97.1 & 94.5 \\
ViT-B/16 & 86M & 95.3 & 91.2 \\
ViT-L/16 & 300M & 96.4 & 91.7 \\
ViT-H+/16 & 840M & 96.3 & 90.8 \\
ViT-7B/16 & 6{,}716M & 95.0 & 90.3 \\
\bottomrule
\end{tabular}
\end{table}

\section{OpenCLIP ZS Family Scaling: Full Results}
\label{sec:clip_zs_family_scaling_wga}

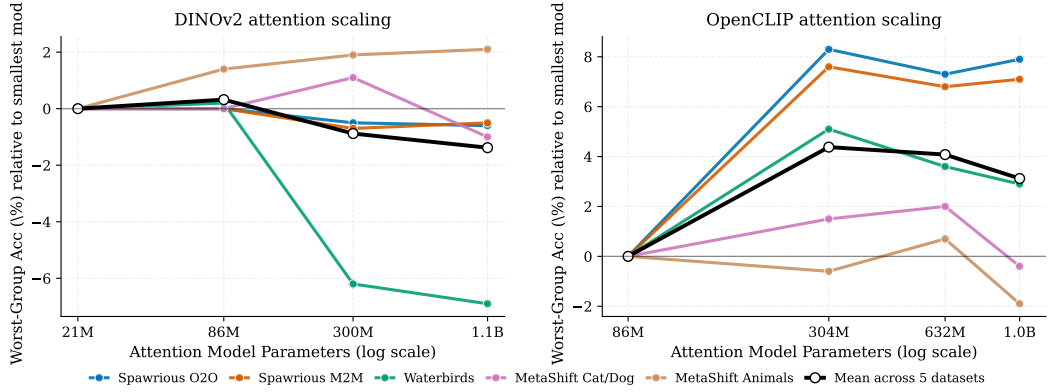
\begin{figure}[H]
\centering
\definecolor{dsSky}{HTML}{1C90DD}%
\definecolor{dsIndigo}{HTML}{4B4589}%
\definecolor{dsGreen}{HTML}{168E4D}%
\definecolor{dsOrchid}{HTML}{9F57A5}%
\definecolor{dsSienna}{HTML}{844500}%
\begin{tikzpicture}[font=\fontsize{8}{9.5}\selectfont]
\begin{groupplot}[
  group style={group size=2 by 1, horizontal sep=0.42in},
  scale only axis, width=2.06in, height=1.05in,
  xmode=log, axis lines*=left, axis line style={line width=0.5pt, -},
  tick align=outside, major tick length=2pt, minor tick length=0pt,
  tick style={line width=0.5pt, black},
  tick label style={font=\fontsize{8}{9.5}\selectfont, /pgf/number format/assume math mode=true},
  label style={font=\fontsize{8}{9.5}\selectfont}, xlabel style={yshift=2pt},
  title style={font=\fontsize{8}{9.5}\selectfont\bfseries, at={(0,1)}, anchor=south west, inner sep=0pt, yshift=4pt},
  clip=false,
]
\nextgroupplot[title={DINOv2 selectors}, xlabel={Selector size (embedder: OpenCLIP ViT-L/14)}, xmin=15.6, xmax=1485.0, xtick={21,86,300,1100}, xticklabels={21M,86M,300M,1.1B}, ymin=-7.62, ymax=2.82, ytick={-6,-4,-2,0,2}, yticklabels={\textminus6,\textminus4,\textminus2,0,+2}, ylabel={WGA change (points)}, ylabel style={yshift=-2pt},
  legend to name=rellegend-clipzs_wga,
  legend style={draw=none, fill=none, legend columns=6, font=\fontsize{8}{9.5}\selectfont,
    /tikz/every even column/.append style={column sep=6pt}},
  legend image code/.code={\draw[mark repeat=2, mark phase=2] plot coordinates {(0cm,0cm) (0.25cm,0cm) (0.5cm,0cm)};}]
\addlegendimage{color=accent, line width=1.2pt, mark=*, mark size=2.4pt, mark options={fill=accent, draw=accent, line width=0pt}}\addlegendentry{\textbf{Mean}}
\addlegendimage{color=dsSky, line width=0.8pt, mark=*, mark size=1.9pt, mark options={fill=dsSky!35, draw=dsSky, line width=0.6pt}}\addlegendentry{Spawrious O2O}
\addlegendimage{color=dsIndigo, line width=0.8pt, mark=square*, mark size=1.9pt, mark options={fill=dsIndigo!35, draw=dsIndigo, line width=0.6pt}}\addlegendentry{Spawrious M2M}
\addlegendimage{color=dsGreen, line width=0.8pt, mark=triangle*, mark size=1.9pt, mark options={fill=dsGreen!35, draw=dsGreen, line width=0.6pt}}\addlegendentry{Waterbirds}
\addlegendimage{color=dsOrchid, line width=0.8pt, mark=diamond*, mark size=1.9pt, mark options={fill=dsOrchid!35, draw=dsOrchid, line width=0.6pt}}\addlegendentry{Cat vs.\ Dog}
\addlegendimage{color=dsSienna, line width=0.8pt, mark=pentagon*, mark size=1.9pt, mark options={fill=dsSienna!35, draw=dsSienna, line width=0.6pt}}\addlegendentry{Animals}
\addplot[refgray, line width=0.5pt, forget plot] coordinates {(15.6,0) (1485.0,0)};
\addplot[color=dsSky, line width=0.8pt, mark=*, mark size=1.9pt, mark options={fill=dsSky!35, draw=dsSky, line width=0.6pt}] table[x=params, y=o2o, col sep=comma] {Figures/relative_scaling/clipzs_wga_left.csv};
\addplot[color=dsIndigo, line width=0.8pt, mark=square*, mark size=1.9pt, mark options={fill=dsIndigo!35, draw=dsIndigo, line width=0.6pt}] table[x=params, y=m2m, col sep=comma] {Figures/relative_scaling/clipzs_wga_left.csv};
\addplot[color=dsGreen, line width=0.8pt, mark=triangle*, mark size=1.9pt, mark options={fill=dsGreen!35, draw=dsGreen, line width=0.6pt}] table[x=params, y=wb, col sep=comma] {Figures/relative_scaling/clipzs_wga_left.csv};
\addplot[color=dsOrchid, line width=0.8pt, mark=diamond*, mark size=1.9pt, mark options={fill=dsOrchid!35, draw=dsOrchid, line width=0.6pt}] table[x=params, y=catdog, col sep=comma] {Figures/relative_scaling/clipzs_wga_left.csv};
\addplot[color=dsSienna, line width=0.8pt, mark=pentagon*, mark size=1.9pt, mark options={fill=dsSienna!35, draw=dsSienna, line width=0.6pt}] table[x=params, y=animals, col sep=comma] {Figures/relative_scaling/clipzs_wga_left.csv};
\addplot[color=accent, line width=1.2pt, mark=*, mark size=2.4pt, mark options={fill=accent, draw=accent, line width=0pt}] table[x=params, y=mean, col sep=comma] {Figures/relative_scaling/clipzs_wga_left.csv};
\node[anchor=west, xshift=3pt, fill=white, inner sep=1pt] at (axis cs:1100,-1.38) {\textminus1.4};
\nextgroupplot[title={OpenCLIP selectors}, xlabel={Selector size (embedder: OpenCLIP ViT-L/14)}, xmin=63.7, xmax=1350.0, xtick={86,304,632,1000}, xticklabels={86M,304M,632M,1.0B}, ymin=-2.72, ymax=9.12, ytick={-2,0,2,4,6,8}, yticklabels={\textminus2,0,+2,+4,+6,+8}]
\addplot[refgray, line width=0.5pt, forget plot] coordinates {(63.7,0) (1350.0,0)};
\addplot[color=dsSky, line width=0.8pt, mark=*, mark size=1.9pt, mark options={fill=dsSky!35, draw=dsSky, line width=0.6pt}] table[x=params, y=o2o, col sep=comma] {Figures/relative_scaling/clipzs_wga_right.csv};
\addplot[color=dsIndigo, line width=0.8pt, mark=square*, mark size=1.9pt, mark options={fill=dsIndigo!35, draw=dsIndigo, line width=0.6pt}] table[x=params, y=m2m, col sep=comma] {Figures/relative_scaling/clipzs_wga_right.csv};
\addplot[color=dsGreen, line width=0.8pt, mark=triangle*, mark size=1.9pt, mark options={fill=dsGreen!35, draw=dsGreen, line width=0.6pt}] table[x=params, y=wb, col sep=comma] {Figures/relative_scaling/clipzs_wga_right.csv};
\addplot[color=dsOrchid, line width=0.8pt, mark=diamond*, mark size=1.9pt, mark options={fill=dsOrchid!35, draw=dsOrchid, line width=0.6pt}] table[x=params, y=catdog, col sep=comma] {Figures/relative_scaling/clipzs_wga_right.csv};
\addplot[color=dsSienna, line width=0.8pt, mark=pentagon*, mark size=1.9pt, mark options={fill=dsSienna!35, draw=dsSienna, line width=0.6pt}] table[x=params, y=animals, col sep=comma] {Figures/relative_scaling/clipzs_wga_right.csv};
\addplot[color=accent, line width=1.2pt, mark=*, mark size=2.4pt, mark options={fill=accent, draw=accent, line width=0pt}] table[x=params, y=mean, col sep=comma] {Figures/relative_scaling/clipzs_wga_right.csv};
\node[anchor=west, xshift=3pt, fill=white, inner sep=1pt] at (axis cs:1000,3.12) {+3.1};
\end{groupplot}
\node[anchor=south] at ([xshift=0.21in, yshift=15pt]group c1r1.north east) {\pgfplotslegendfromname{rellegend-clipzs_wga}};
\end{tikzpicture}%
\caption{Worst-group accuracy version of Figure~\ref{fig:clip_zs_attn_scaling}, also plotted relative to the smallest model in each family. The same family contrast holds: DINOv2 attention shows a mild downward WGA trend with size, while OpenCLIP attention shows large positive offsets at ViT-L/14 and larger (Section~\ref{sec:attention_iou}). Absolute per-dataset values are reported in Table~\ref{tab:clip_zs_attn_scaling}.}
\label{fig:clip_zs_family_scaling_wga}
\end{figure}

\begin{table}[H]
\centering
\small
\caption{OpenCLIP ViT-L-14 zero-shot classification on $A^2$ attention-guided $128\times128$ crops. The rows represent the model used to compute attention (and select crops). The backbone used to embed crops and class text labels is fixed to OpenCLIP ViT-L-14. DINOv2 attention shows inverse scaling (smaller is better); OpenCLIP attention is non-monotonic (ViT-B-16 underperforms despite being the smallest), this finding mirrors the localization finding from Section~\ref{sec:attention_iou}. For OpenCLIP, we report the number of parameters in the vision encoder.}
\label{tab:clip_zs_attn_scaling}
\setlength{\tabcolsep}{4pt}
\begin{tabular}{@{}llcccccccccc@{}}
\toprule
& & \multicolumn{2}{c}{\textbf{O2O-Hard}} & \multicolumn{2}{c}{\textbf{M2M-Hard}} & \multicolumn{2}{c}{\textbf{Waterbirds}} & \multicolumn{2}{c}{\textbf{Cat vs.\ Dog}} & \multicolumn{2}{c}{\textbf{Animals}} \\
\textbf{Attn Family} & \textbf{Attn Model} & Acc & WGA & Acc & WGA & Acc & WGA & Acc & WGA & Acc & WC \\
\midrule
\multicolumn{2}{l}{\textbf{Full image (no crops)}} & 89.8 & 82.0 & 92.4 & 82.1 & 73.5 & 46.9 & 92.4 & 88.8 & 90.4 & 58.8 \\
\addlinespace[3pt]
\multirow{4}{*}{\textbf{DINOv2}} & ViT-S(21M) & \textbf{92.8} & \textbf{85.0} & \textbf{94.0} & \textbf{85.4} & \textbf{80.3} & 64.5 & 89.7 & 88.8 & \textbf{89.7} & 54.7 \\
 & ViT-B(86M) & 92.5 & \textbf{85.0} & \textbf{94.0} & \textbf{85.4} & 80.7 & \textbf{64.7} & \textbf{90.3} & 88.8 & 89.6 & 56.1 \\
 & ViT-L(300M) & 91.0 & 84.5 & 93.4 & 84.7 & 77.8 & 58.3 & 90.1 & \textbf{89.9} & 89.0 & 56.6 \\
 & ViT-G(1.1B) & 91.5 & 84.4 & 93.5 & 84.9 & 77.0 & 57.6 & 90.1 & 87.8 & 89.5 & 56.8 \\
\addlinespace[3pt]
\multirow{4}{*}{\textbf{OpenCLIP}} & ViT-B-16(86M) & 84.0 & 75.9 & 87.2 & 76.8 & 77.1 & 58.4 & 87.7 & 87.5 & 87.5 & 59.4 \\
 & ViT-L-14(304M) & 91.9 & 84.2 & 93.4 & 84.4 & 79.6 & 63.5 & 89.0 & 89.0 & 89.3 & 58.8 \\
 & ViT-H-14(632M) & 91.2 & 83.2 & 92.9 & 83.6 & 78.9 & 62.0 & 89.5 & 89.5 & 88.6 & \textbf{60.1} \\
 & ViT-G-14(1.01B) & 91.4 & 83.8 & 93.4 & 83.9 & 78.6 & 61.3 & 89.5 & 87.1 & 88.8 & 57.5 \\
\bottomrule
\end{tabular}
\end{table}

\section{TTR Implementation Details}
\label{sec:ttr}

We reimplement TTR \cite{ttr} based on the author's official codebase \url{https://github.com/ShengyuLu/TTR}. We use the OpenCLIP ViT-L-14 backbone (\texttt{laion2b\_s32b\_b82k}) for TTR and $A^2$ to ensure fair comparison. For Waterbirds, we were able to reproduce the reported results (achieving slightly higher WGA than was originally reported). We were unable to match the MetaShift Cat vs.\ Dog result, and report our best numbers instead.

\begin{table}[H]
\centering
\small
\caption{TTR hyperparameters used in our reimplementation}
\label{tab:ttr-hyperparams}
\begin{tabular}{lcccc}
\toprule
 & Waterbirds & \makecell{Spawrious\\O2O/M2M} & \makecell{MetaShift\\Cat vs.\ Dog} & \makecell{MetaShift\\Animals} \\
\midrule
Queue length       & 28   & 28   & 24   & 24   \\
Gaussian $\sigma$  & 0.72 & 0.72 & 0.71 & 0.71 \\
KMeans $K$         & 3    & 3    & 3    & 3    \\
PCA components     & 3    & 3    & 3    & 3    \\
SVD rank ($q$)     & 16   & 16   & 16   & 16   \\
\bottomrule
\end{tabular}%
\end{table}

Below we also include our class prompts and general prompts (to isolate the main subject in each image from the clusters)

\begin{table}[H]
\centering
\small
\caption{TTR text prompts used in our reimplementation.}
\label{tab:ttr-prompts}
\begin{tabular}{lll}
\toprule
Dataset & Class prompts & General label \\
\midrule
Waterbirds & ``a photo of a landbird.'', ``a photo of a waterbird.'' & ``a photo of a bird.'' \\
Spawrious O2O/M2M & ``a photo of a \{breed\}.'' for each breed & ``a photo of a dog.'' \\
MetaShift Cat vs.\ Dog & ``a photo of a cat.'', ``a photo of a dog.'' & ``a photo of a pet.'' \\
MetaShift Animals & ``a photo of a \{class\}.'' for each class & ``a photo of an animal.'' \\
\bottomrule
\end{tabular}%
\end{table}

\section{iFAM Comparison Runs}
\label{sec:ifam_protocol}

We re-trained iFAM \cite{ifam} on every benchmark in Tables~\ref{tab:main-wga} and~\ref{tab:main-acc} using the official code release with all default hyperparameters from the iFAM paper (its Appendix~A).

\textbf{Number of foreground parts $K$.} The iFAM paper reports $K{=}8$ as its Waterbirds default and $K{=}4$ as its MetaShift Cat vs.\ Dog default; the remaining benchmarks we evaluate (Spawrious O2O Hard, Spawrious M2M Hard, MetaShift Animals) are not in the iFAM paper, so no published default exists. To give iFAM the strongest comparison, we trained at both $K{=}4$ and $K{=}8$ on every benchmark and report both rows in Tables~\ref{tab:main-wga} and~\ref{tab:main-acc}.

Our Waterbirds numbers ($K{=}8$: $98.8$ accuracy, $96.1$ WGA) match the iFAM paper's reported Waterbirds results closely, which we take as evidence that our implementation is faithful; the same default hyperparameters are used across all five benchmarks (excluding $K$ which we varied).

\section{Comparing $A^2$ to using a peak-localization selector}

\begin{table}[H]
\centering
\caption{\textbf{ImageNet-calibrated layer transfer.} Changes in downstream worst-group accuracy, computed as the ImageNet-selected layer minus the last layer. MetaShift Animals reports worst-class accuracy. For each selector, the chosen layer maximizes mean patch-normalized attention mass within ImageNet validation bounding boxes and stays fixed across all five tasks. Layer indices are zero-indexed. Positive changes are bold. Values of $0.0$ are present when the last layer is the best attention localization layer on ImageNet validation.}
\label{tab:imagenet_layer_transfer}
\begingroup
\small
\setlength{\tabcolsep}{3pt}
\renewcommand{\arraystretch}{0.9}
\begin{tabular}{lcrrrrr}
\toprule
Selector & \shortstack{Terminal $\to$\\selected index} & Waterbirds & \shortstack{Spawrious\\O2O} & \shortstack{Spawrious\\M2M} & \shortstack{MetaShift\\CatDog} & \shortstack{MetaShift\\Animals} \\
\midrule
DINOv1-S & \(11\to 10\) & \(-0.47\) & \(-1.74\) & \(-2.62\) & \(-3.23\) & \(\mathbf{+1.55}\) \\
DINOv1-B & \(11\to 9\) & \(-3.12\) & \(\mathbf{+1.70}\) & \(\mathbf{+6.47}\) & \(-0.78\) & \(-2.38\) \\
\midrule
DINOv2-S & \(11\to 11\) & \(0.00\) & \(0.00\) & \(0.00\) & \(0.00\) & \(0.00\) \\
DINOv2-B & \(11\to 11\) & \(0.00\) & \(0.00\) & \(0.00\) & \(0.00\) & \(0.00\) \\
DINOv2-L & \(23\to 21\) & \(\mathbf{+8.72}\) & \(\mathbf{+2.16}\) & \(\mathbf{+8.05}\) & \(\mathbf{+3.88}\) & \(\mathbf{+0.61}\) \\
DINOv2-G & \(39\to 33\) & \(\mathbf{+2.96}\) & \(\mathbf{+2.89}\) & \(\mathbf{+6.12}\) & \(\mathbf{+4.01}\) & \(-2.15\) \\
\midrule
iBOT-S & \(11\to 9\) & \(-3.27\) & \(-1.50\) & \(\mathbf{+0.22}\) & \(\mathbf{+1.55}\) & \(\mathbf{+2.56}\) \\
iBOT-B & \(11\to 9\) & \(-0.93\) & \(\mathbf{+2.64}\) & \(\mathbf{+5.65}\) & \(\mathbf{+0.39}\) & \(\mathbf{+3.38}\) \\
iBOT-L & \(23\to 15\) & \(\mathbf{+2.96}\) & \(\mathbf{+5.57}\) & \(-1.48\) & \(\mathbf{+6.86}\) & \(-0.27\) \\
\midrule
MAE-B & \(11\to 11\) & \(0.00\) & \(0.00\) & \(0.00\) & \(0.00\) & \(0.00\) \\
MAE-L & \(23\to 15\) & \(\mathbf{+15.26}\) & \(-8.10\) & \(-3.72\) & \(\mathbf{+2.07}\) & \(\mathbf{+12.60}\) \\
MAE-H & \(31\to 18\) & \(\mathbf{+23.52}\) & \(\mathbf{+13.43}\) & \(\mathbf{+2.90}\) & \(\mathbf{+13.84}\) & \(\mathbf{+20.97}\) \\
\midrule
DINOv3-S & \(11\to 11\) & \(0.00\) & \(0.00\) & \(0.00\) & \(0.00\) & \(0.00\) \\
DINOv3-S+ & \(11\to 11\) & \(0.00\) & \(0.00\) & \(0.00\) & \(0.00\) & \(0.00\) \\
DINOv3-B & \(11\to 11\) & \(0.00\) & \(0.00\) & \(0.00\) & \(0.00\) & \(0.00\) \\
DINOv3-L & \(23\to 22\) & \(0.00\) & \(-1.17\) & \(\mathbf{+2.59}\) & \(\mathbf{+4.14}\) & \(-4.61\) \\
DINOv3-H+ & \(31\to 30\) & \(\mathbf{+3.58}\) & \(\mathbf{+2.87}\) & \(\mathbf{+1.89}\) & \(\mathbf{+0.52}\) & \(-4.91\) \\
DINOv3-7B & \(39\to 34\) & \(\mathbf{+1.40}\) & \(\mathbf{+2.07}\) & \(-0.38\) & \(\mathbf{+0.52}\) & \(\mathbf{+1.02}\) \\
\bottomrule
\end{tabular}%
\endgroup
\end{table}

\begin{table}[H]
\centering
\caption{
We compare $A^2$ performance (using the smallest model in a family to select crops and the largest to embed) with a classifier that instead selects crops using the layer that maximizes mean patch attention mass in ImageNet validation bounding boxes. Once calibrated on ImageNet validation, this layer is used throughout all five tasks. $A^2$ uses the last layer attention from the smallest model in the family, and the largest model in the family to embed those crops. Scores shown are worst group accuracy except for MetaShift Animals which is worst class accuracy. A positive delta favors $A^2$. $A^2$ outperforms in $16$ out of $25$ cases, showing that decoupled selection and embedding is still useful even after calibration. Layers are 0 indexed.}
\label{tab:a2_vs_peak_localization}
\begingroup
\small
\setlength{\tabcolsep}{4pt}
\renewcommand{\arraystretch}{1.08}
\begin{tabular}{lcrrrrr}
\toprule
Selector (model) & Layer & Waterbirds & \shortstack{Spawrious\\O2O} & \shortstack{Spawrious\\M2M} & \shortstack{MetaShift\\CatDog} & \shortstack{MetaShift\\Animals} \\
\midrule
\multicolumn{7}{l}{\textbf{DINOv1} (crop embedder: ViT-B/16)} \\
Largest at peak (B) & 9 & \(68.54\) & \(\mathbf{65.83}\) & \(\mathbf{37.75}\) & \(64.17\) & \(53.41\) \\
$A^2$ (S, terminal) & 11 & \(\mathbf{72.12}\) & \(64.84\) & \(37.37\) & \(\mathbf{66.49}\) & \(\mathbf{55.79}\) \\
$\Delta$: $A^2$ $-$ largest (pp) & -- & \(\mathbf{+3.58}\) & \(-0.99\) & \(-0.38\) & \(\mathbf{+2.33}\) & \(\mathbf{+2.38}\) \\
\midrule
\multicolumn{7}{l}{\textbf{DINOv2} (crop embedder: ViT-G/14)} \\
Largest at peak (G) & 33 & \(95.02\) & \(76.44\) & \(\mathbf{67.87}\) & \(\mathbf{77.75}\) & \(67.04\) \\
$A^2$ (S, terminal) & 11 & \(\mathbf{95.33}\) & \(\mathbf{80.10}\) & \(64.65\) & \(76.58\) & \(\mathbf{70.83}\) \\
$\Delta$: $A^2$ $-$ largest (pp) & -- & \(\mathbf{+0.31}\) & \(\mathbf{+3.66}\) & \(-3.22\) & \(-1.16\) & \(\mathbf{+3.79}\) \\
\midrule
\multicolumn{7}{l}{\textbf{iBOT} (crop embedder: ViT-L/16)} \\
Largest at peak (L) & 15 & \(73.36\) & \(64.05\) & \(27.02\) & \(66.36\) & \(62.85\) \\
$A^2$ (S, terminal) & 11 & \(\mathbf{76.17}\) & \(\mathbf{64.79}\) & \(\mathbf{28.47}\) & \(\mathbf{69.21}\) & \(\mathbf{63.15}\) \\
$\Delta$: $A^2$ $-$ largest (pp) & -- & \(\mathbf{+2.80}\) & \(\mathbf{+0.74}\) & \(\mathbf{+1.45}\) & \(\mathbf{+2.85}\) & \(\mathbf{+0.31}\) \\
\midrule
\multicolumn{7}{l}{\textbf{MAE} (crop embedder: ViT-H/14)} \\
Largest at peak (H) & 18 & \(\mathbf{39.56}\) & \(42.60\) & \(18.40\) & \(\mathbf{43.86}\) & \(\mathbf{29.65}\) \\
$A^2$ (B, terminal) & 11 & \(37.23\) & \(\mathbf{46.64}\) & \(\mathbf{21.02}\) & \(41.27\) & \(26.76\) \\
$\Delta$: $A^2$ $-$ largest (pp) & -- & \(-2.34\) & \(\mathbf{+4.04}\) & \(\mathbf{+2.62}\) & \(-2.59\) & \(-2.89\) \\
\midrule
\multicolumn{7}{l}{\textbf{DINOv3} (crop embedder: ViT-7B/16)} \\
Largest at peak (7B) & 34 & \(88.32\) & \(\mathbf{76.80}\) & \(64.49\) & \(59.90\) & \(\mathbf{76.77}\) \\
$A^2$ (S, terminal) & 11 & \(\mathbf{92.83}\) & \(73.56\) & \(\mathbf{65.03}\) & \(\mathbf{67.14}\) & \(76.36\) \\
$\Delta$: $A^2$ $-$ largest (pp) & -- & \(\mathbf{+4.52}\) & \(-3.24\) & \(\mathbf{+0.54}\) & \(\mathbf{+7.24}\) & \(-0.41\) \\
\bottomrule
\end{tabular}%
\endgroup
\end{table}

\section{Compute Resources}
\label{sec:compute_resources}

All experiments were run on a single node with 8$\times$NVIDIA A100 GPUs (80\,GB memory each).

\end{document}